\pdfoutput=1

\documentclass[11pt]{article}

\usepackage[preprint]{acl}

\usepackage{CJKutf8} 
\usepackage[utf8]{inputenc} 
\usepackage{times}
\usepackage{latexsym}
\usepackage{makecell} 
\usepackage{amsmath,amssymb,units}
\usepackage[ruled,vlined,linesnumbered]{algorithm2e}
\usepackage{subfigure}
\usepackage{enumitem}
\usepackage{import}
\usepackage{bm}
\usepackage{booktabs}
\usepackage{multirow}
\usepackage{array}
\usepackage{caption}
\usepackage{adjustbox}
\usepackage{balance}
\usepackage{hyperref}
\newcolumntype{C}{>{\centering\arraybackslash}p{0.7cm}}
\usepackage[T1]{fontenc}

\usepackage[utf8]{inputenc}

\usepackage{microtype}

\usepackage{inconsolata}

\usepackage{graphicx}

\newtheorem{definition}{Definition}
\newtheorem{Theorem}{Theorem}
\title{CoKV: Optimizing KV Cache Allocation via Cooperative Game}

\author{
Qiheng Sun$^{1,2}$, Hongwei Zhang$^{1,2}$, Haocheng Xia$^3$, Jiayao Zhang$^{1,2}$, Jinfei Liu$^{1,2}$\thanks{Jinfei Liu is the corresponding author.} Kui Ren$^1$\\
$^1$the State Key Laboratory of Blockchain and Data Security, Zhejiang University\\
$^2$Hangzhou High-Tech Zone (Binjiang) Institute of Blockchain and Data Security\\
$^3$Siebel School of Computing and Data Science\\
University of Illinois Urbana-Champaign\\
\texttt{\{qiheng\_sun,hongweizhang, jiayaozhang, jinfeiliu, kuiren\}@zju.edu.cn}\\
\texttt{hxia7@illinois.edu}\\
}

\begin{document}
\maketitle
\begin{abstract}
Large language models (LLMs) have achieved remarkable success on various aspects of human life. However, one of the major challenges in deploying these models is the substantial memory consumption required to store key-value pairs (KV), which imposes significant resource demands. Recent research has focused on KV cache budget allocation, with several approaches proposing head-level budget distribution by evaluating the importance of individual attention heads. These methods, however, assess the importance of heads independently, overlooking their cooperative contributions within the model, which may result in a deviation from their true impact on model performance. In light of this limitation, we propose CoKV, a novel method that models the cooperation between heads in model inference as a cooperative game. By evaluating the contribution of each head within the cooperative game, CoKV can allocate the cache budget more effectively. Extensive experiments show that CoKV achieves state-of-the-art performance on the LongBench benchmark using LLama-3-8B-Instruct and Mistral-7B models. Code is provided in \url{https://github.com/nawei1010/CoKV}.
\end{abstract}

\section{Introduction}
Large language models (LLMs) are widely applied across various domains, including content generation~\cite{li2024pre}, automated services~\cite{chen2024large}, and decision support systems~\cite{bao2023tallrec}. To enhance the application capabilities of large language models, it is essential for them to handle long texts. For example, GPT-4~\cite{gpt4} and Llama-3~\cite{llama3} support a context size of 128k tokens, while the context size of Claude 3~\cite{claude3} is up to 200k tokens. LLMs consist of multiple transformer blocks that store key and value states (KV) during inference. KV cache allows efficient decoding in token generation without recomputing key and value states by using previously cached KV pairs. However, the KV cache grows excessively large when dealing with long texts, inevitably straining GPU memory and increasing decoding latency.

Eviction of less important key and value states in the cache has garnered significant attention. Many studies have explored methods for ranking the importance of tokens within a single attention head, retaining only the top $k$ most significant ones. For example, H2O~\cite{H2O} evaluates token importance using the sum of attention weights. StreamingLLM~\cite{StreamingLLM} directly removes KV from the middle segment of the cache to reduce the cache size as they incorporate less information. SnapKV~\cite{SnapKV} calculates token scores by pooling the attention weights between tokens in the local window and those in the cache. Recently, some studies have recognized that the importance of each attention head varies, enabling methods like AdaKV~\cite{AdaKV} and HeadKV~\cite{HeadKV}. AdaKV improves budget utilization by adaptively allocating the overall budget across different attention heads based on their varied concentration degrees. Heads with sparse concentrations require a small cache proportion, whereas more dispersed heads demand larger allocations. HeadKV evaluates the retrieval-reasoning scores of different heads and allocates a larger cache size to those with higher scores.

Motivated by evidence that attention heads vary in importance, we propose a novel approach to better evaluate and utilize this variability. We identify two key insights. First, existing methods evaluate attention head importance independently. For example, AdaKV evaluates the concentration degrees of heads while HeadKV assesses the retrieval-reasoning capability of each head in isolation as a measure of importance. However, these approaches treat heads as isolated units, overlooking the fact that their true importance emerges from their cooperation rather than individual capabilities. As a result, independently assessing head importance may lead to suboptimal allocation. Second, existing methods evaluate head importance in a task-agnostic manner. However, heads that play a critical role in query answering may not hold the same level of significance in code generation. Consequently, applying the same importance scores to heads across all tasks within a model may fail to reflect the practical need of each specific task accurately. Based on these insights, we propose \textbf{CoKV} 
 (\underline{\textbf{Co}}operation-based \underline{\textbf{K}}ey-\underline{\textbf{V}}alue), a method that evaluates the contribution of all attention heads in their cooperation within the model and dynamically allocates cache budgets based on their contribution to the specific task.

CoKV is inspired by the Shapley value~\cite{shapley} from cooperative game theory. The Shapley value of a player $p_i$ measures the expected marginal contribution that $p_i$ provides to a coalition of players. Similarly, we can use the Shapley value to assess the importance of each attention head by viewing each head as a player. Marginal contribution is defined as $\mathcal{U}(\mathcal{S} \cup \{p_i\}) - \mathcal{U}(\mathcal{S})$ where \( \mathcal{S} \) is a coalition of players excluding $i$ and \( \mathcal{U} \) is the utility function. A simple intuition for computing the Shapley value of each head in the model is to define $\mathcal{U}$ as the model performance metric. However, calculating the Shapley value is \#P-hard~\cite{deng1994complexity}, as there are an exponential number of coalitions and corresponding marginal contributions. As a result, evaluating the Shapley value for each head in LLMs requires an enormous number of model inferences. Although many studies~\cite{jia2019towards, mitchell2022sampling} have explored approximating the Shapley value to reduce computational costs, the process remains costly.

The computational bottleneck in calculating the Shapley value arises from the fact that each sample of the marginal contribution only can be applied to a single player. Fortunately, Shapley value can be expressed as the expectation of the weighted complementary contribution, defined as \( \mathcal{U}(\mathcal{S}) - \mathcal{U}(\mathcal{N} \setminus \mathcal{S}) \), where \( \mathcal{N} \) represents the set of all players~\cite{cc}. Complementary contribution has an advantage over marginal contribution is that \( \mathcal{U}(\mathcal{S}) - \mathcal{U}(\mathcal{N} \setminus \mathcal{S}) \) can be used to update the Shapley values for all players \( i \in \mathcal{S} \). By expressing the Shapley value in terms of complementary contributions, we can interpret it as an expectation over these contributions computed at different coalition sizes \( |\mathcal{S}| \). However, in the LLM setting, the cost of computing the complementary contributions in all coalition sizes is still prohibitively high. We observe that the average complementary contribution at each coalition size exhibits a strong correlation with the Shapley value of the players in Appendix Section~\ref{sec:coalition_distribution}. This insight allows us to approximate attention head importance by computing complementary contributions at only a few selected coalition sizes, rather than evaluating all possible sizes (i.e., from 1 to $|\mathcal{N}|$). By focusing on a few representative coalition sizes, we can significantly reduce the computational cost of estimating the contributions of heads. Additionally, we provide a theoretical analysis of this approach and demonstrate its efficiency.

CoKV is a simple-yet-effective method and can integrate well with other inference optimization techniques. We integrate CoKV with widely used methods including FlashAttention~\cite{flashattention} and grouped-query attention (GQA)~\cite{gqa}. CoKV achieves state-of-the-art performance in LongBench~\cite{longbench} using Llama-3-8B-Instruct~\cite{llama3} and Mistral-7B~\cite{mistral} models. Results from the Llama-3-8B-Instruct model show that when each KV cache retains an average of 128 KV pairs ($1.6\%$ of the full cache), it achieves $97.29\%$ of the performance of the full KV cache. Furthermore, when each cache retains just 512 tokens on average, CoKV outperforms the full KV cache in terms of average accuracy. This demonstrates that CoKV not only reduces computational costs but also improves inference performance by identifying which heads benefit from cache retention and which may have a detrimental effect. Additionally, we evaluate all methods within the token range of 1k to 31k in the Needle-in-a-Haystack test, where CoKV also demonstrated the best retrieval capability.

\section{Preliminaries}
\subsection{Key-Value Caching and Compression} 
In Multi-Head Attention (MHA), for each attention head $h_i$ in one layer, the embedded input $X = \{x_1, x_2, \dots, x_m\} \in \mathbb{R}^{m \times d_{\text{model}}}$ of $m$ tokens is mapped into different subspaces using query $W^Q_i$, key $W^K_i$, and value $W^V_i \in \mathbb{R}^{d_{\text{model}} \times d_h}$ matrices:
\begin{align}\nonumber
    Q_i = X W^Q_i, K_i = X W^K_i, V_i = X W^V_i \in \mathbb{R}^{m \times d_h}
\end{align}
where $d_h$ is the dimension of attention heads, $d_h=d/{\tau}$, and $\tau$ is the number of heads in one layer.

All the computed $KV$ for the input sequence are cached to avoid recalculating them during the subsequent decoding stages. Assume there is a new input token $x\in \mathbb{R}^{1 \times d_{\text{model}}}$, then it will be mapped to a new query, key, and value as follows,
\begin{align}\nonumber
    q_i = x W^Q_i, k_i = x W^K_i, v_i = x W^V_i \in \mathbb{R}^{1 \times d_h}.
\end{align}

The KV cache is updated by adding the new key and value pair
$$K_i = \operatorname{Cat}[K_i,k_i], V_h=\operatorname{Cat}[V_i,v_i].$$
The attention output is computed as follows,
$$O_i=A_iV_i$$ where $A_i=\operatorname{softmax}(q_iK_i^T/\sqrt{d_h})$. The final output $y \in \mathbb{R}^{1\times d_{\text{model}}}$ is obtained through a linear transformation $$
y=\operatorname{Cat}[O_1,\cdots,O_\tau]W^O$$ where $W^O\in \mathbb{R}^{d \times d_{\text{model}}}$ output weight matrix.

Furthermore, KV cache eviction methods can be employed to discard unimportant KV cache entries while preserving performance. For each head $h_i$, the compressed KV cache is reduced to $\hat{K}_i \in \mathbb{R}^{s \times d_h}$ and $\hat{V}_i \in \mathbb{R}^{s \times d_h}$, where some unimportant KV pairs are evicted and $s \ll m$, resulting in a significant improvement in computational efficiency and memory usage. Specifically, the compressed KV cache is updated by appending the new key and value pair:
   \[
   \hat{K}_i = \text{Cat}[\hat{K}_i, k_i], \quad \hat{V}_i = \text{Cat}[\hat{V}_i, v_i].
   \]
The attention output for each head \( h_i \) is computed using the compressed KV cache:
   \[
   \hat{O}
   _i = \hat{A}_i \hat{V}_i,
   \]
   where the attention weights \( A_i \) are calculated as:
   $\hat{A}_i = \text{softmax}(q_i \hat{K}_i^T/\sqrt{d_h}).$

\subsection{Shapley Value}

Consider a set of players $\mathcal{N}=\{p_1,\ldots,p_n\}$. A \emph{coalition} $\mathcal{S}$ is a subset of $\mathcal{N}$ that cooperates to complete a task. A utility function $\mathcal{U}(\mathcal{S})$ $(\mathcal{S} \subseteq \mathcal{N})$ is the utility of coalition $\mathcal{S}$ for the task. The marginal contribution of player $p_i$ with respect to a coalition $\mathcal{S}$ is $\mathcal{U}(\mathcal{S}\cup \{p_i\})-\mathcal{U}(\mathcal{S})$. 
The Shapley value measures the expectation of marginal contribution of player $p_i$ in all possible coalitions. That is
\begin{equation}\label{equ:SV}
  \mathcal{SV}_i=\frac{1}{n} \sum_{\mathcal{S}\subseteq \mathcal{N} \setminus \{p_i\}}\frac{\mathcal{U}(\mathcal{S}\cup \{p_i\})-\mathcal{U}(\mathcal{S})}{\binom{n-1}{|\mathcal{S}|}}.
\end{equation} 
According to Equation~\ref{equ:SV}, it is evident that computing the exact Shapley value requires enumerating the utilities for all possible subsets of players and each marginal contribution can only be used to update the Shapley value of a single player. Therefore, the computational complexity of exactly calculating the Shapley value is exponential. Recently, the Shapley value of player $p_i$ is proven to be equal to the weighted complementary contributions~\cite{cc} as follows,
\begin{equation}\label{equ:svs}
    \mathcal{SV}_i = \frac{1}{n}\sum_{\mathcal{S}\subseteq \mathcal{N}\setminus\{p_i\}}\frac{\mathcal{U}(\mathcal{S})-\mathcal{U}(\mathcal{N} \setminus\mathcal{S})}{\binom{n-1}{|\mathcal{S}|}}.
\end{equation}
$\mathcal{U}(\mathcal{S})-\mathcal{U}(\mathcal{N} \setminus\mathcal{S})$ is called complementary contribution which has an advantage that can be reused to update Shapley value estimation for all players in $\mathcal{S}$. In the context of KV caches, attention heads are treated as players for evaluating their importance to each specific task. $\mathcal{U}(\mathcal{S})$ is defined as the model accuracy when the attention heads in $\mathcal{N}\setminus{\mathcal{S}}$ are masked, we retain only the KV pairs within the local window for masked heads.

\section{Method}
\begin{figure*}[htbp] 
    \centering
    \includegraphics[width=1\textwidth]{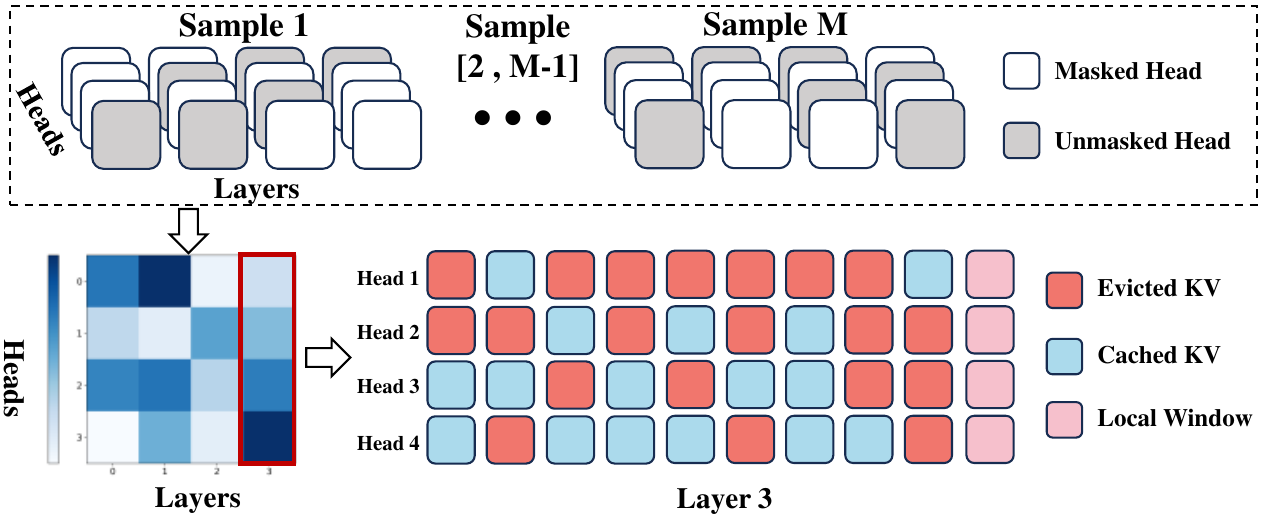} 
    \caption{Overview of our proposed method:
    (1) \textbf{Head Importance Evaluation (Upper Part):} For a 4-layer × 4-head model, We measure head importance using the Sliced Shapley Value (SSV). To approximate SSV, we sample $M$ different sets of masked heads and compute their complementary contributions. The average complementary contribution of each head is its estimated SSV.
    (2) \textbf{KV Cache Compression (Lower Part):} Using the 4 heads in Layer 3 as an example, all heads store KV pairs for a small local window of recent tokens, while heads with higher SSV (darker in the heatmap) are allocated more cache size to retain KV pairs before the local window.
    }  
    \label{fig:overview}
\end{figure*}

Our method consists of two phases. First, we precompute the importance scores for each attention head. Second, these scores are utilized for KV cache compression during inference. The overview of our approach is illustrated in Figure~\ref{fig:overview}.

\subsection{Head Importance Evaluation}
Although the complementary contribution helps in increasing efficiency when approximating the Shapley value, it is still computationally costly, especially in the LLM setting. Given a set of players $\mathcal{N}=\{p_1,\ldots,p_n\}$, a coalition of $j$ players $(1 \leq j \leq n)$ is called a \textit{$j$-coalition}. Moreover, for a player $p_i$ $(1\leq i \leq n)$, a $j$-coalition that contains $p_i$ is called a \textit{$(i, j)$-coalition}. Denote by $\mathfrak{S}_{i,j} = \{\mathcal{S}\cup\{p_i\}|\mathcal{S}\subseteq\mathcal{N}\setminus\{p_i\},|\mathcal{S}|=j-1\}$ the set of $(i, j)$-coalitions, and by $\mathcal{SV}_{i,j}$ \textit{the expected complementary contributions of} $(i, j)$-coalitions. That is,
$$\mathcal{SV}_{i,j} = \sum_{\mathcal{S} \in \mathfrak{S}_{i,j}} \frac{\mathcal{U}(\mathcal{S})-\mathcal{U}(\mathcal{N}\setminus\mathcal{S})}{\binom{n-1}{j-1}}.$$
It is clear that $\mathcal{SV}_i = \frac{1}{n}\sum_{j=1}^{n}\mathcal{SV}_{i, j}$. Computing the Shapley value needs to calculate $\mathcal{SV}_{i,j}$ for $j$ ranging from 1 to $n$, which becomes costly when $n$ is large. 

We observe that the expected complementary contributions of $j$-coalitions for heads in LLMs follow a similar distribution across different $j$ values, as shown in Appendix Section~\ref{sec:coalition_distribution}. This suggests that the contributions of heads can be effectively captured using a subset of $j$-coalitions.  Based on this insight, we propose assessing the importance of heads using the expected complementary contribution of several $j$-coalitions, which can significantly reduce the computation cost while maintaining effectiveness. Formally, we introduce a new definition called the Sliced Shapley value as follows.

\begin{definition}
\normalfont \textbf{(Sliced Shapley Value)} Let $\mathcal{H} \subseteq \{1,\cdots,n\}$ denote the selected set of $j$-coalitions, representing a specific slice of the coalition size space. The \emph{Sliced Shapley value} of head $h_i$ with respect to $\mathcal{H}$ is defined as:
\[
\mathcal{SSV}_i^{\mathcal{H}} = \frac{1}{|\mathcal{H}|} \sum_{j=1}^{n} \mathcal{SV}_{i,j} \cdot \mathbb{I}_j^{|\mathcal{H}|},
\]
where $\mathbb{I}_j^{\mathcal{H}}$ is an indicator function, which is 1 if $j$ is the element in $\mathcal{H}$ and 0 otherwise. 
\end{definition}

\paragraph{Algorithm Description.}The detailed steps of approximating $\mathcal{SSV}_i^{\mathcal{H}}$ are shown in Algorithm~\ref{Alg:approSV}. In each iteration, sample a random permutation \( \pi^k \) of the heads \( \{h_1, \dots, h_n\} \), which defines a random ordering of the heads. Randomly select a split point and create a set \( \mathcal{S} \) of selected heads. Mask heads in the set \( \mathcal{N} \setminus \mathcal{S} \), and evaluate the model accuracy after masking, which is denoted as \( \mathcal{U}(\mathcal{S}) \). Similarly, calculate $\mathcal{U}(\mathcal{N} \setminus \mathcal{S})$ by masking heads in $\mathcal{S}$ (Lines 3-6). For each head in $\mathcal{S}$, update \( \mathcal{SV}_{\pi^k(j), i} \) and count matrix \( m_{\pi^k(j), i} \) (Lines 7-10). After $\mathcal{M}$ iterations are completed, calculate the approximated Sliced Shapley value for each head by averaging the complementary contributions.

\begin{Theorem}\label{theorem:SSV_sampling}
    Algorithm~\ref{Alg:approSV} returns an $(\epsilon,\delta)$-approximation of Sliced Shapley value with time complexity $\mathcal{O}( \frac{T|\mathcal{H}|ln\frac{2|\mathcal{H}|}{\delta}}{\epsilon^2})$ where T is the time cost of evaluating a complementary contribution which is the time to inference on the validation dataset of each task in our setting. In contrast, Shapley value requires the time complexity of $\mathcal{O}( \frac{Tnln\frac{2n}{\delta}}{\epsilon^2})$ to achieve an $(\epsilon,\delta)$-approximation. The proof is provided in Appendix Section~\ref{sec:proof}.
\end{Theorem}{}

\begin{algorithm}[t] \caption{Evaluating Head Importance in LLMs.}\label{Alg:approSV}
\SetKwInOut{Input}{input}\SetKwInOut{Output}{output}

\Input{Heads $\mathcal{N} = \{h_1, \ldots, h_n\}$ and sampling number $\mathcal{M} >0$}
\Output{approximate Sliced Shapley value $\overline{\mathcal{SSV}^\mathcal{H}_i}$ for each head $h_i$ $(1 \leq i \leq n)$}

    $\overline{\mathcal{SV}_i^{\mathcal{H}}} \gets 0$ $(1 \leq i \leq n)$;
    $\overline{\mathcal{SV}_{i,j}},m_{i,j} \gets 0$ $(1 \leq i,j \leq n)$;

    \For{k=1 to $\mathcal{M}$}{
        let $\pi^k$ be a random permutation of $\{1, \ldots, n\}$;
        
        let $i$ be a randomly selected element from the set $\mathcal{H}$;
        
        $\mathcal{S} \gets \{\pi^k(1), \ldots,\pi^k(i)\}$;
        
        $\mathcal{N}\setminus\mathcal{S} \gets \{\pi^k(i+1), \ldots, \pi^k(n)\}$;\\
        {\scriptsize \tcp{ $\mathcal{U}(\mathcal{S})$ is the model performance when heads in $\mathcal{N} \setminus \mathcal{S}$ are masked and vice versa for $\mathcal{U}(\mathcal{N} \setminus \mathcal{S})$.}}
        
        $u \gets \mathcal{U}(\mathcal{S})-\mathcal{U}(\mathcal{N}\setminus\mathcal{S})$;\\
        
        \For{j=1 to i}{
            $\overline{\mathcal{SV}_{\pi^k(j),i}}+=u$;
            
            $m_{\pi^k(j),i} += 1$;
        }
    }
    \For{i = 1 to $n$}{
        $\overline{\mathcal{SSV}_{i}^{\mathcal{H}}}=\frac{1}{\mathcal{H}}\sum_{j=1}^{n} \overline{\mathcal{SV}_{i,j}} / m_{i, j}$\;
    }
    \Return $\overline{\mathcal{SSV}_1^{\mathcal{H}}}, \ldots, \overline{\mathcal{SSV}_n^{\mathcal{H}}}$.
\end{algorithm}

\subsection{KV Cache Compression}

Existing KV cache compression methods have partially addressed the importance of layers, yet this consideration remains insufficient during cache allocation. While AdaKV attempts to preserve tokens with larger attention weights across all heads when allocating cache size, it overlooks the varying importance of different attention heads. Conversely, HeadKV acknowledges the differential importance of attention heads but suffers from several limitations. First, its evaluation primarily relies on the retrieval capability of individual heads, incorporating only basic reasoning abilities that prove inadequate for more complex scenarios, such as few-shot learning. Second, it assesses each head in isolation, ignoring the discrepancy between a head's individual contribution and its collaborative importance when working in conjunction with other heads. Our proposed method addresses these limitations by introducing a SSV-based scoring mechanism, which evaluates each head's importance based on its collaborative contribution to the task. This approach offers a more comprehensive and accurate representation of each head's significance in the overall model inference process.

\paragraph{Budget Allocation.}
An intuitive approach suggests that the least important heads, which contribute minimally or even negatively to the model performance, may not require cache allocation. Let \( \alpha \) represent the number of such heads, which serves as the sole hyperparameter in our allocation scheme. For the remaining \( n - \alpha \) heads, we employ a 
normalization method to normalize their importance scores and allocate the cache size proportionally based on their normalized scores.

Specifically, we normalize their contributions using min-max normalization for the \( n - \alpha\) heads:
\[
\mathcal{NSV}_i^{\mathcal{H}} = \frac{\mathcal{SSV}_i^{\mathcal{H}} - \min^{\alpha}(\mathcal{SSV}^{\mathcal{H}})}{\max(\mathcal{SSV}^{\mathcal{H}}) - \min^{\alpha}(\mathcal{SSV}^{\mathcal{H}})},
\]
where \( \min^{\alpha}(\cdot) \) and \( \max(\cdot) \) extract the $\alpha$-th smallest and maximum value, respectively. For the $\alpha$ heads with the smallest Sliced Shapley values, we set the normalized score as 0. This ensures that all normalized scores lie in the range \([0, 1]\).

Next, the cache size \( c_i \) allocated to head \( h_i \) is determined by the local window size $s$ and linearly distributing the remaining shared cache size \( B \) based on the normalized scores:

\begin{equation}\label{equ:alloc}
    c_i = B \cdot \frac{\mathcal{NSV}_i^{\mathcal{H}}}{\sum_{j=1}^{n} \mathcal{NSV}_j^{\mathcal{H}}}+s.
\end{equation}

\paragraph{Algorithm Description.}First, we allocate the KV cache size for each head based on their normalized Sliced Shapley values. Next, we rank the importance of KV pairs within each head using SnapKV. Specifically, the most recent tokens within local windows guide the KV cache selection. Attention scores from these local windows to the remaining tokens are aggregated via pooling, with higher-scoring tokens retained in the cache for each head. The detailed eviction steps for a single head are outlined in Algorithm~\ref{Alg:CoKV}.

\begin{algorithm}[t] \caption{Token Eviction Using CoKV.}\label{Alg:CoKV}
\SetKwInOut{Input}{input}\SetKwInOut{Output}{output}

\Input{Shared budget size $B$, local window size $s$, tokens in local window $X^{win}\in \mathbb{R}^{s \times d}$, KV in local window $\{K_i^{win},V_i^{win}\}$, KV outside local window $\{K_i^{out},V_i^{out}\}$}
\Output{Retained KV Cache $\{\hat{K}_i,\hat{V}_i\}$}
    $Q_i^{win} = X^{win}W_i^Q$;\\
    {\scriptsize \tcp{Compute attention weights of queries in local window and prefix Keys.}}
    $\overline{A}_i=\operatorname{softmax}(Q_i^{win}K_i^T)$;\\
    $\overline{A}_i=\overline{A}_i.maxpooling(dim=1).mean(dim=0)$;\\
    {\scriptsize \tcp{Calculate token scores outside the local window.}}
    Get $c_i$ using Algorithm~\ref{Alg:approSV} and Equation~\ref{equ:alloc};\\
    $indices=\overline{A}_i.topk(c_i).indices$;\\
    Select $\{\hat{K}_i,\hat{V}_i\}$ from $\{K_i^{out},V_i^{out}\}$ according $indices$;\\
    $\{\hat{K}_i,\hat{V}_i\}=$  Cat($\{\hat{K}_i,\hat{V}_i\},\{K_i^{win},V_i^{win}\}$);\\
    {\scriptsize \tcp{Keep top $c_i$ KV pairs in the cache.}}
    \Return Retained KV Cache $\{\hat{K}_i,\hat{V}_i\}$.
\end{algorithm}

\section{Experiments}

\subsection{Experiment Settings}
\paragraph{Datasets.}LongBench 
is a multitask benchmark for long context understanding and exhibits a wide range of average input lengths, spanning from 1,235 to 18,409 tokens.

\paragraph{Baselines and Settings.}
We compare CoKV with four strong KV cache compression methods. All methods keep the same total cache size for fair comparison. Besides, we implement all methods with GQA~\cite{gqa} and FlashAttention~\cite{flashattention} for efficient computation.
\begin{itemize}[leftmargin=*, itemsep=0pt, parsep=0pt]
\item \textbf{SnapKV}~\cite{SnapKV} uses the last several tokens as local windows to guide KV cache selection. Attention scores from these windows to the remaining tokens are pooled to cluster and guide the selection process.
\item \textbf{PyramidKV}~\cite{PyramidKV} allocates more KV cache to lower layers to retain key information while reducing the budget for higher layers where information is already aggregated.
\item  \textbf{Ada-KV}~\cite{AdaKV} dynamically allocates budgets to heads within each layer based on their concentration degrees, and can be combined with SnapKV or PyramidKV. Ada-SnapKV is used as the baseline due to its superior performance over Ada-PyramidKV.
\item \textbf{HeadKV-R2}~\cite{HeadKV} allocate budgets to heads based on their retrieval-reasoning score, and it uses SnapKV to rank the importance of KV pairs in each head.
\end{itemize}

In CoKV, we allocate the KV cache size for each head based on the normalized Sliced Shapley value of $\mathcal{H}=\{32, 64, 96, 128\}$. Following HeadKV-R2, we set the local window size to 8, and randomly split each dataset into a validation dataset and a test dataset, with proportions of 15\% and 85\%, respectively. The hyperparameter $\alpha$ is selected from the set $\{1, 5, 10, 15, 20, 30, 40\}$. The validation dataset is used to compute Sliced Shapley value and determine the optimal $\alpha$ for each task. We evaluate CoKV on the Llama-3-8B-Instruct and Mistral-7B-Instruct-v0.2 models. Due to the page limit, the Mistral-7B-Instruct-v0.2 results are provided in \nameref{sec:appendix}.  For test data that exceeds the maximum input length of Llama-3-8B-Instruct, we adopt the approach of HeadKV by utilizing the first 4k tokens and the last 4k tokens.  Following standard practices in prior studies~\cite{AdaKV,HeadKV}, we perform cache eviction after the prefilling phase of each layer for consistent comparison. In GQA, a group of 4 heads shares the same KV cache. We treat each cache within a group as a player in a cooperative game, evaluating their Sliced Shapley value to determine their importance scores. For HeadKV-R2, we calculate the importance score of each group by averaging the retrieval-reasoning scores of the 4 heads within the group. This adaptation ensures compatibility with GQA, as HeadKV is implemented with MHA in the original paper. For the efficiency and computation cost analysis of Sliced Shapley value, please refer to Appendix Section~\ref{section:compute_efficiency}. For the test in Needle-in-a-Haystack, please refer to Appendix Section~\ref{sec:needle_test}.

\subsection{Main Results}
\begin{figure}[h] 
    \centering
    \includegraphics[width=0.4\textwidth]{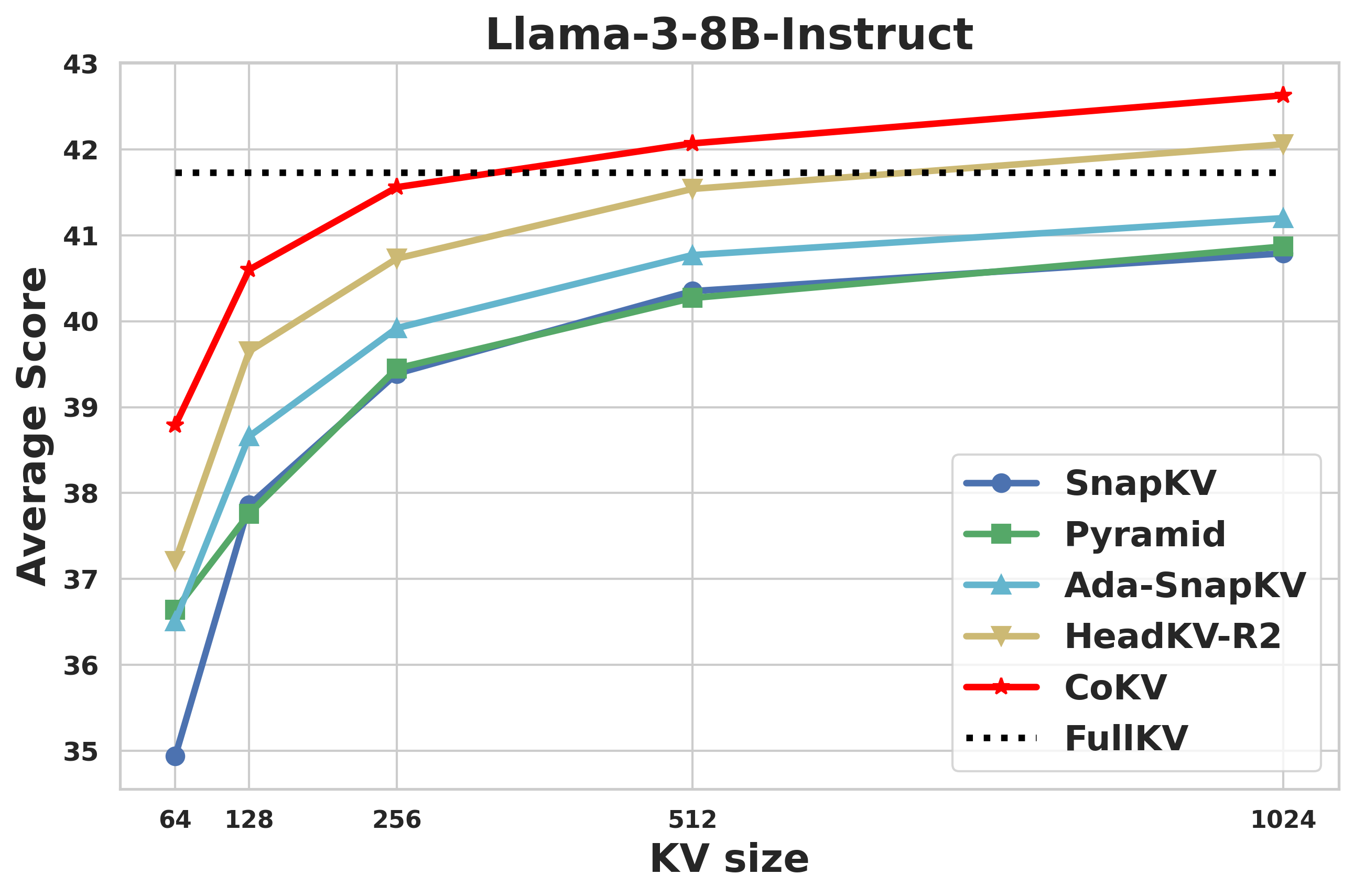}
    \includegraphics[width=0.4\textwidth]{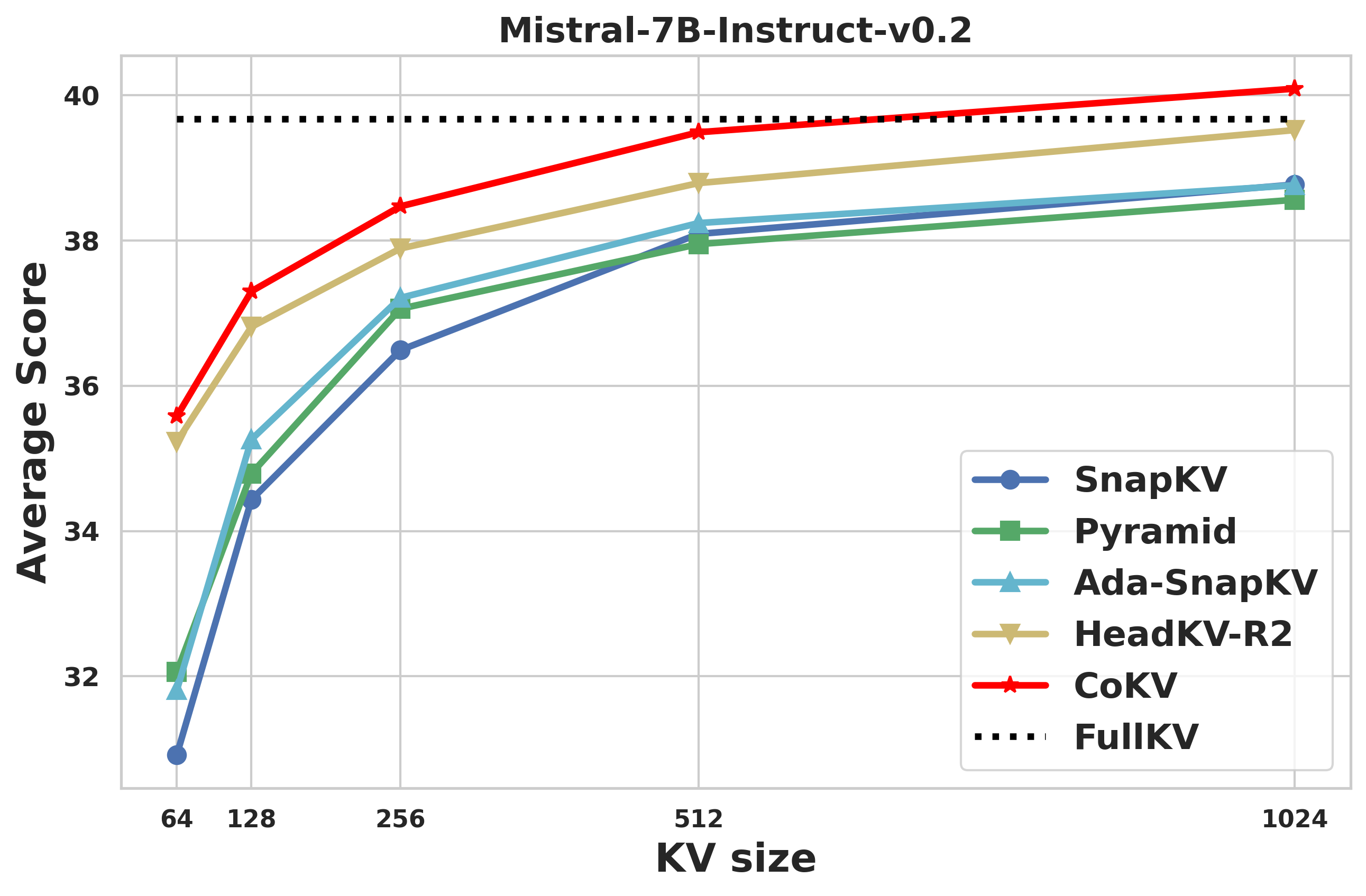} 
    \caption{Results for varying KV cache sizes (64, 128, 256, 512, 1024), showing the average accuracy across 16 datasets from the LongBench benchmark.}
    \label{fig:Llama_Mistral_Longbench}
\end{figure}

\paragraph{Benchmark Results.} 
The complete benchmark results are presented in Tables~\ref{tab:llama_results} and~\ref{tab:mistral_results} in the appendix. We include a simplified table (Table~\ref{tab:short_llama_results}), showing the performance of Llama-3-8B-Instruct when keeping 64-128 KV pairs on average.
\begin{table*}[h]
\centering \caption{{Benchmark Results of Llama-3-8B-Instruct}} \label{tab:short_llama_results}
\begin{adjustbox}{width=\textwidth}
\begin{tabular}{lCCCCCCCCCCCCCCCCC}
\toprule
\multirow{2}{*}{Method} & \multicolumn{3}{c}{Single-Doc. QA} & \multicolumn{3}{c}{Multi-Doc. QA} & \multicolumn{3}{c}{Summarization} & \multicolumn{3}{c}{Few-shot Learning} & \multicolumn{2}{c}{Synthetic} & \multicolumn{2}{c}{Code} \\
\cmidrule(lr){2-4} \cmidrule(lr){5-7} \cmidrule(lr){8-10} \cmidrule(lr){11-13} \cmidrule(lr){14-15} \cmidrule(lr){16-17}
& \rotatebox{-45}{\small NtrQA} 
& \rotatebox{-45}{\small Qasper} 
& \rotatebox{-45}{\small MF-en} 
& \rotatebox{-45}{\small HotpotQA} 
& \rotatebox{-45}{\small 2WikiMQA} 
& \rotatebox{-45}{\small Musique} 
& \rotatebox{-45}{\small GovReport} 
& \rotatebox{-45}{\small QMSum} 
& \rotatebox{-45}{\small MultiNews} 
& \rotatebox{-45}{\small TREC} 
& \rotatebox{-45}{\small TriviaQA} 
& \rotatebox{-45}{\small SAMSum} 
& \rotatebox{-45}{\small PCount} 
& \rotatebox{-45}{\small PRe} 
& \rotatebox{-45}{\small Lcc} 
& \rotatebox{-45}{\small RB-P} \\
\midrule
Full Cache       & 24.12 & 31.24 & 39.85 & 45.23 & 34.56 & 21.09 & 28.38 & 23.24 & 26.52 & 74.12 & 90.96 & 42.37 & 4.55 & 71.76 & 58.1 & 51.64 \\
\bottomrule

\multicolumn{17}{c}{\textbf{KV size=64}} \\  
\midrule[0.5pt]

SnapKV           
&  19.94
&  13.21
&  28.91
&  40.06
&  28.58
&  \textbf{18.12}
&  17.29
&  21.71
&  17.05
&  49.41
&  89.00
&  35.48
&  3.99
&  71.57
&  54.35
&  50.42\\
Pyramid          
&  20.11
&  16.54
&  32.67
&  40.25
&  27.71
&  17.54
&  18.67
&  \textbf{22.37}
&  20.03
&  62.55
&  89.89
&  36.63
&  4.30
&  71.76
&  54.27
&  50.96\\
Ada-SnapKV       
&  20.40
&  14.46
&  32.62
&  42.39
&  31.48
&  17.58
&  18.57
&  22.18
&  18.71
&  58.82
&  90.13
&  35.25
&  4.41
&  71.57
&  54.02
&  51.68\\
HeadKV-R2        
&  20.30
&  16.76
&  \textbf{35.96}
&  38.08
&  26.41
&  17.98
&  \textbf{18.68}
&  21.75
&  \textbf{20.58}
&  67.06
&  88.19
&  37.30
&  3.21
&  71.76
&  \textbf{56.20}
&  54.49\\
CoKV             &\textbf{20.77} & \textbf{19.67} & 35.11 & \textbf{44.37} & \textbf{34.36} & 17.83 & 17.89 & 22.33 & 18.55 & \textbf{71.76} & \textbf{90.73} & \textbf{38.51} & \textbf{4.71} &  \textbf{71.76} & 55.45 & \textbf{55.82} \\
\bottomrule

\multicolumn{17}{c}{\textbf{KV size=1024}} \\  
\midrule[0.5pt]

SnapKV          
&  23.95
&  26.95
&  37.81
&  44.03
&  30.88
&  20.93
&  24.26
&  23.09
&  25.79
&  72.35
&  90.87
&  41.43
&  4.31
&  71.76
&  59.29
&  54.91 \\
Pyramid          
& 23.62 & 26.76 & 39.44 & 45.79 & 33.41 & 19.87 & 23.57 & 22.98 & 25.13 & 73.02 & 90.93 & 40.86 & 4.71 & 71.76 & 58.43 & 53.67 \\
Ada-SnapKV       
&  23.52
&  28.33
&  40.39
&  45.20
&  32.95
&  20.11
&  24.55
&  23.33
&  25.37
&  73.53
&  90.87
&  41.38
&  4.46
&  71.76
&  58.88
&  54.65\\
HeadKV-R2        
&  23.35
&  29.60
&  40.09
&  45.82
&  35.81
&  \textbf{21.39}
&  \textbf{25.57}
&  23.32
&  \textbf{26.30}
&  74.12
&  90.77
&  40.27
&  4.19
&  \textbf{71.76}
&  61.58
&  59.03\\
CoKV             & \textbf{24.01} & \textbf{31.70} & \textbf{40.64} & \textbf{48.13} & \textbf{37.89} & 20.64 & 23.02 & \textbf{23.89} & 25.71 & \textbf{74.12} & \textbf{91.01} & \textbf{42.02} & \textbf{4.71} & 71.20 & \textbf{63.33} & \textbf{63.74} \\
\bottomrule

\end{tabular}
\vspace{0.5em}
\footnotesize
\end{adjustbox}
\end{table*}
The results demonstrate that CoKV consistently outperforms all baseline methods. The average accuracy of the two models on 16 datasets are presented in Figure ~\ref{fig:Llama_Mistral_Longbench}. 
Notably, in Llama-3-8B-Instruct, with an average of 128 tokens cached per group KV cache, CoKV retains 97.29\% of the model performance. Furthermore, CoKV significantly surpasses FullKV when it maintains an average of over 512 KV pairs per group cache. When retains an average of 1024 KV, the average results of both models outperform FullKV. This demonstrates that CoKV achieves near-lossless performance under resource-constrained settings. The superior performance of CoKV arises from its ability to effectively evaluate the importance of each cache within a group while considering the cooperation among all groups. It is not only capable of identifying which groups are important but also able to recognize those groups that do not contribute or even have a negative contribution. By optimizing the cache size to enhance overall cooperation, CoKV ensures efficient and high-quality inference.

\begin{figure}[h] 
    \centering
    \includegraphics[width=0.4\textwidth]{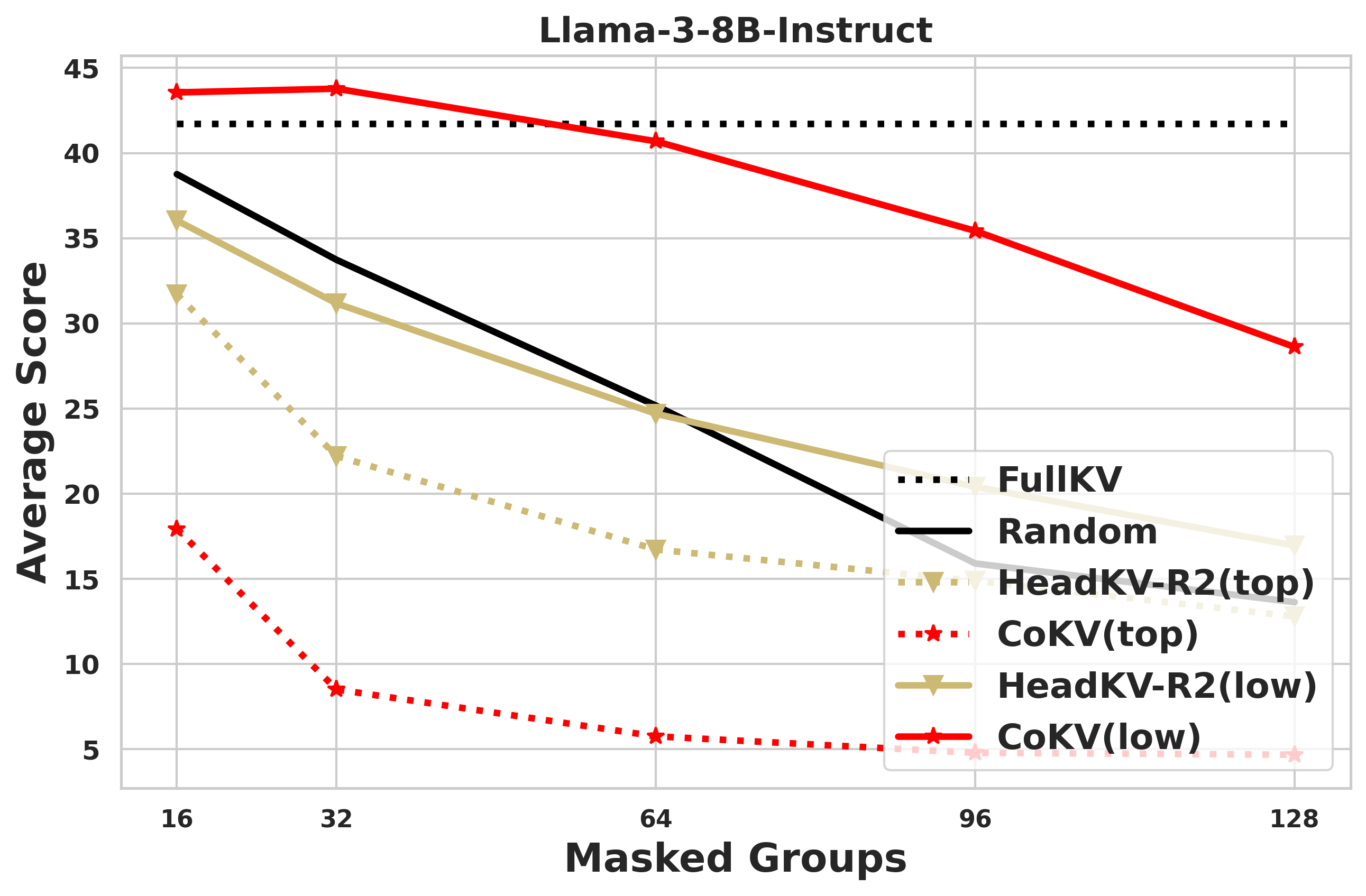} 
    \includegraphics[width=0.4\textwidth]{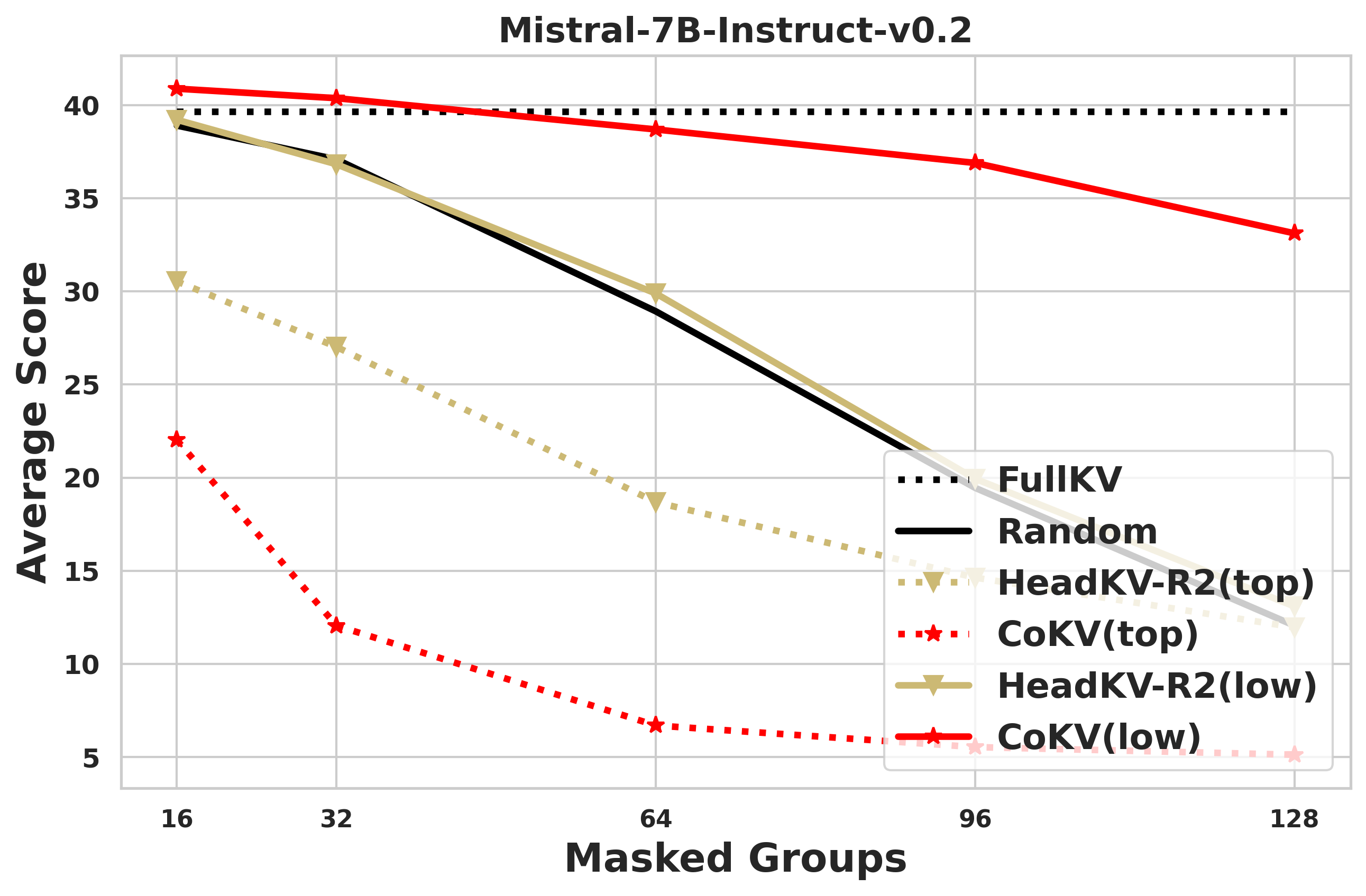} 
    \caption{Results for varying masked groups (16,32,64,96,128), showing the average accuracy across 16 datasets from the LongBench benchmark.}
    \label{fig:mask_llama_mistral_average}
\end{figure}

\paragraph{Hyperparameter Free Results.}\label{experiment:mask} Since both HeadKV-R2 and CoKV provide importance scores for each group, we conduct an experiment to compare their effectiveness without introducing any additional hyperparameters. In this experiment, we mask the caches of groups based on the importance scores assigned by each algorithm. Specifically, we mask the caches of both the highest-ranked (\textbf{top}) and lowest-ranked groups (\textbf{low}). The complete results are shown in Tables~\ref{tab:mask_llama} and~\ref{tab:mask_mistral} in the appendix. We include a simplified table for the results of masking 16,128 groups of Llama-3-8B-Instruct model in Table~\ref{tab:short_mask_llama}. 
\begin{table*}[h]
\centering 
\caption{{Results of masking groups with Llama-3-8B-Instruct}}
\label{tab:short_mask_llama}
\begin{adjustbox}{width=\textwidth}
\begin{tabular}{lCCCCCCCCCCCCCCCCC}
\toprule
\multirow{2}{*}{Method} & \multicolumn{3}{c}{Single-Doc. QA} & \multicolumn{3}{c}{Multi-Doc. QA} & \multicolumn{3}{c}{Summarization} & \multicolumn{3}{c}{Few-shot Learning} & \multicolumn{2}{c}{Synthetic} & \multicolumn{2}{c}{Code} \\
\cmidrule(lr){2-4} \cmidrule(lr){5-7} \cmidrule(lr){8-10} \cmidrule(lr){11-13} \cmidrule(lr){14-15} \cmidrule(lr){16-17}
& \rotatebox{-45}{\small NtrQA} 
& \rotatebox{-45}{\small Qasper} 
& \rotatebox{-45}{\small MF-en} 
& \rotatebox{-45}{\small HotpotQA} 
& \rotatebox{-45}{\small 2WikiMQA} 
& \rotatebox{-45}{\small Musique} 
& \rotatebox{-45}{\small GovReport} 
& \rotatebox{-45}{\small QMSum} 
& \rotatebox{-45}{\small MultiNews} 
& \rotatebox{-45}{\small TREC} 
& \rotatebox{-45}{\small TriviaQA} 
& \rotatebox{-45}{\small SAMSum} 
& \rotatebox{-45}{\small PCount} 
& \rotatebox{-45}{\small PRe} 
& \rotatebox{-45}{\small Lcc} 
& \rotatebox{-45}{\small RB-P} \\
\midrule
Full Cache       & 24.12 & 31.24 & 39.85 & 45.23 & 34.56 & 21.09 & 28.38 & 23.24 & 26.52 & 74.12 & 90.96 & 42.37 & 4.55 & 71.76 & 58.10 & 51.64 \\
\bottomrule

\multicolumn{17}{c}{\textbf{Masking 16 groups}} \\  
\midrule[0.5pt]

Random
&  20.93
&  28.48
&  33.69
&  44.93
&  20.01
&  20.6
&  28.43
&  23.7
&  26.67
&  74.12
&  91.07
&  41.12
&  4.26
&  71.76
&  49.83
&  40.55\\

HeadKV-R2(top)
&  19.45
&  12.97
&  27.75
&  34.2
&  17.33
&  14.32
&  19.74
&  22.76
&  22.05
&  67.06
&  87.91
&  35.53
&  4.71
&  68.49
&  26.62
&  26.53\\

CoKV(top)
&  \textcolor{blue}{6.55}
&  \textcolor{blue}{9.46}
&  \textcolor{blue}{9.47}
&  \textcolor{blue}{10.19}
&  \textcolor{blue}{12.27}
&  \textcolor{blue}{5.67}
&  \textcolor{blue}{5.73}
&  \textcolor{blue}{16.96}
&  \textcolor{blue}{4.47}
&  \textcolor{blue}{43.53}
&  \textcolor{blue}{71.21}
&  \textcolor{blue}{23.77}
&  \textcolor{blue}{3.91}
&  \textcolor{blue}{34.98}
&  \textcolor{blue}{11.58}
&  \textcolor{blue}{17.18}\\

HeadKV-R2(low)
&  21.83
&  14.36
&  33.34
&  31.37
&  27.23
&  12.55
&  27.29
&  \textcolor{red}{23.82}
&  26.99
&  74.12
&  91.03
&  42.18
&  4.12
&  \textcolor{red}{70.59}
&  37.35
&  38.55\\

CoKV(low)
&  \textcolor{red}{23.74}
&  \textcolor{red}{33.76}
&  \textcolor{red}{41.71}
&  \textcolor{red}{49.27}
&  \textcolor{red}{40.48}
&  \textcolor{red}{19.99}
&  \textcolor{red}{29.13}
&  23.25
&  \textcolor{red}{27.79}
&  \textcolor{red}{74.12}
&  \textcolor{red}{91.45}
&  \textcolor{red}{42.37}
&  \textcolor{red}{4.71}
&  70.55
&  \textcolor{red}{63.38}
&  \textcolor{red}{61.26}\\
\bottomrule

\multicolumn{17}{c}{\textbf{Masking 128 groups}} \\  
\midrule[0.5pt]

Random           
&  3.34
&  2.50
&  5.33
&  10.59
&  5.12
&  2.73
&  2.15
&  9.19
&  0.16
&  44.12
&  31.33
&  9.05
&  4.18
&  66.74
&  12.27
&  9.23\\
HeadKV-R2(top)           
&  2.34
&  2.17
&  5.38
&  7.21
&  7.19
&  1.85
&  1.80
&  10.34
&  0.31
&  34.71
&  26.08
&  7.87
&  4.71
&  66.92
&  13.94
&  11.76\\
CoKV(top)             
&  \textcolor{blue}{0.59}
&  \textcolor{blue}{0.80}
&  \textcolor{blue}{1.38}
&  \textcolor{blue}{2.96}
&  \textcolor{blue}{3.42}
&  \textcolor{blue}{1.11}
&  \textcolor{blue}{1.16}
&  \textcolor{blue}{4.05}
&  \textcolor{blue}{0.13}
&  \textcolor{blue}{34.12}
&  \textcolor{blue}{2.89}
&  \textcolor{blue}{7.17}
&  \textcolor{blue}{1.09}
&  \textcolor{blue}{7.52}
&  \textcolor{blue}{2.91}
&  \textcolor{blue}{3.55}\\

HeadKV-R2(low) 
&  12.02
&  7.97
&  8.92
&  14.87
&  12.83
&  5.26
&  2.41
&  9.12
&  1.42
&  55.88
&  40.96
&  10.2
&  4.71
&  \textcolor{red}{68.42}
&  10.14
&  6.03\\

CoKV(low) 
&  \textcolor{red}{15.31}
&  \textcolor{red}{12.15}
&  \textcolor{red}{28.44}
&  \textcolor{red}{35.35}
&  \textcolor{red}{23.27}
&  \textcolor{red}{10.67}
&  \textcolor{red}{2.93}
&  \textcolor{red}{12.24}
&  \textcolor{red}{9.41}
&  \textcolor{red}{73.82}
&  \textcolor{red}{76.32}
&  \textcolor{red}{37.70}
&  \textcolor{red}{4.71}
&  68.24
&  \textcolor{red}{22.20}
&  \textcolor{red}{24.93}\\

\bottomrule

\end{tabular}
\vspace{0.5em}
\footnotesize
\end{adjustbox}
\end{table*}
The results show that when masking the top-ranked groups identified by each method, the performance of CoKV degrades more significantly than that of HeadKV-R2. Conversely, when masking the unimportant groups (\textbf{low}), the performance of CoKV declines more gradually than HeadKV-R2. This suggests that CoKV is more effective at ranking group importance, as it better distinguishes between critical and non-critical caches. The results of masking 16 groups in both models outperformed the FullKV approach as shown in Figure~\ref{fig:mask_llama_mistral_average}. 
This further demonstrates that CoKV can identify groups that have a negative impact on the model. By removing the KV pairs from these groups, the model inference not only optimizes storage and decoding speed but also enhances overall performance.

\subsection{Decoding Latency and Memory Usage}
We conduct experiments using the Mistral-7B-Instruct-v0.2 model, which supports a maximum context window of 32k tokens, with FlashAttention enabled as the default setting, on an A100 GPU with 40GB of memory. We design two key experiments with the average KV cache size set to 128 tokens(comparative experiments showed less than 2\% variation across 64/256/512/1024 tokens).

\begin{figure}[h]
    \centering
    \includegraphics[width=0.4\textwidth]{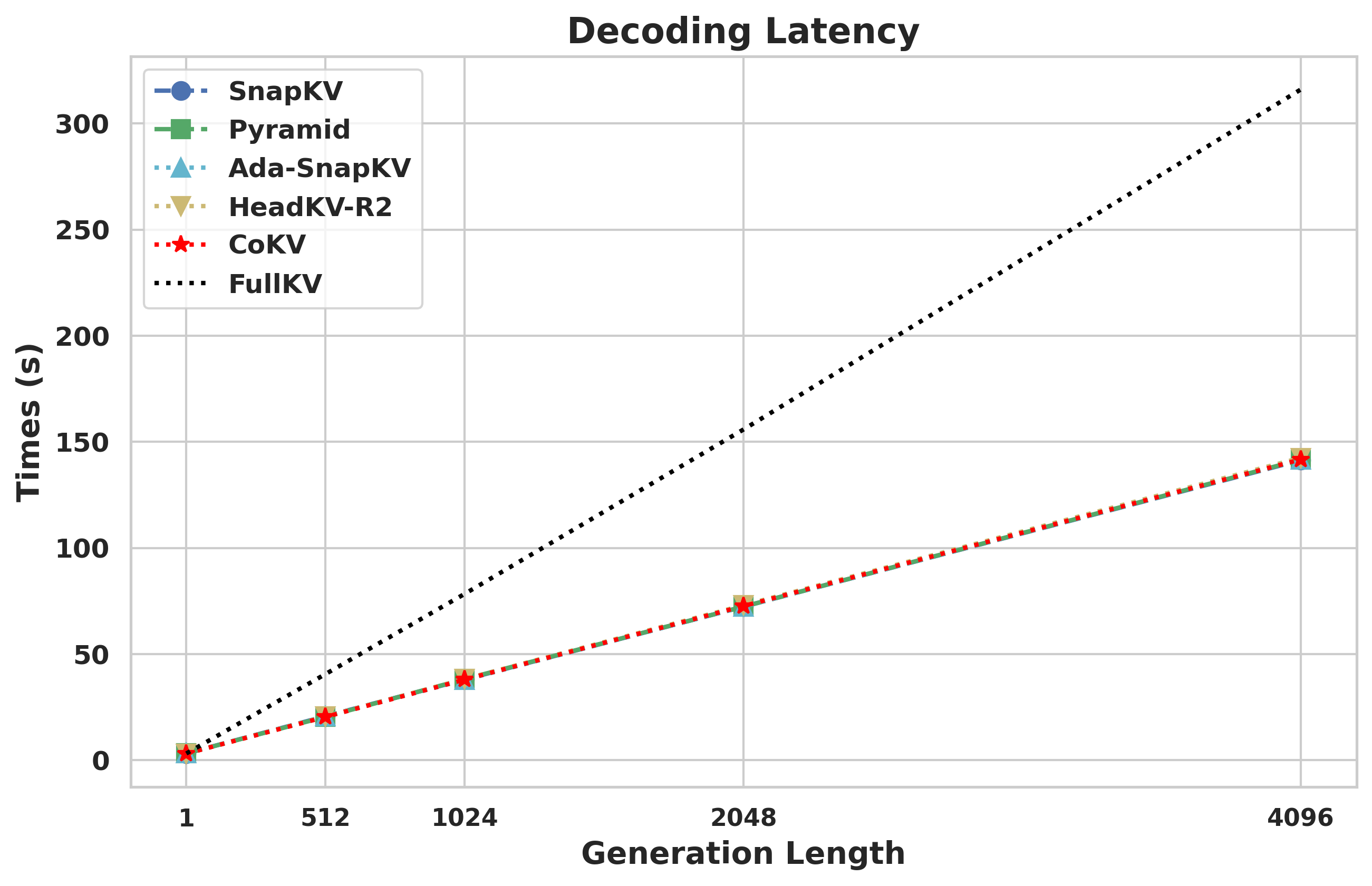} \\[1em] 
    \includegraphics[width=0.4\textwidth]{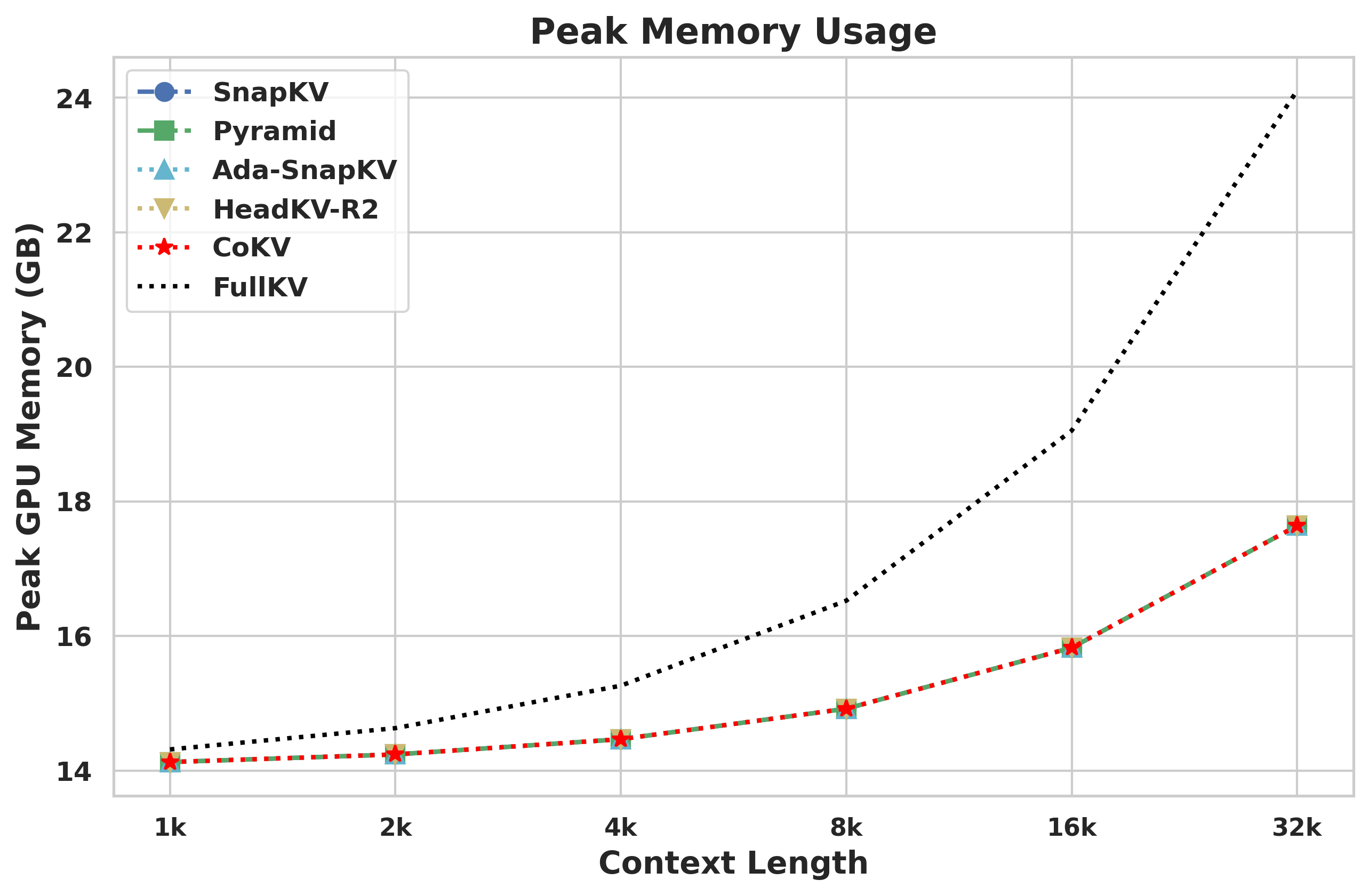} 
    \caption{Results of Decoding Latency and Peak Memory Usage, demonstrating that CoKV maintains comparable performance with other baseline methods while achieving significant improvements over FullKV.}
    \label{fig:decoding}
\end{figure}

\paragraph{Decoding Latency} With a fixed input context length of 28k tokens, we measure decoding latency (including both the pre-filling time and the decoding time)  across different generation lengths (1/512/1024/2048/4096 tokens). As shown in the Decoding Latency of Figure~\ref{fig:decoding},  CoKV achieves less than 50\% of the total latency compared to the FullKV baseline, with negligible differences observed between the other baselines.

\paragraph{Peak Memory Usage} Under fixed generation length (1 token), we measure the peak GPU memory usage (including model parameters and runtime states) across varying input contexts (1k/2k/4k/8k/16k/32k tokens). As shown in the Peak Memory Usage of Figure~\ref{fig:decoding}, CoKV reduces memory usage by 64\% compared to FullKV baseline at 32k input length.

\section{Conclusion}
Large language models (LLMs) face significant challenges in handling long texts due to the excessive memory and latency overhead caused by the growing size of the KV cache. To this end, we introduce the Sliced Shapley value (SSV) to evaluate the collaborative importance of attention heads and a novel method called  CoKV to dynamically allocate cache sizes based on SSV. Our experimental results demonstrate that CoKV achieves state-of-the-art performance across 16 LongBench datasets, outperforming the full KV cache in 9 datasets while reducing memory and latency overhead. CoKV provides a scalable and practical solution for enhancing the efficiency of LLMs in real-world applications.

\section*{Limitations}
Our work has two main limitations that suggest future research directions:

\paragraph{Task-specific constraint:}CoKV requires calculating head importance scores for different tasks. While experiments in Appendix Section~\ref{section:generalization} demonstrate strong generalizability across datasets within the same task category. Despite this constraint, CoKV is highly practical for LLM providers serving diverse users. Users can simply select their task type, and the model will apply the corresponding head importance scores for KV cache compression. Importantly, the underlying inference process remains consistent across all tasks; only the cache budget allocation varies based on the task-specific importance scores. This ensures both flexibility and efficiency, enabling the model to adapt to various user needs without requiring significant changes to its core architecture.

\paragraph{Precomputation cost:}The computation of importance based on cooperative game theory for attention heads is computationally intensive. Although we propose the Sliced Shapley Value (SSV), which significantly reduces the computational cost, our precomputation overhead remains higher than that of baseline methods. However, our experiments in Appendix Section~\ref{section:compute_efficiency} demonstrate that this precomputation is still entirely acceptable. We plan to address optimizing computational complexity as one of our future research directions by developing efficient approximation algorithms and parallel computing strategies.

\bibliography{custom}

\clearpage

\appendix

\section*{Appendix}

\label{sec:appendix}

\section{Related Works}

\paragraph{KV Cache Compression} The memory overhead of storing key-value (KV) pairs for LLM has motivated extensive research on KV cache compression. StreamingLLM ~\cite{StreamingLLM} preserves the initial and recent tokens, which empirically exhibit higher informativeness during generation. Similarly, Scissorhands ~\cite{scissorhands} proposes the persistence of importance to identify and retain pivotal tokens. H2O ~\cite{H2O} employs a heavy-hitter oracle to drop tokens with low attention scores.  SnapKV ~\cite{SnapKV} uses the attention scores of the recent tokens to retain critical tokens. While these methods reduce memory usage and accelerate inference, they implicitly assume uniform importance across attention heads, limiting their applicability.  Recent works address head diversity through layer-wise and head-wise optimizations. PyramidKV~\cite{PyramidKV} implements a hierarchical allocation strategy, assigning larger cache budgets to lower layers based on the observed attention patterns across layers. FastGen~\cite{FastGen} is an adaptive KV cache compression method that reduces LLMs' memory usage by profiling attention modules and constructing caches adaptively. RazorAttention ~\cite{Razorattention} and Duoattention~\cite{Duoattention} divide attention heads into retrieval heads(critical for long-context processing~\cite{retrievalheads}) and non-retrieval heads, apply full KV cache to retrieval heads and compressed KV cache to non-retrieval heads. ArkVale~\cite{chen2025arkvale} proposes a page-based KV cache manager that asynchronously copies filled pages into external memory (e.g., CPU memory) as a backup and supports the recall of important tokens that were previously evicted. AdaKV ~\cite{AdaKV} dynamically adjusts cache budgets across heads based on their concentration degrees and HeadKV ~\cite{HeadKV} calculates head importance scores to allocate cache budget before inference.
However, these methods assess heads in isolation, neglecting their collaborative interactions. For example, the standalone score of a head may not reflect its true contribution when working synergistically with others. Additionally, these approaches overlook the task-dependent variations in head importance. 
Our approach tackles these limitations by modeling head interactions as a cooperative game, dynamically allocating cache resources based on the varying complementary contributions of heads across different tasks.

In addition to KV cache eviction methods, KV cache quantization is also one of the mainstream approaches for KV cache compression~\cite{yang2024no,liukivi}. However, while eviction methods can be used to retain less than 1\% of the cache, KV cache compression cannot be applied to such an extent because it must preserve at least 1 bit. Nevertheless, the combination of these two methods is an interesting direction for future research.

\paragraph{Model Architecture and Computation Optimization}
Modern LLMs employ architectural optimizations to balance efficiency and performance. Multi Query Attention (MQA) ~\cite{mqa} shares a single key-value pair across all attention heads, drastically reducing memory usage but potentially sacrificing performance. Group Query Attention (GQA) ~\cite{gqa} strikes a balance by grouping heads to share key-value pairs, preserving performance while maintaining memory efficiency, which is widely adopted in recent LLMs like Llama~\cite{llama3} and Mistral~\cite{mistral}. Concurrently, Flash Attention ~\cite{flashattention} optimizes hardware utilization by minimizing memory reads/writes during attention computation, significantly accelerating inference. Notably, our approach is fully compatible with GQA and Flash Attention, ensuring seamless integration with current LLMs.

\paragraph{Cooperative Game Theory } Cooperative game theory offers a principled framework for understanding how multiple players can jointly contribute to overall system performance. Shapley value ~\cite{shapley}, a classic solution in cooperative game theory, provides a method for fairly allocating joint benefits based on the marginal contribution of each player. However, traditional Shapley value computation methods allow each sample to be used to calculate the marginal contribution of only a single player.
Recent works ~\cite{cc,sun2024shapley} address this limitation through complementary contributions that enable simultaneous estimation of multiple players' contributions. In the context of LLMs, these methods still encounter scalability issues, as the cost of computing complementary contributions across all coalition sizes remains prohibitively high.
We propose the Sliced Shapley value, which samples only a subset of coalition sizes. This approach not only accelerates the computation but also accurately reflects the contributions of different heads.
\section{Supplementary experiments}

We introduce the detailed information of LongBench in Table~\ref{tab:longbench_metric}, including the task types, evaluation metrics, average length, languages, and the number of samples for each task.
\begin{table*}[h]
\centering
\caption{Details of 16 Datasets in LongBench} \label{tab:longbench_metric}
\begin{adjustbox}{width=0.85\textwidth}
\small {
\begin{tabular}{lllClC}
\toprule
Label  & Task Type & Eval metric & Avg len & Language & Sample Num \\ \midrule
NrtvQA  & Single-Doc. QA & F1 & 18,409 & EN & 200 \\
Qasper  & Single-Doc. QA & F1 & 3,619 & EN & 200 \\
MF-en  & Single-Doc. QA & F1 & 4,559 & EN & 150 \\
HotpotQA  & Multi-Doc. QA  & F1 & 9,151 & EN & 200 \\
2WikiMQA  & Multi-Doc. QA  & F1 & 4,887 & EN & 200 \\
Musique  & Multi-Doc. QA  & F1 & 11,214 & EN & 200 \\
GovReport & Summarization & Rouge-L & 8,734 & EN & 200 \\
QMSum  & Summarization & Rouge-L & 10,614 & EN & 200 \\
MultiNews  & Summarization & Rouge-L & 2,113 & EN & 200 \\
TREC  & Few-shot Learning & Accuracy & 5,177 & EN & 200 \\
TriviaQA  & Few-shot Learning & F1 & 8,209 & EN & 200 \\
SAMSum  & Few-shot Learning & Rouge-L & 6,258 & EN & 200 \\
PCount & Synthetic  & Accuracy & 11,141 & EN & 200 \\
PRe  & Synthetic  & Accuracy & 9,289 & EN & 200 \\
Lcc  & Code & Edit Sim & 1,235 & \makecell{Python/\\C\#/Java} & 500 \\
RB-P  & Code & Edit Sim & 4,206 & \makecell{Python/\\Java} & 500 \\ \bottomrule
\end{tabular}
}
\end{adjustbox}
\vspace{0.5em}
\end{table*}
.

\subsection{Computation Efficiency}
\label{section:compute_efficiency}
We conduct experiments to demonstrate the efficiency of approximating the Sliced Shapley value using the \emph{qasper} dataset with the Llama-3-8B-Instruct model. We randomly select 15\% of the \emph{qasper} dataset as the validation set to compute the Sliced Shapley value. The experiments are performed on a server equipped with 8 RTX 3090 GPUs. We compute the Sliced Shapley value for coalition sizes of $\{32, 64, 96, 128\}$. GPUs 0-3 are assigned to compute the complementary contributions for coalitions of sizes $\{32, 64, 96, 128\}$, respectively, while GPUs 4-7 compute another independent Sliced Shapley value. Table~\ref{table:qasper_mse} shows the computation time for each GPU from 50 to 500 samples of complementary contributions, as well as the mean absolute error (MAE) between the two independently computed Sliced Shapley values. The MAE is calculated as:
\[
MAE = \frac{\sum_{i=1}^n |\overline{\mathcal{SSV}}_i^{\mathcal{H}} - \overline{\mathcal{SSV}}_i^{\mathcal{H}'}|}{n},
\]
where $\overline{\mathcal{SSV}}_i^{\mathcal{H}}$ and $\overline{\mathcal{SSV}}_i^{\mathcal{H}'}$ represent the Sliced Shapley values from the two independent computations. The experimental results show that when the number of samples reaches 250 for each coalition size, the MAE is $3.8e-3 \leq 1/256$ with 20.93 hours. In GQA inference, the Llama-3-8B-Instruct model has a total of $32 \times 8 = 256$ groups. Since the model accuracy lies in the range $[0, 1]$, when the MAE between two sampling runs is less than $1/256$, the sum of absolute errors across all groups is less than 1. At this point, the Sliced Shapley value can reliably reflect the contributions of the groups. 

We recommend performing two independent sampling runs when computing the Sliced Shapley value for a task. The sampling results are considered stable when the mean absolute error between the two runs is less than $1/n$, where $n$ represents the number of players in the cooperative game. At this point, the results from the two sampling runs can be averaged and used as the importance scores of the heads in the model.

\subsection{Distribution of Sliced Shapley Value}
\begin{figure*}[t]
    \centering
    \subfigure{
		\begin{minipage}[b]{0.19\textwidth}
			\includegraphics[width=1\textwidth]{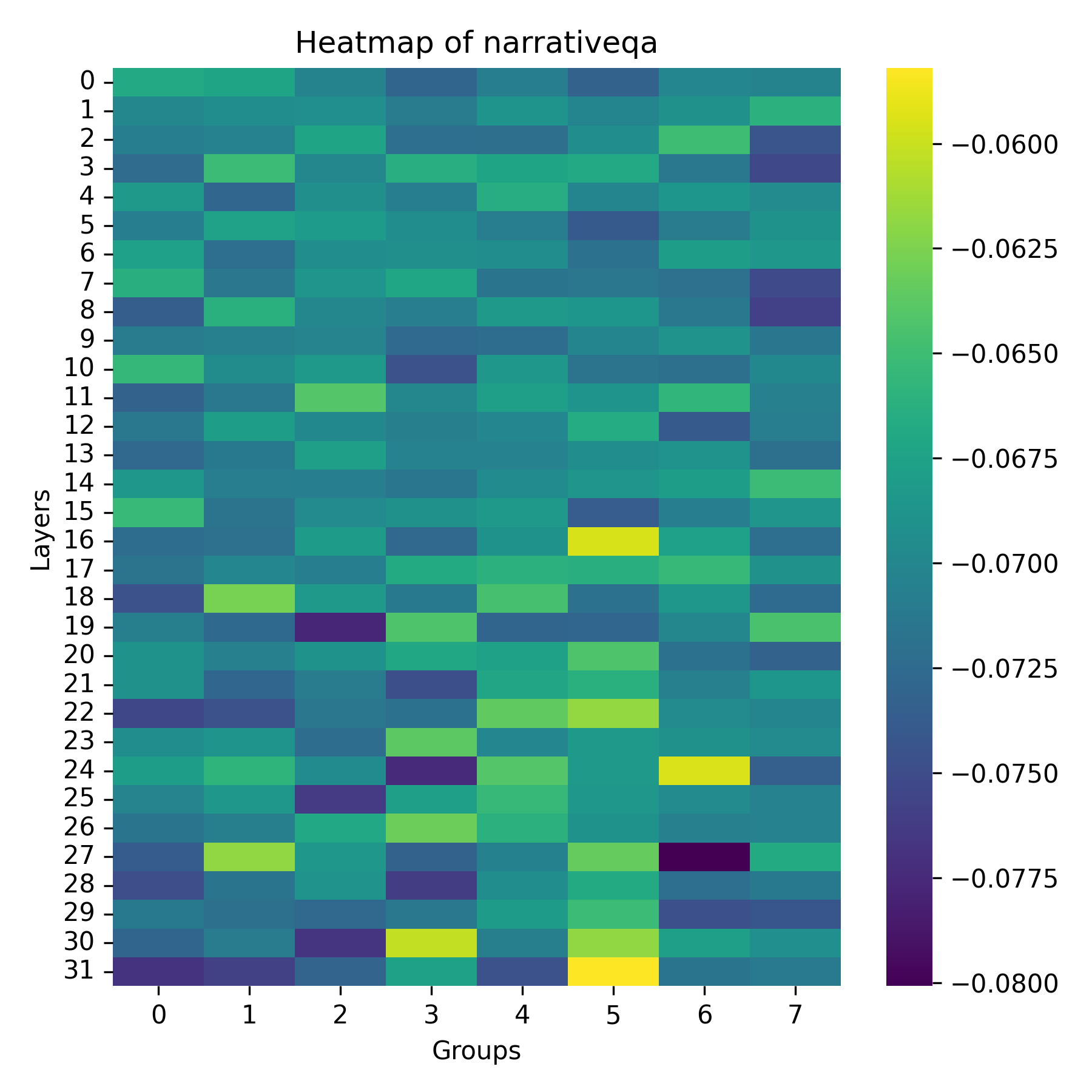}
			\caption* {(1) NtrQA}
			\label{cifar:longtail}
		\end{minipage}%
	}
	\subfigure{
		\begin{minipage}[b]{0.19\textwidth}
			\includegraphics[width=1\textwidth]{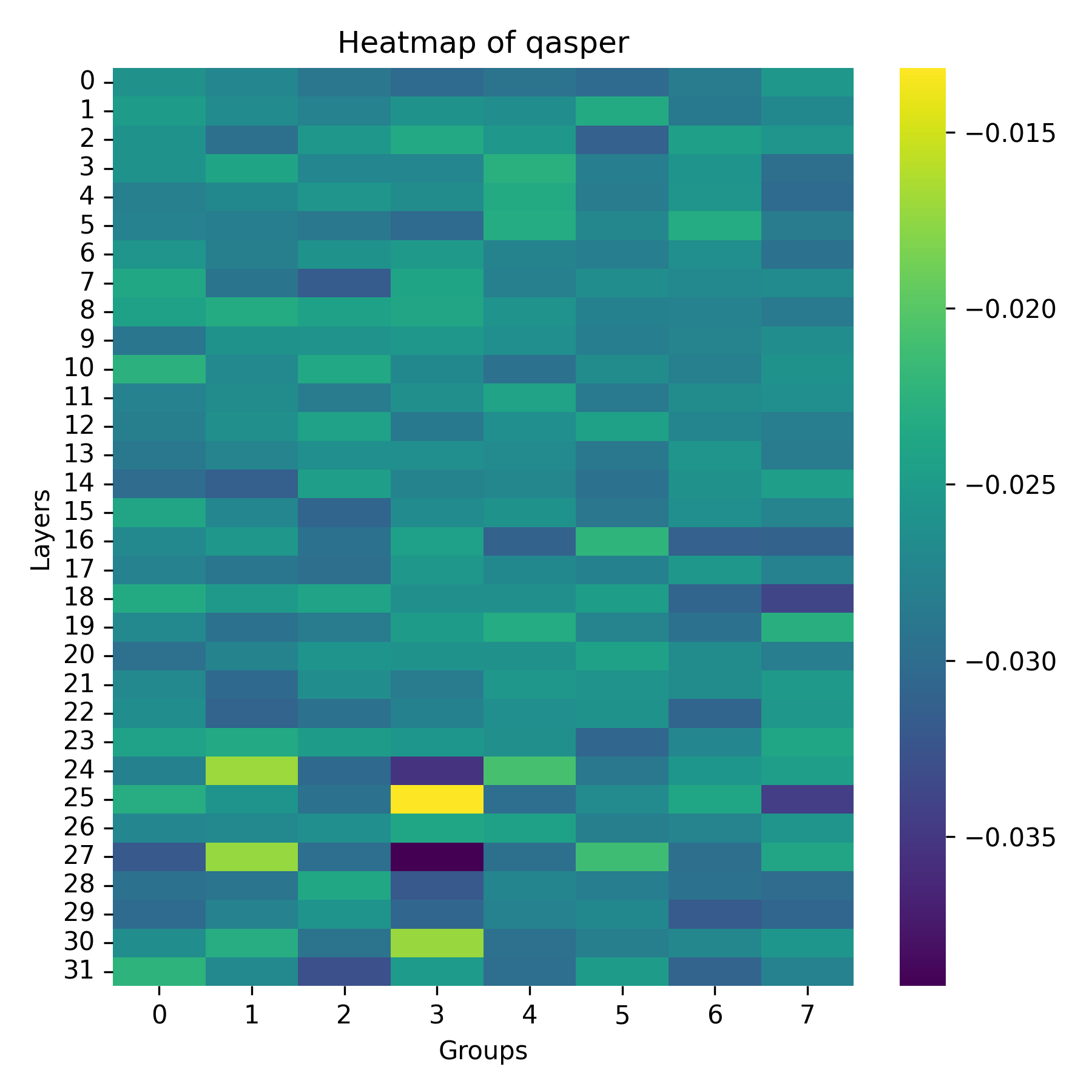}
			\caption* {(2) Qasper}
			\label{cifar:opensetnoise}
		\end{minipage}%
	}%
    \subfigure{
		\begin{minipage}[b]{0.19\textwidth}
			\includegraphics[width=1\textwidth]{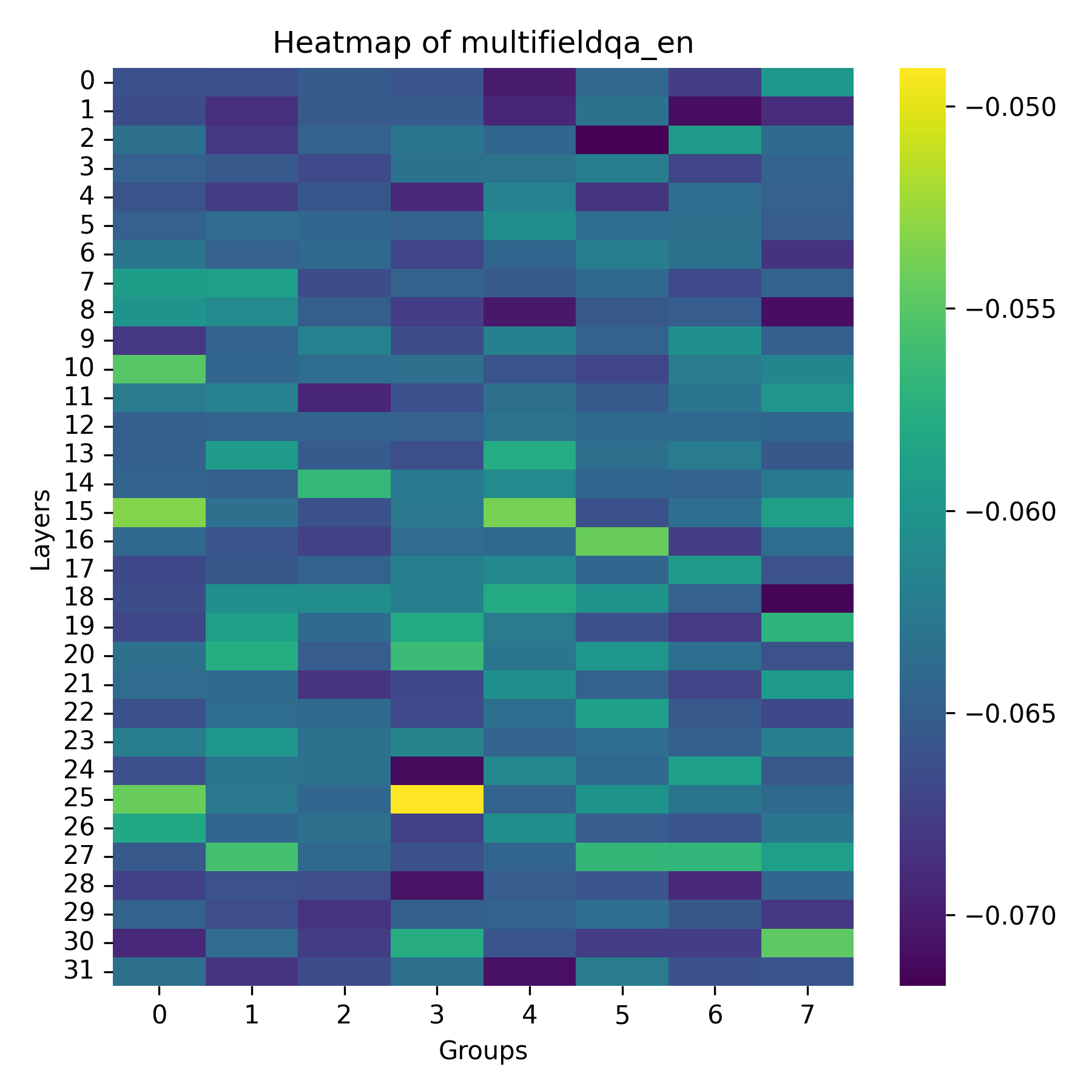}
			\caption* {(3) MF-en}
			\label{cifar:closedsetnoise}
		\end{minipage}%
	}%
    \subfigure{
    	\begin{minipage}[b]{0.19\textwidth}
    		\includegraphics[width=1\textwidth]{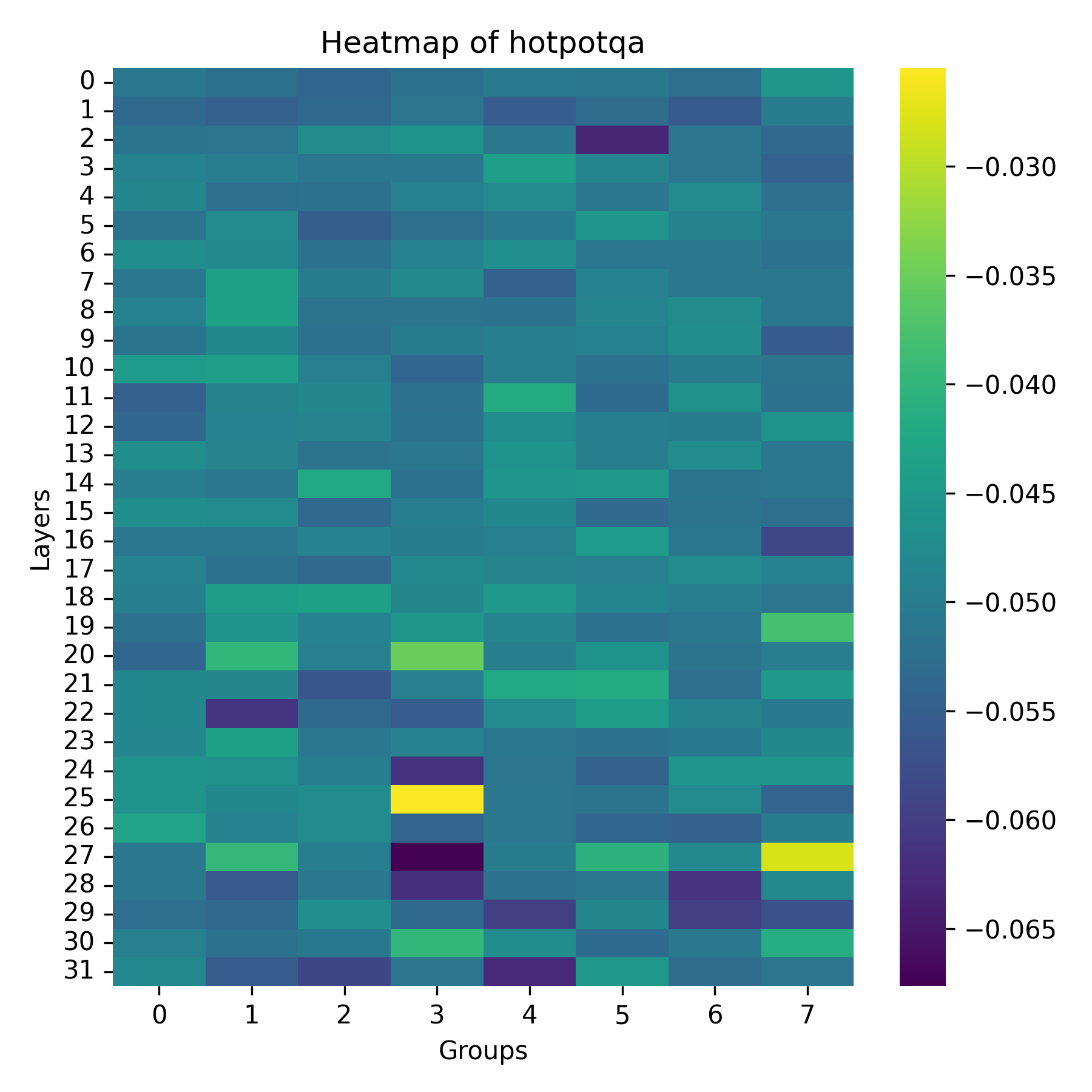}
    		\caption* {(4) HotpotQA}
    		\label{cifar:datanoise}
    	\end{minipage}%
    }%

    \subfigure{
		\begin{minipage}[b]{0.19\textwidth}
			\includegraphics[width=1\textwidth]{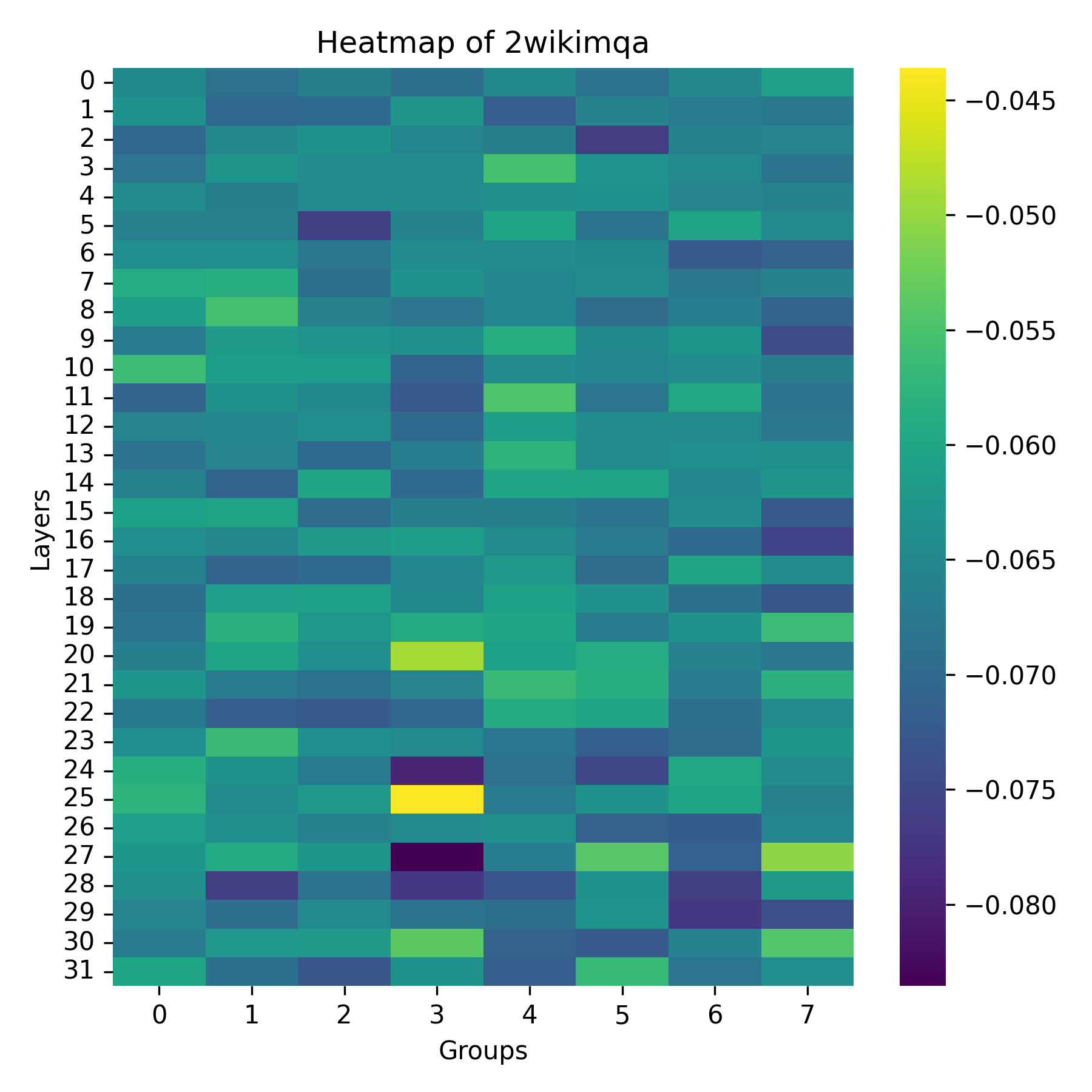}
			\caption* {(5) 2WikiMQA}
			\label{cifar:longtail}
		\end{minipage}%
	}
	\subfigure{
		\begin{minipage}[b]{0.19\textwidth}
			\includegraphics[width=1\textwidth]{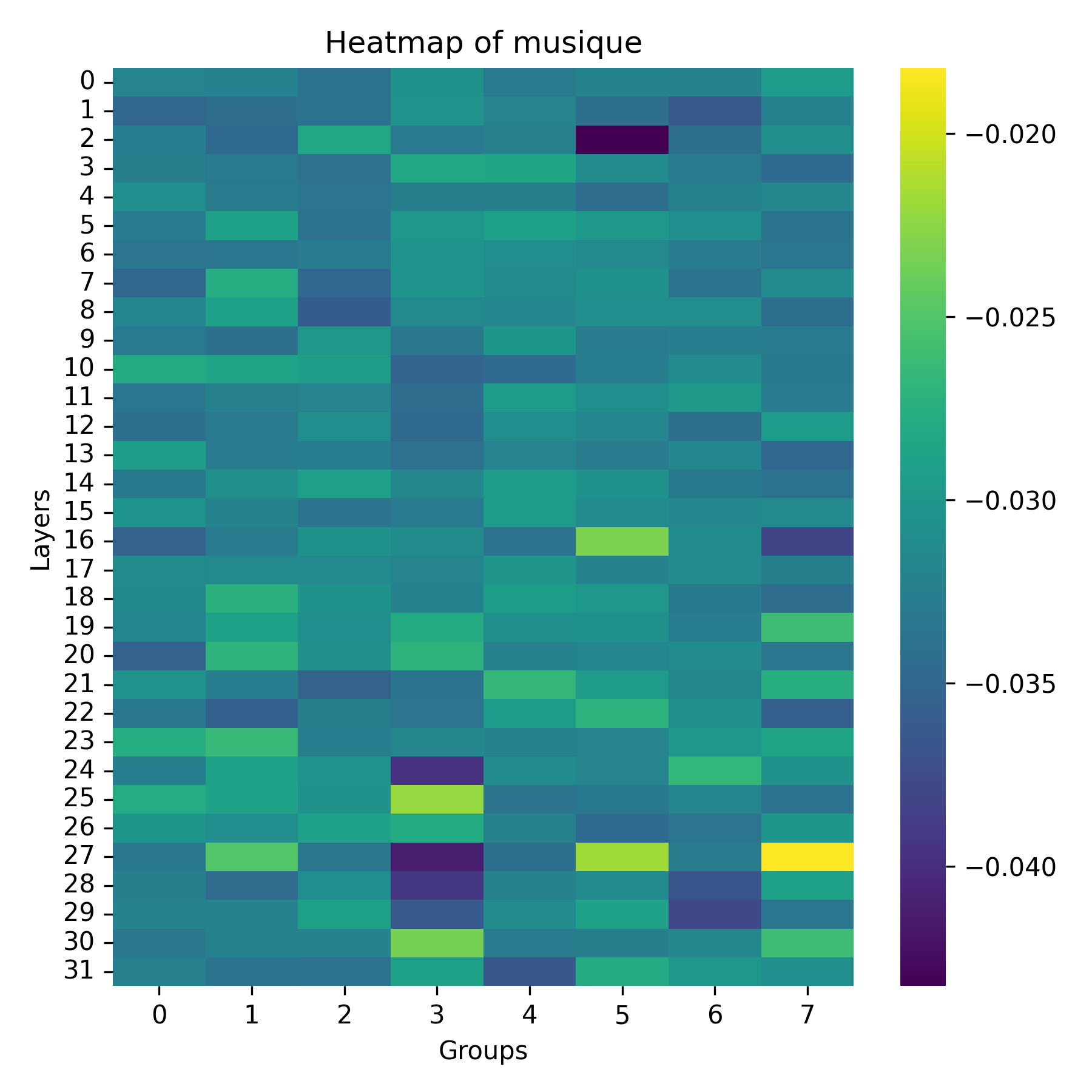}
			\caption* {(6) Musique}
			\label{cifar:opensetnoise}
		\end{minipage}%
	}%
    \subfigure{
		\begin{minipage}[b]{0.19\textwidth}
			\includegraphics[width=1\textwidth]{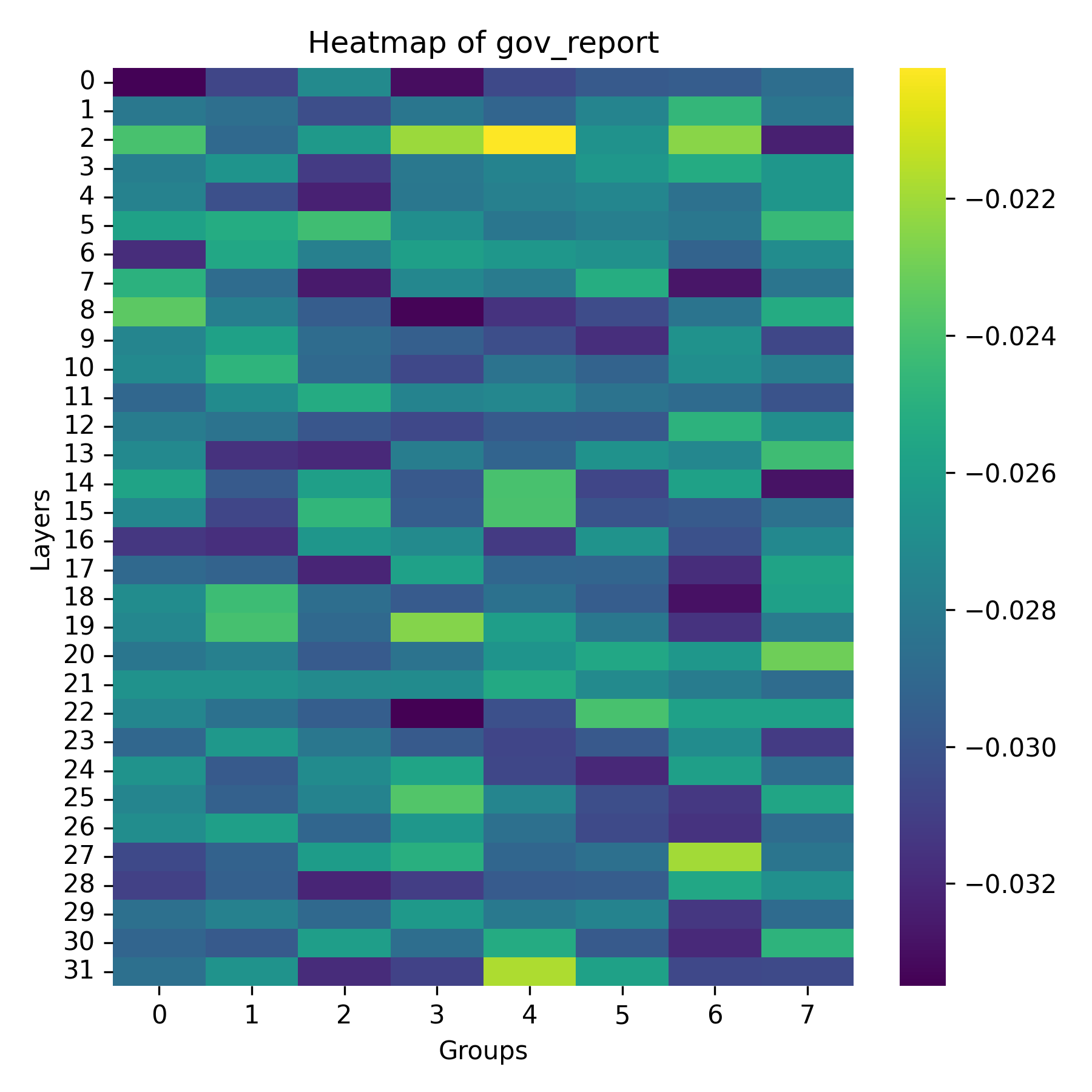}
			\caption* {(7) GovReport}
			\label{cifar:closedsetnoise}
		\end{minipage}%
	}%
    \subfigure{
    	\begin{minipage}[b]{0.19\textwidth}
    		\includegraphics[width=1\textwidth]{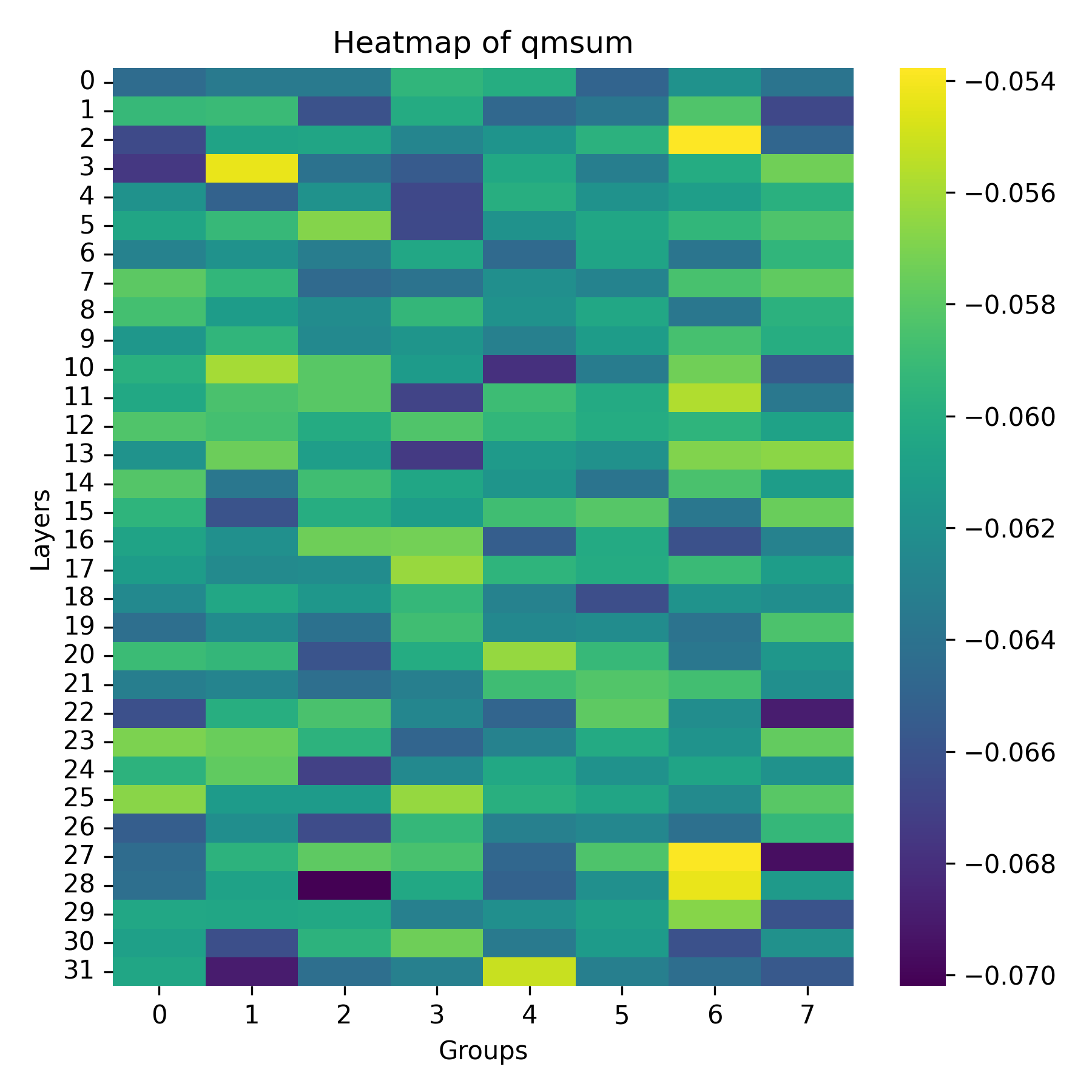}
    		\caption* {(8) QMSum}
    		\label{cifar:datanoise}
    	\end{minipage}%
    }%

    \subfigure{
		\begin{minipage}[b]{0.19\textwidth}
			\includegraphics[width=1\textwidth]{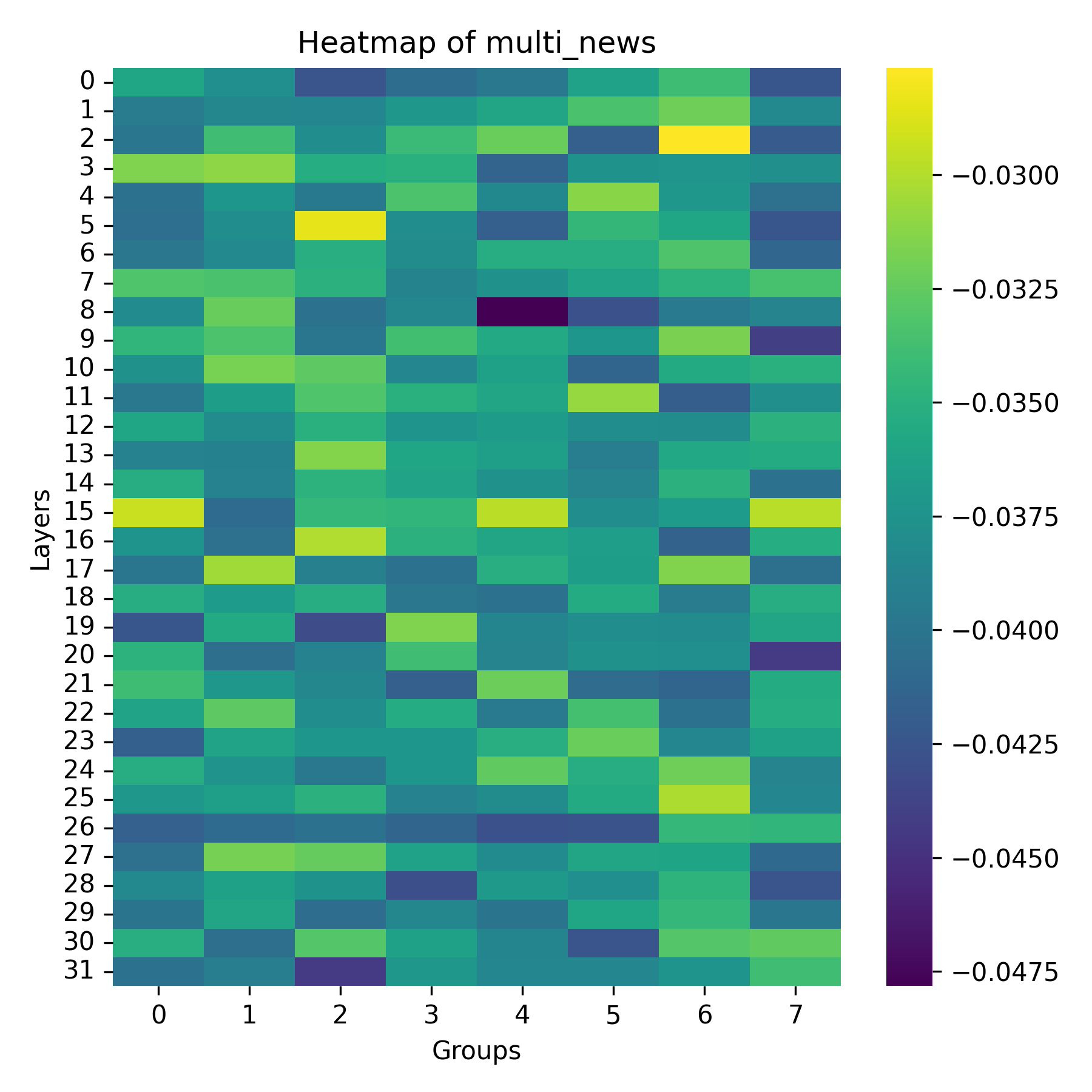}
			\caption* {(9) MultiNews}
			\label{cifar:longtail}
		\end{minipage}%
	}
	\subfigure{
		\begin{minipage}[b]{0.19\textwidth}
			\includegraphics[width=1\textwidth]{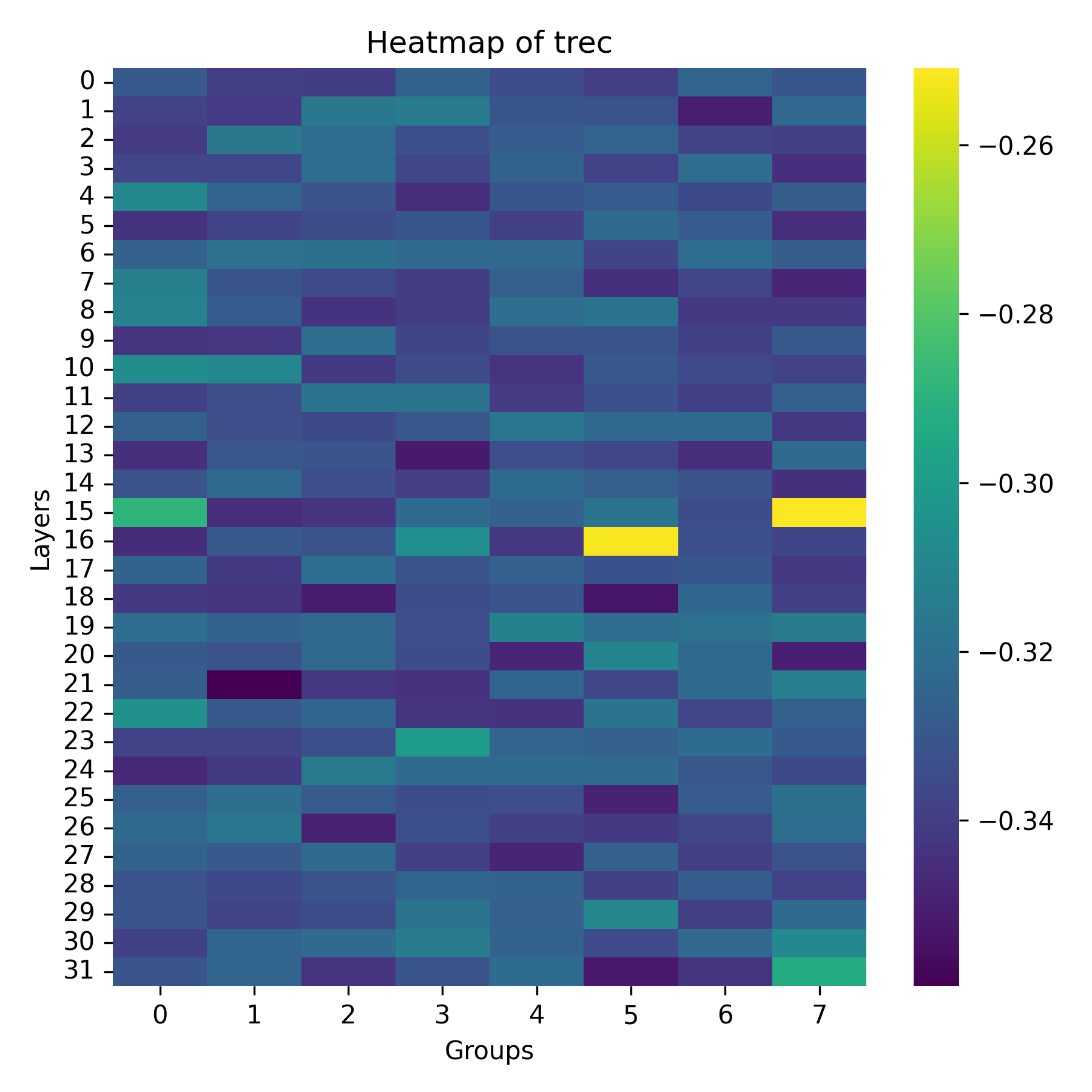}
			\caption* {(10) TREC}
			\label{cifar:opensetnoise}
		\end{minipage}%
	}%
    \subfigure{
		\begin{minipage}[b]{0.19\textwidth}
			\includegraphics[width=1\textwidth]{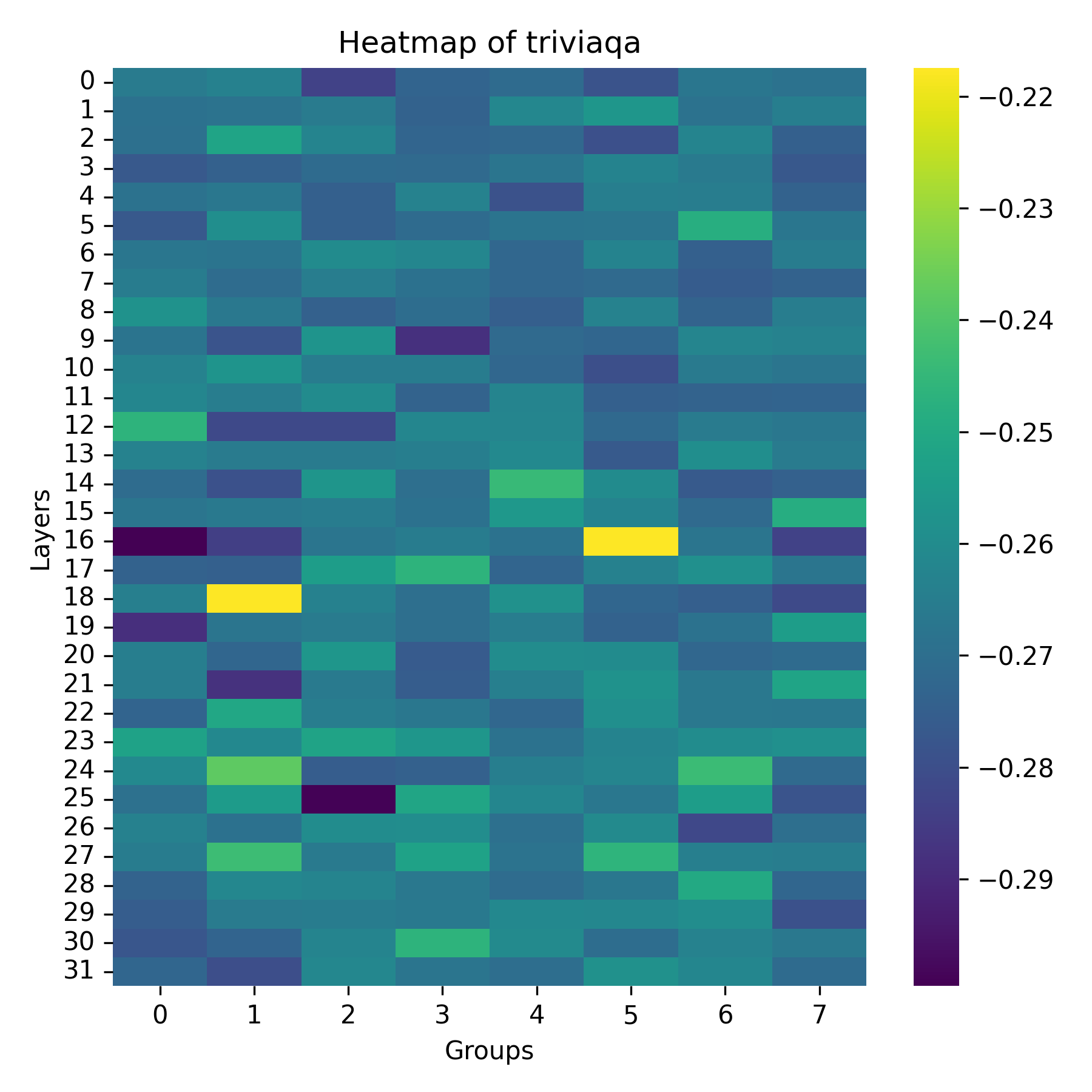}
			\caption* {(11) TriviaQA}
			\label{cifar:closedsetnoise}
		\end{minipage}%
	}%
    \subfigure{
    	\begin{minipage}[b]{0.19\textwidth}
    		\includegraphics[width=1\textwidth]{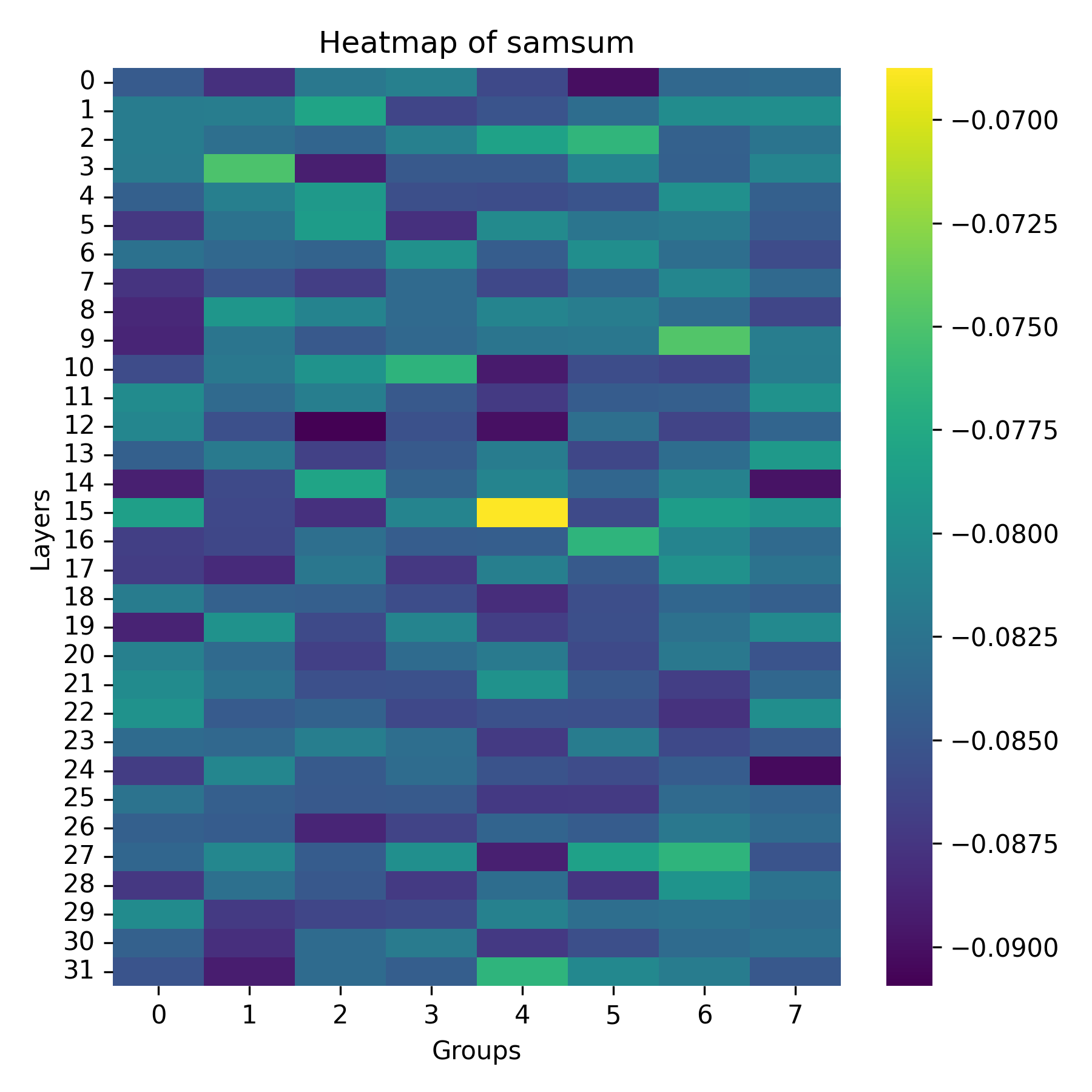}
    		\caption* {(12) SAMSum}
    		\label{cifar:datanoise}
    	\end{minipage}%
    }%

    \subfigure{
		\begin{minipage}[b]{0.19\textwidth}
			\includegraphics[width=1\textwidth]{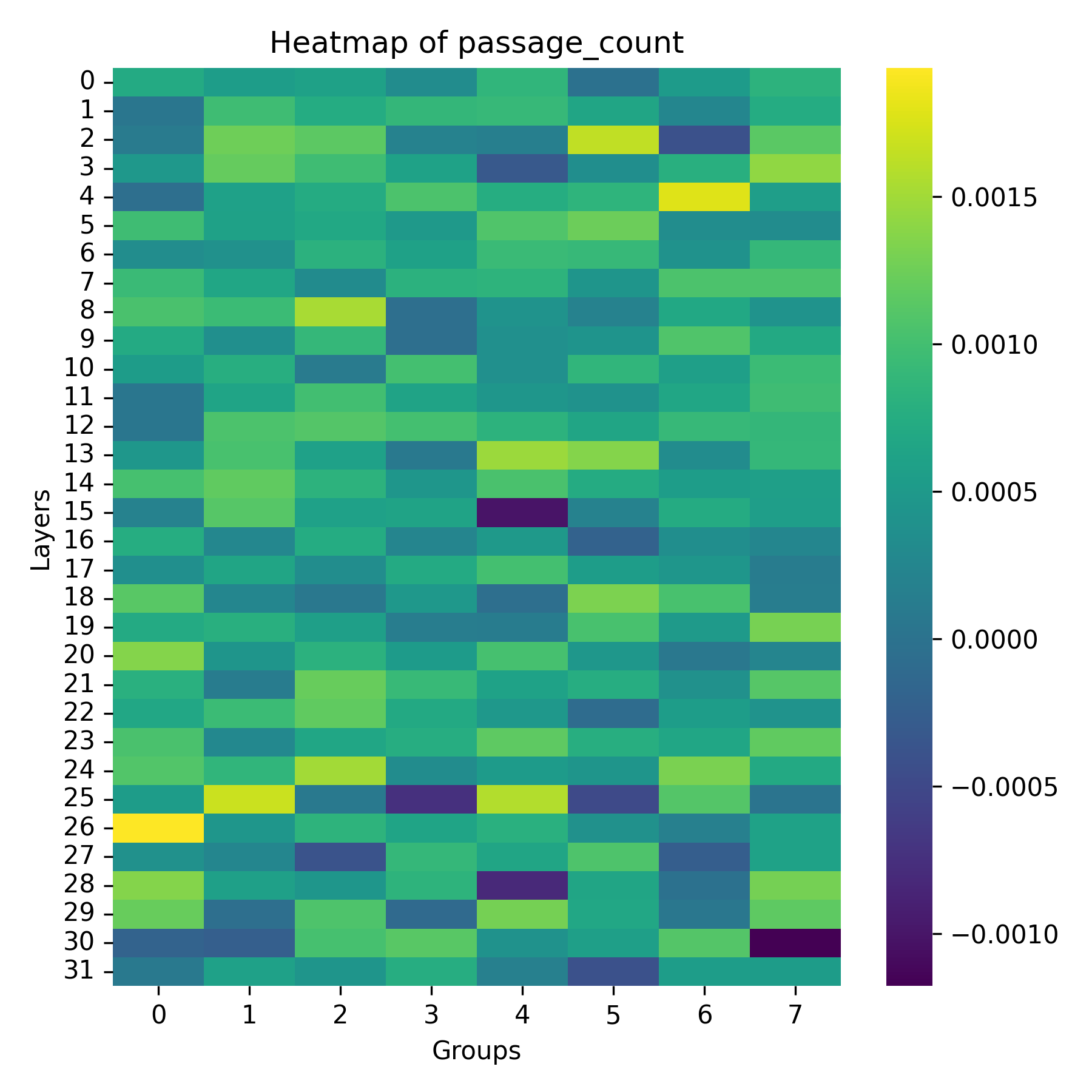}
			\caption* {(13) PCount}
			\label{cifar:longtail}
		\end{minipage}%
	}
	\subfigure{
		\begin{minipage}[b]{0.19\textwidth}
			\includegraphics[width=1\textwidth]{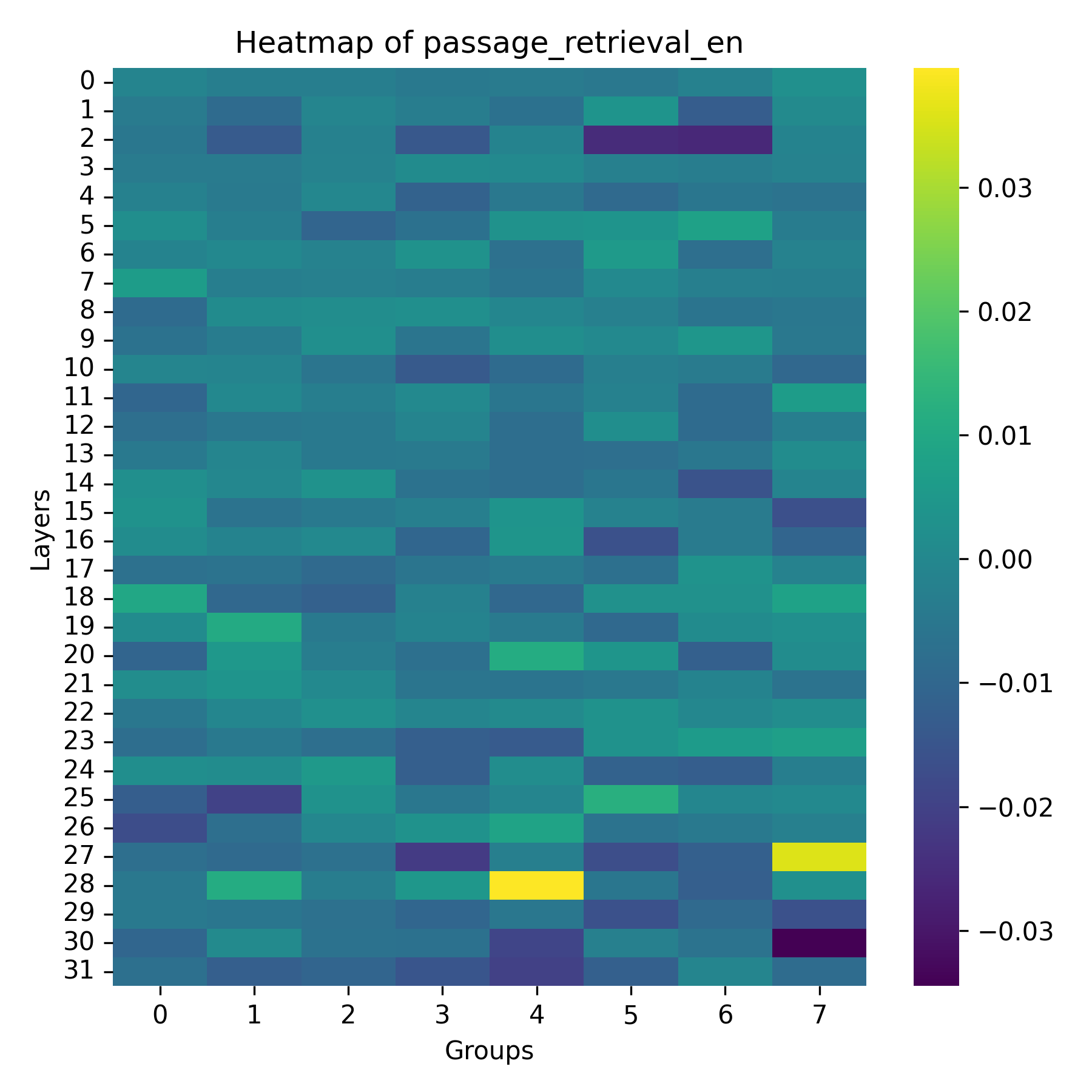}
            \caption* {(14) PRe}
			\label{cifar:opensetnoise}
		\end{minipage}%
	}%
    \subfigure{
		\begin{minipage}[b]{0.19\textwidth}
			\includegraphics[width=1\textwidth]{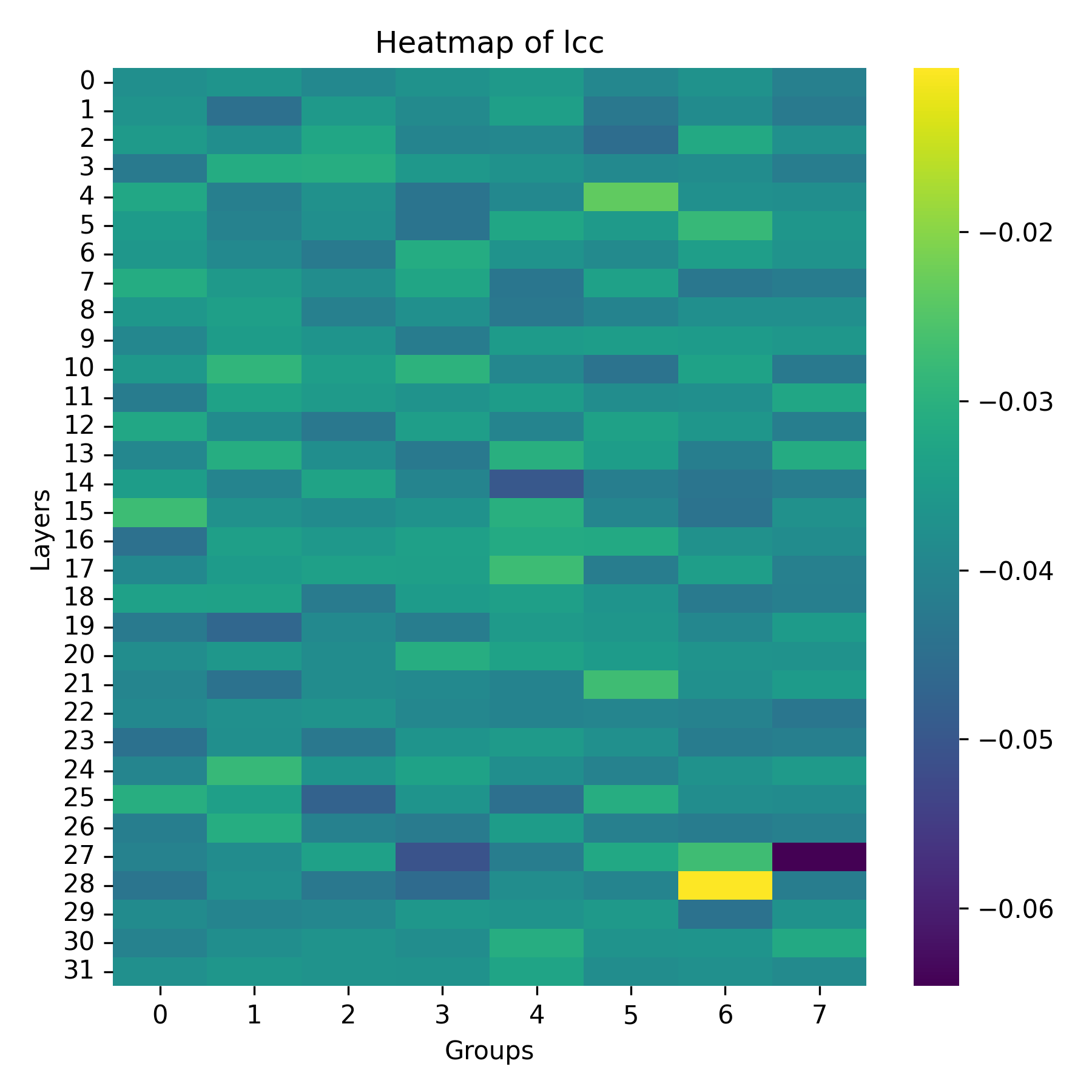}
			\caption* {(15) Lcc}
			\label{cifar:closedsetnoise}
		\end{minipage}%
	}%
    \subfigure{
    	\begin{minipage}[b]{0.19\textwidth}
    		\includegraphics[width=1\textwidth]{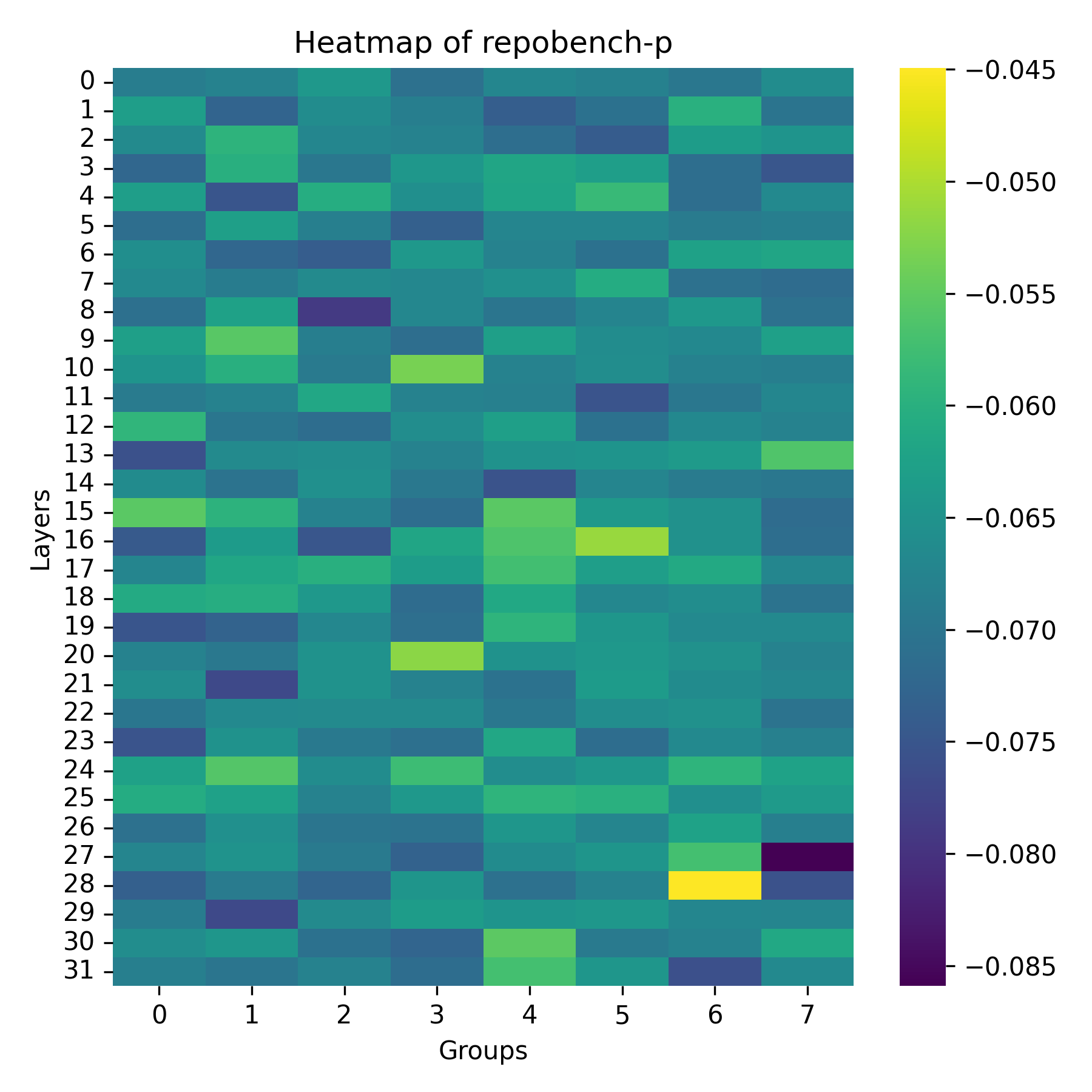}
    		\caption* {(16) RB-P}
    		\label{cifar:datanoise}
    	\end{minipage}%
    }%
    
\caption {Heatmap of Llama-3-8B-Instruct. }
\label{fig:heatmap_llama}
\end{figure*}

\begin{figure*}[t]
    \centering
    \subfigure{
		\begin{minipage}[b]{0.19\textwidth}
			\includegraphics[width=1\textwidth]{pics/cc_pictures/llama_narrativeqa.png}
			\caption* {(1) NtrQA}
			\label{cifar:longtail}
		\end{minipage}%
	}
	\subfigure{
		\begin{minipage}[b]{0.19\textwidth}
			\includegraphics[width=1\textwidth]{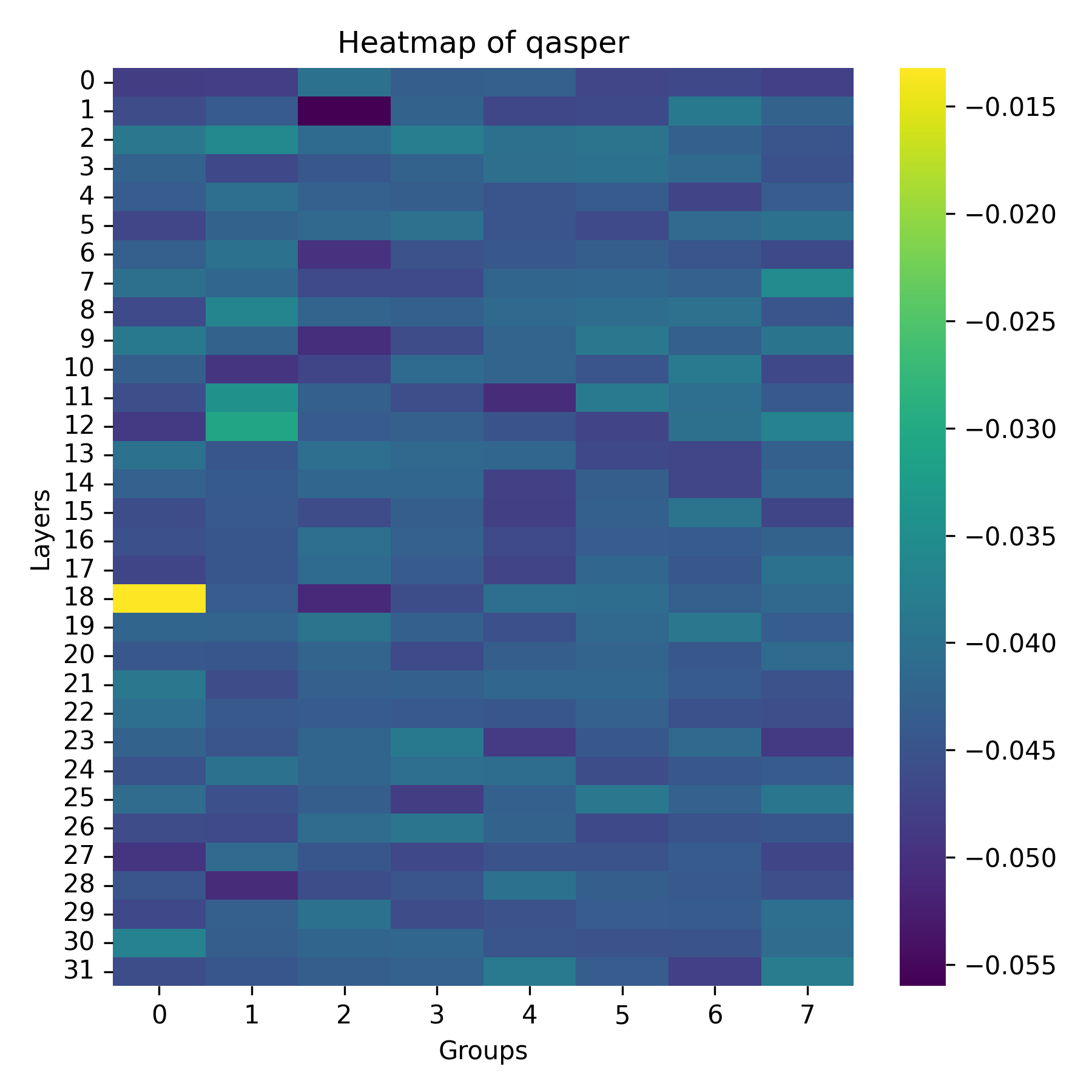}
			\caption* {(2) Qasper}
			\label{cifar:opensetnoise}
		\end{minipage}%
	}%
    \subfigure{
		\begin{minipage}[b]{0.19\textwidth}
			\includegraphics[width=1\textwidth]{pics/cc_pictures/llama_multifieldqa_en.png}
			\caption* {(3) MF-en}
			\label{cifar:closedsetnoise}
		\end{minipage}%
	}%
    \subfigure{
    	\begin{minipage}[b]{0.19\textwidth}
    		\includegraphics[width=1\textwidth]{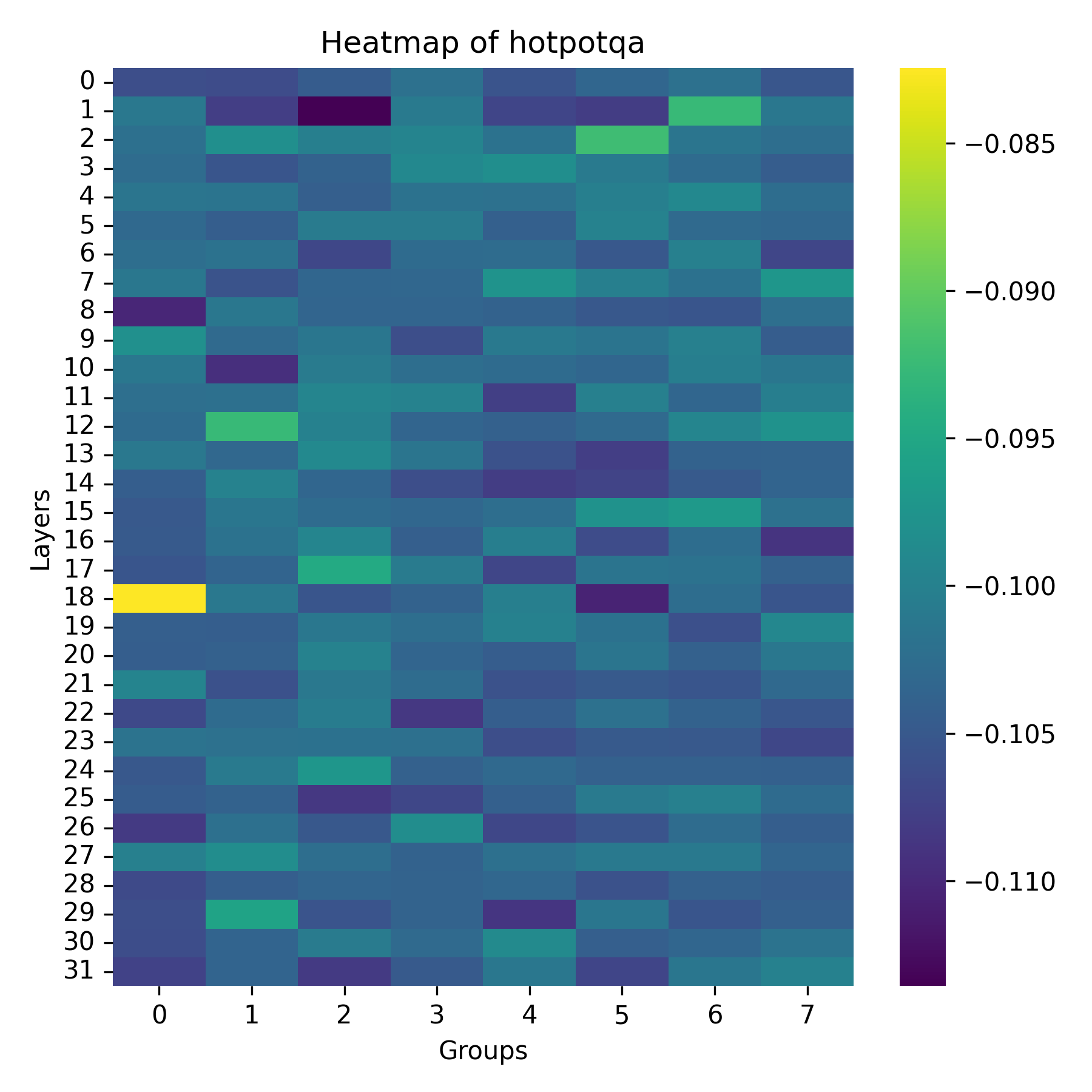}
    		\caption* {(4) HotpotQA}
    		\label{cifar:datanoise}
    	\end{minipage}%
    }%

    \subfigure{
		\begin{minipage}[b]{0.19\textwidth}
			\includegraphics[width=1\textwidth]{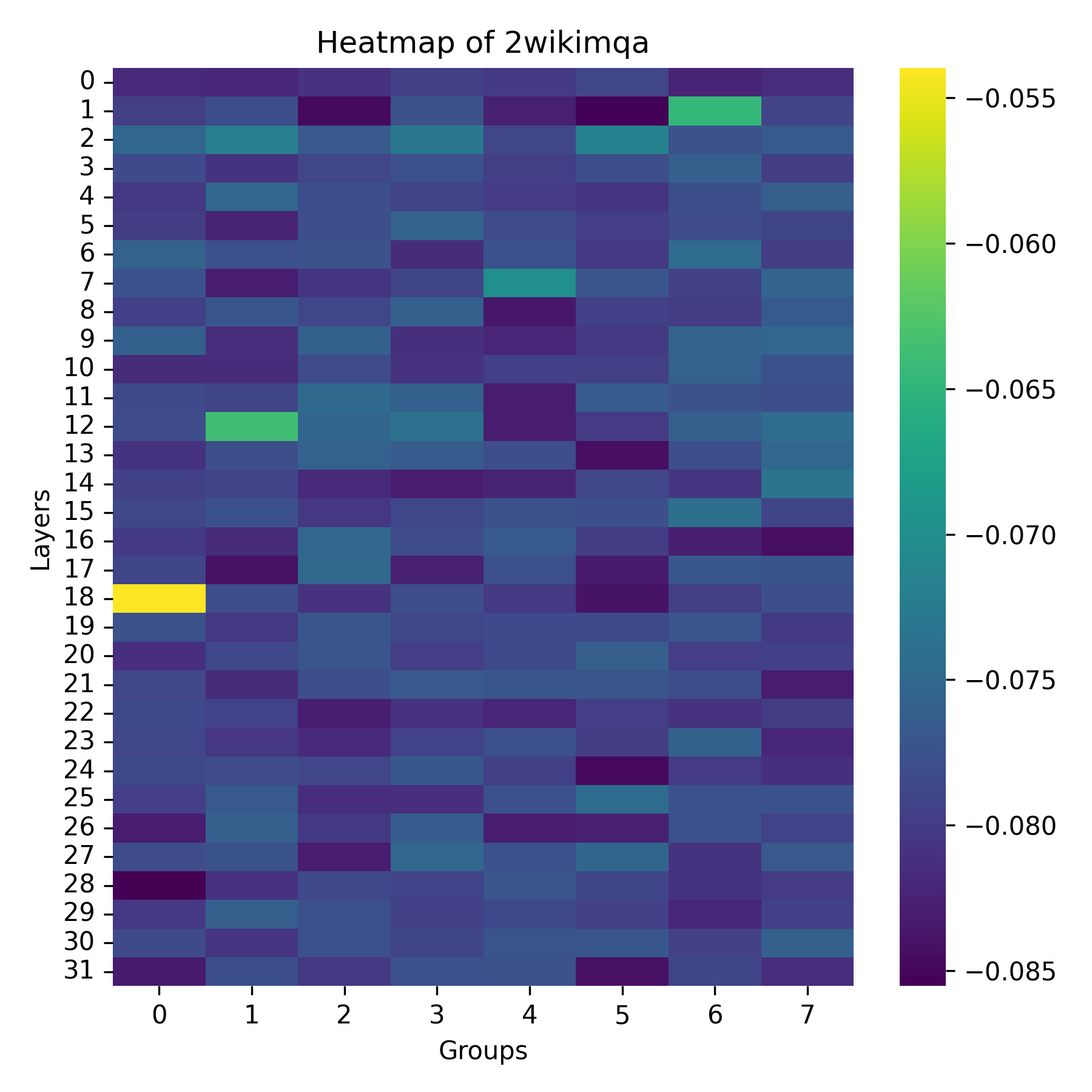}
			\caption* {(5) 2WikiMQA}
			\label{cifar:longtail}
		\end{minipage}%
	}
	\subfigure{
		\begin{minipage}[b]{0.19\textwidth}
			\includegraphics[width=1\textwidth]{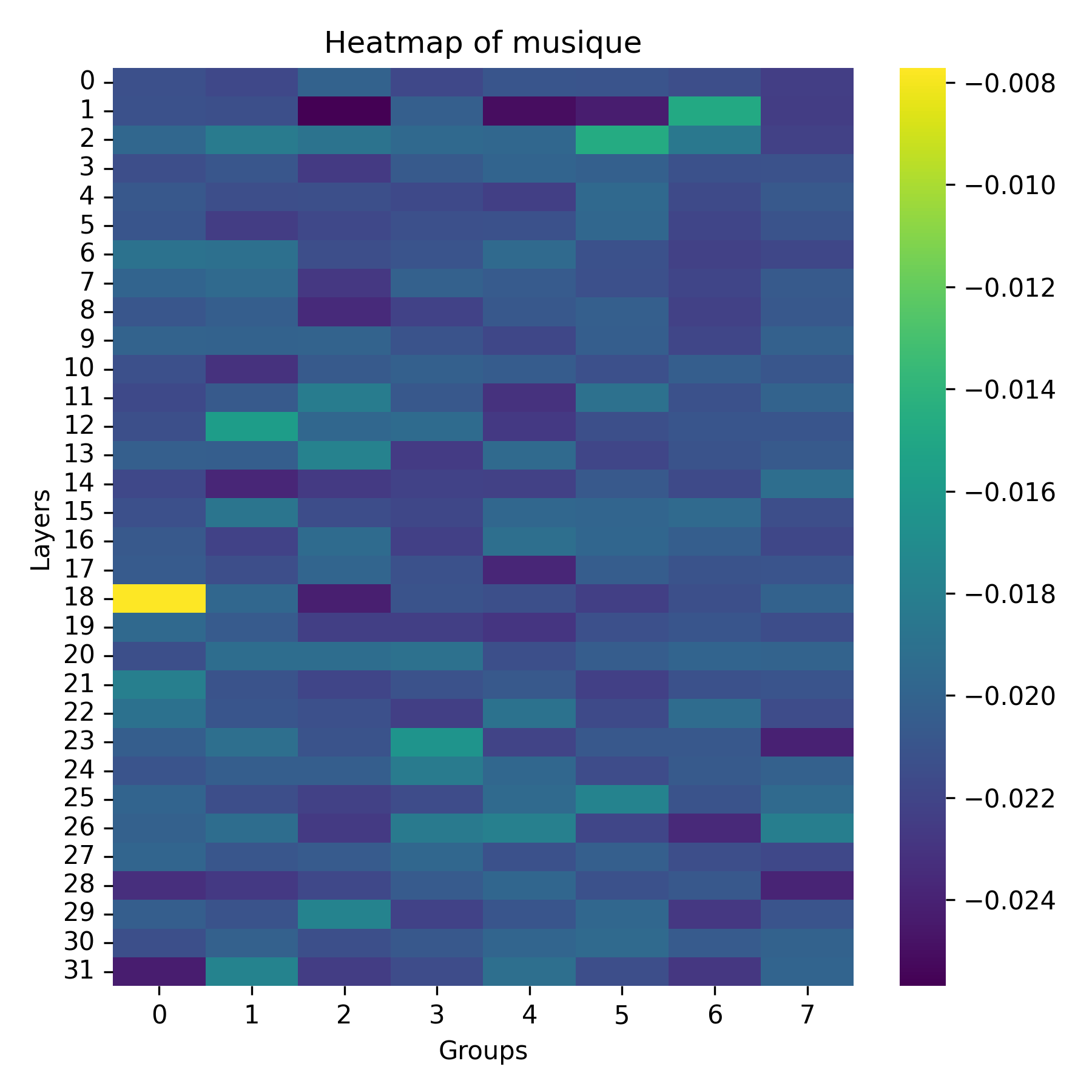}
			\caption* {(6) Musique}
			\label{cifar:opensetnoise}
		\end{minipage}%
	}%
    \subfigure{
		\begin{minipage}[b]{0.19\textwidth}
			\includegraphics[width=1\textwidth]{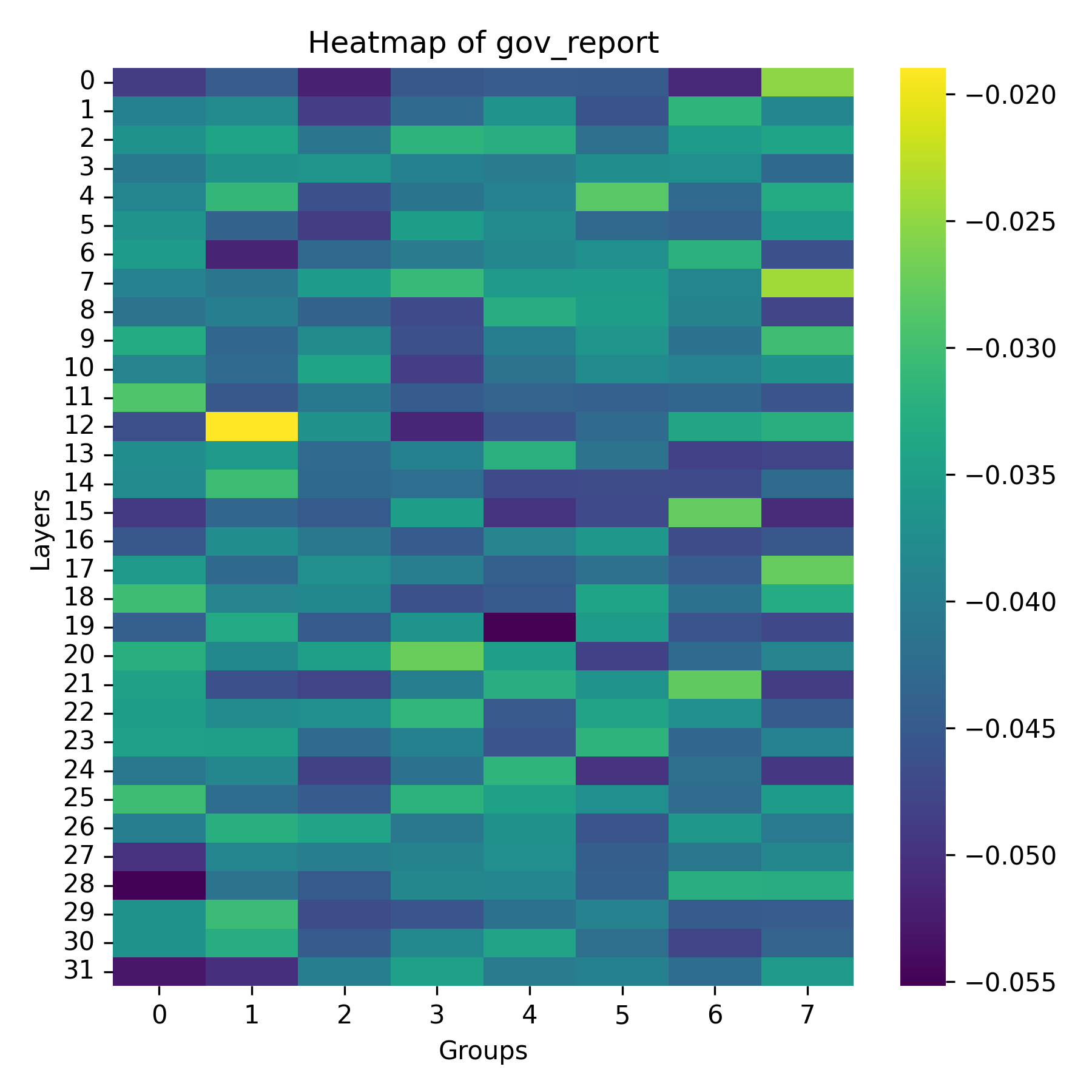}
			\caption* {(7) GovReport}
			\label{cifar:closedsetnoise}
		\end{minipage}%
	}%
    \subfigure{
    	\begin{minipage}[b]{0.19\textwidth}
    		\includegraphics[width=1\textwidth]{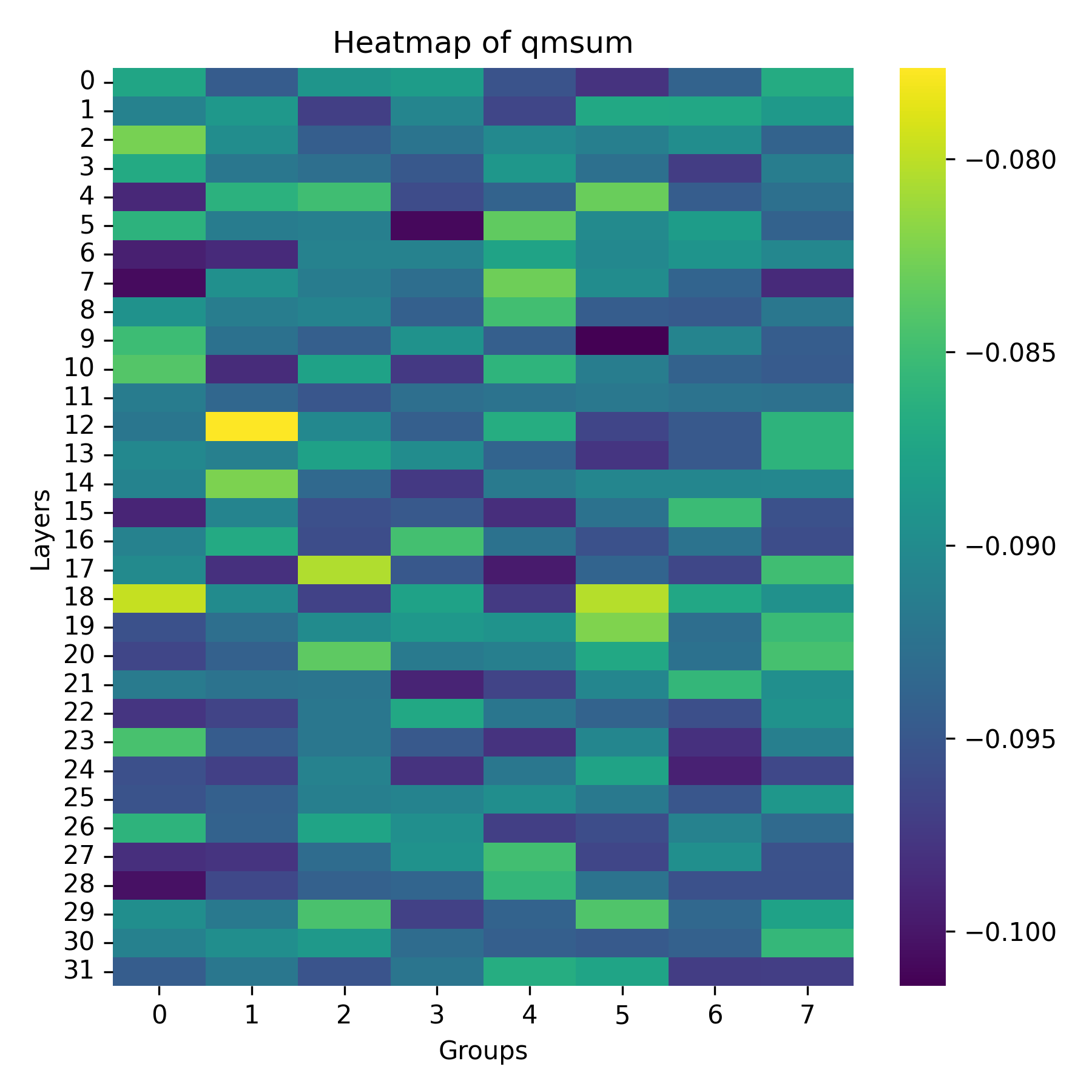}
    		\caption* {(8) QMSum}
    		\label{cifar:datanoise}
    	\end{minipage}%
    }%

    \subfigure{
		\begin{minipage}[b]{0.19\textwidth}
			\includegraphics[width=1\textwidth]{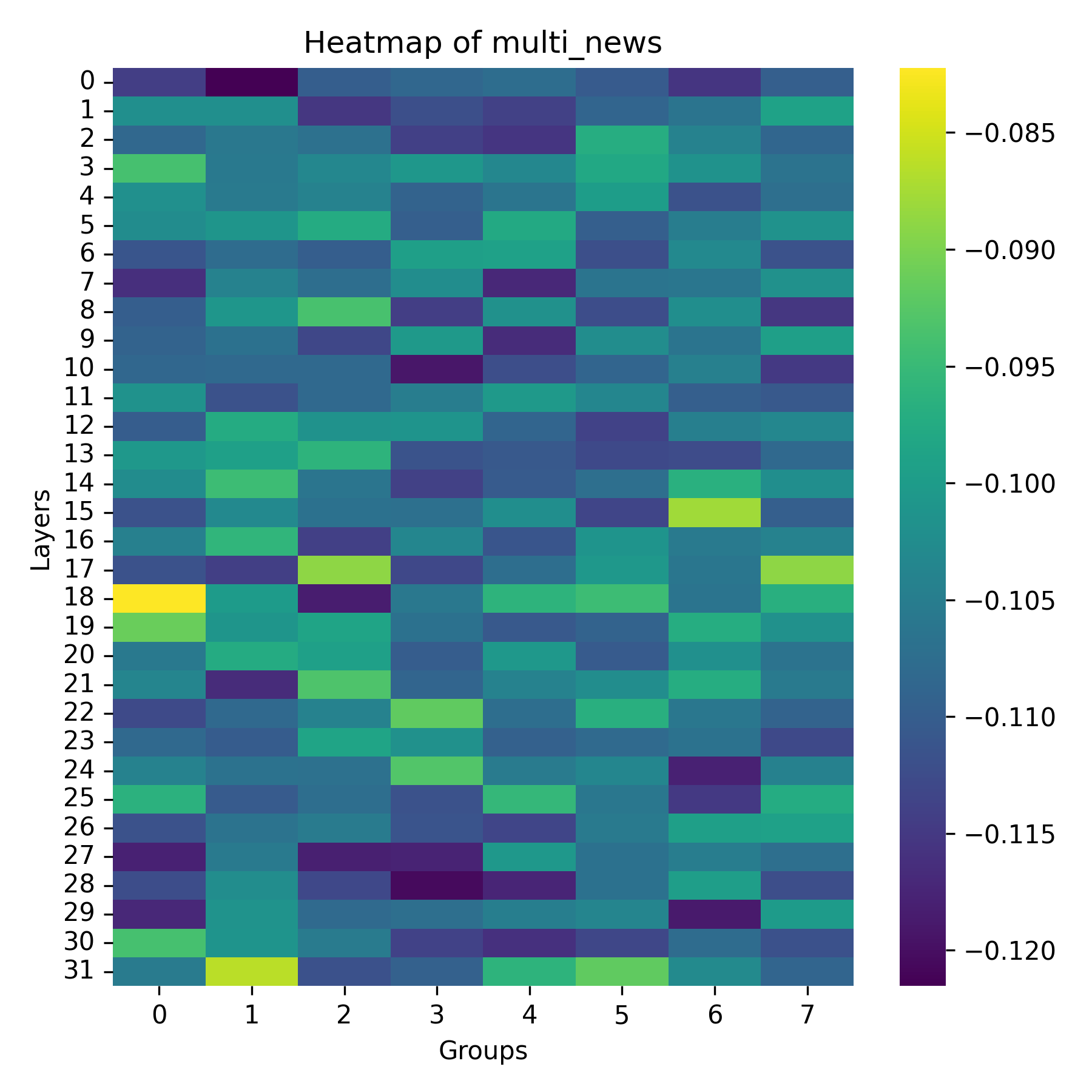}
			\caption* {(9) MultiNews}
			\label{cifar:longtail}
		\end{minipage}%
	}
	\subfigure{
		\begin{minipage}[b]{0.19\textwidth}
			\includegraphics[width=1\textwidth]{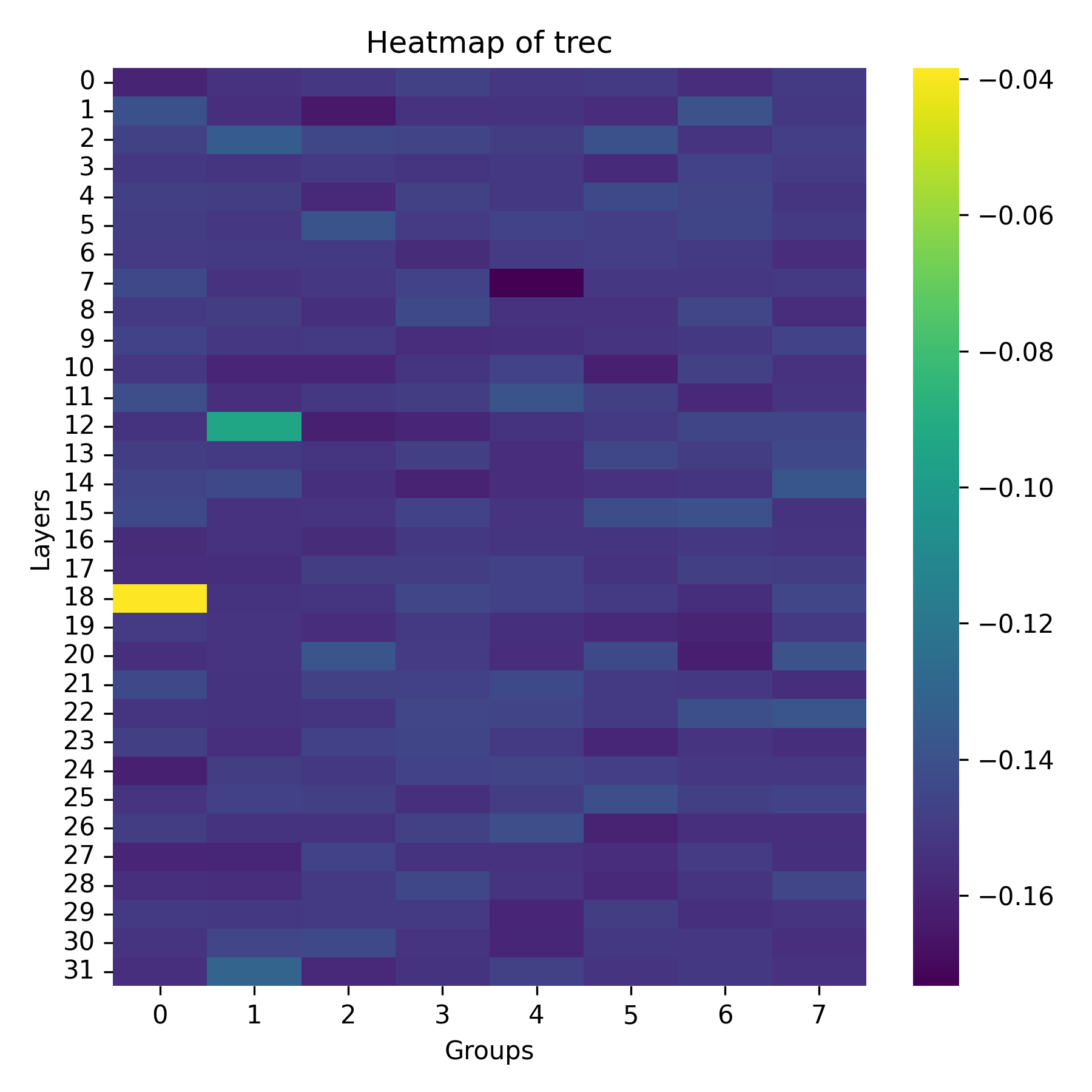}
			\caption* {(10) TREC}
			\label{cifar:opensetnoise}
		\end{minipage}%
	}%
    \subfigure{
		\begin{minipage}[b]{0.19\textwidth}
			\includegraphics[width=1\textwidth]{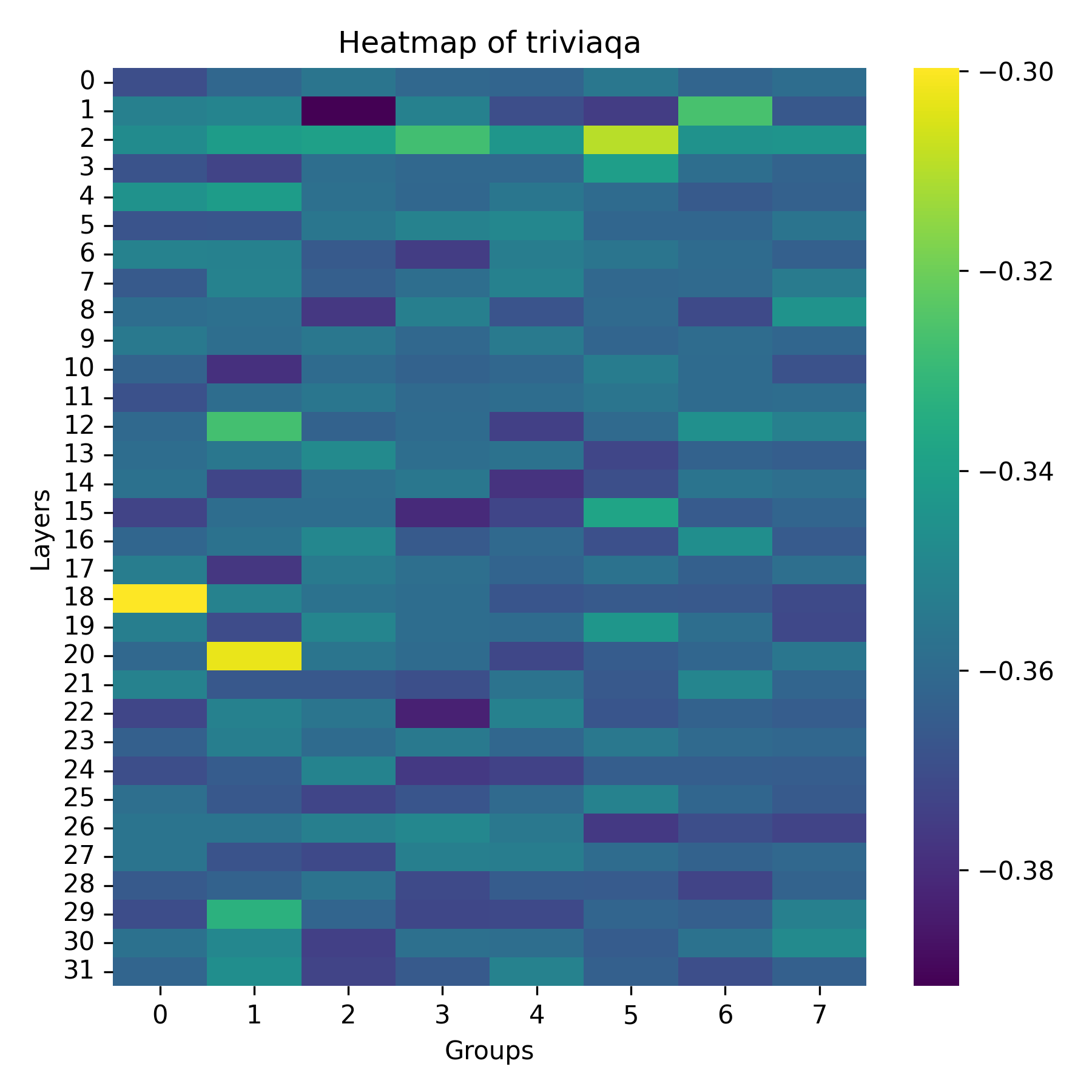}
			\caption* {(11) TriviaQA}
			\label{cifar:closedsetnoise}
		\end{minipage}%
	}%
    \subfigure{
    	\begin{minipage}[b]{0.19\textwidth}
    		\includegraphics[width=1\textwidth]{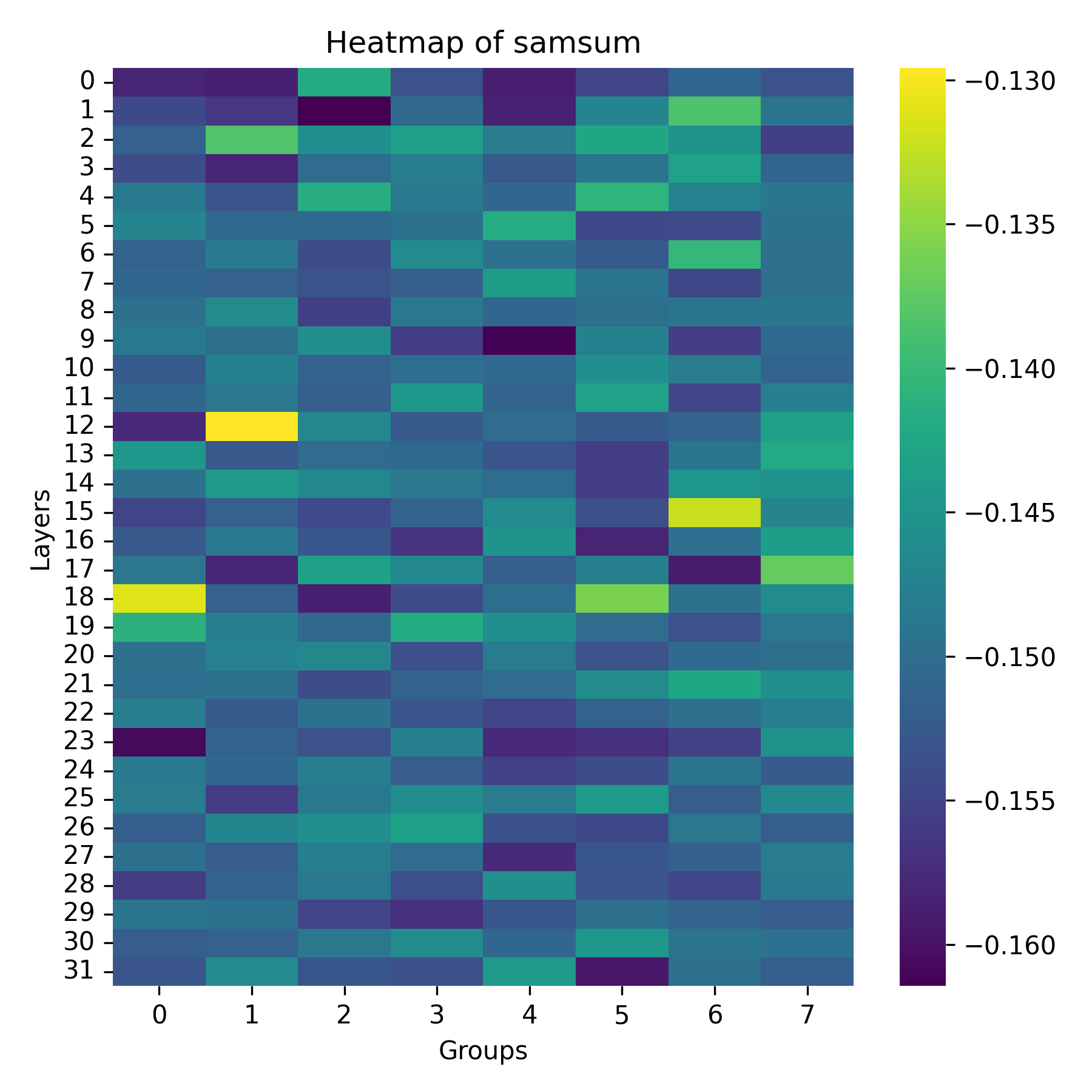}
    		\caption* {(12) SAMSum}
    		\label{cifar:datanoise}
    	\end{minipage}%
    }%

    \subfigure{
		\begin{minipage}[b]{0.19\textwidth}
			\includegraphics[width=1\textwidth]{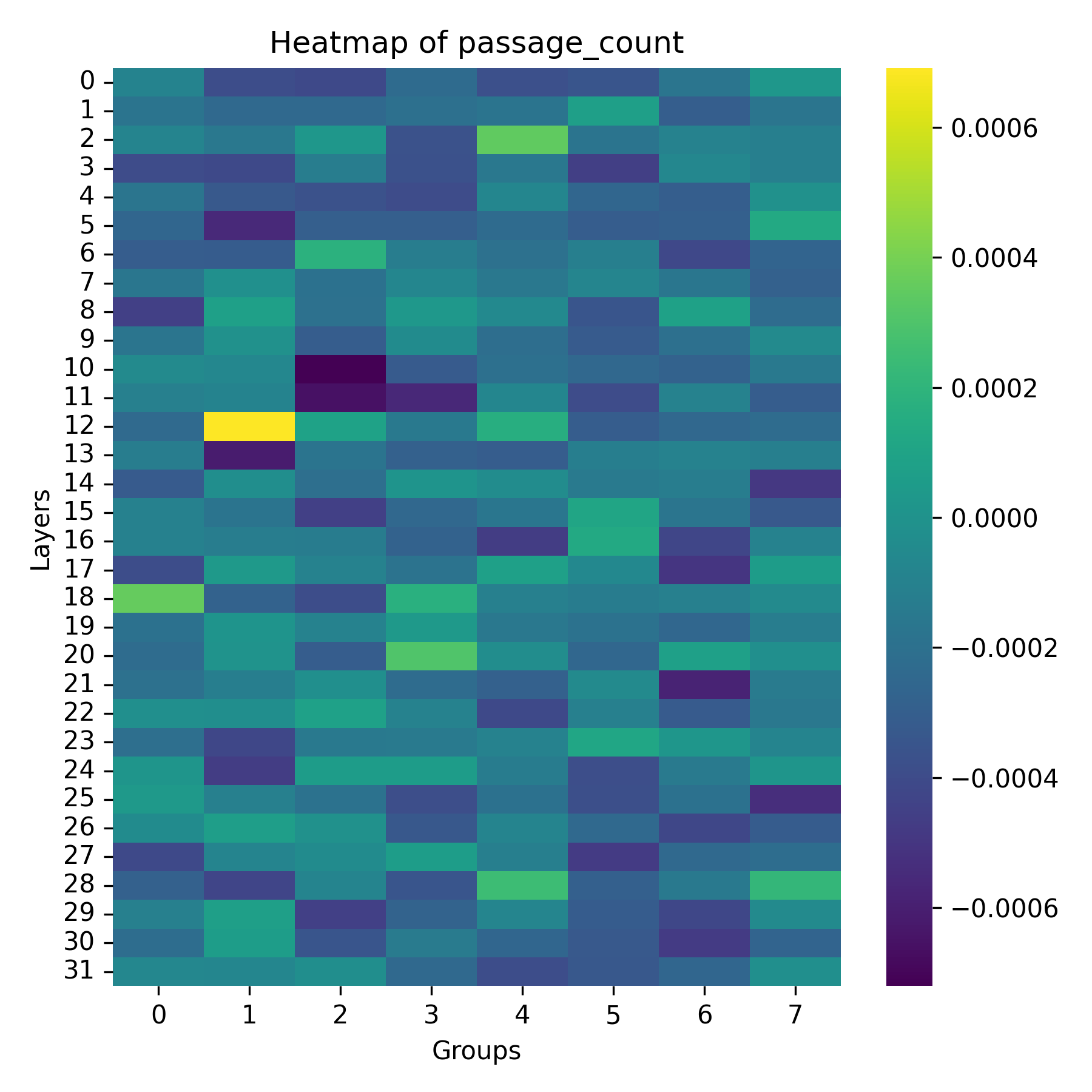}
			\caption* {(13) PCount}
			\label{cifar:longtail}
		\end{minipage}%
	}
	\subfigure{
		\begin{minipage}[b]{0.19\textwidth}
			\includegraphics[width=1\textwidth]{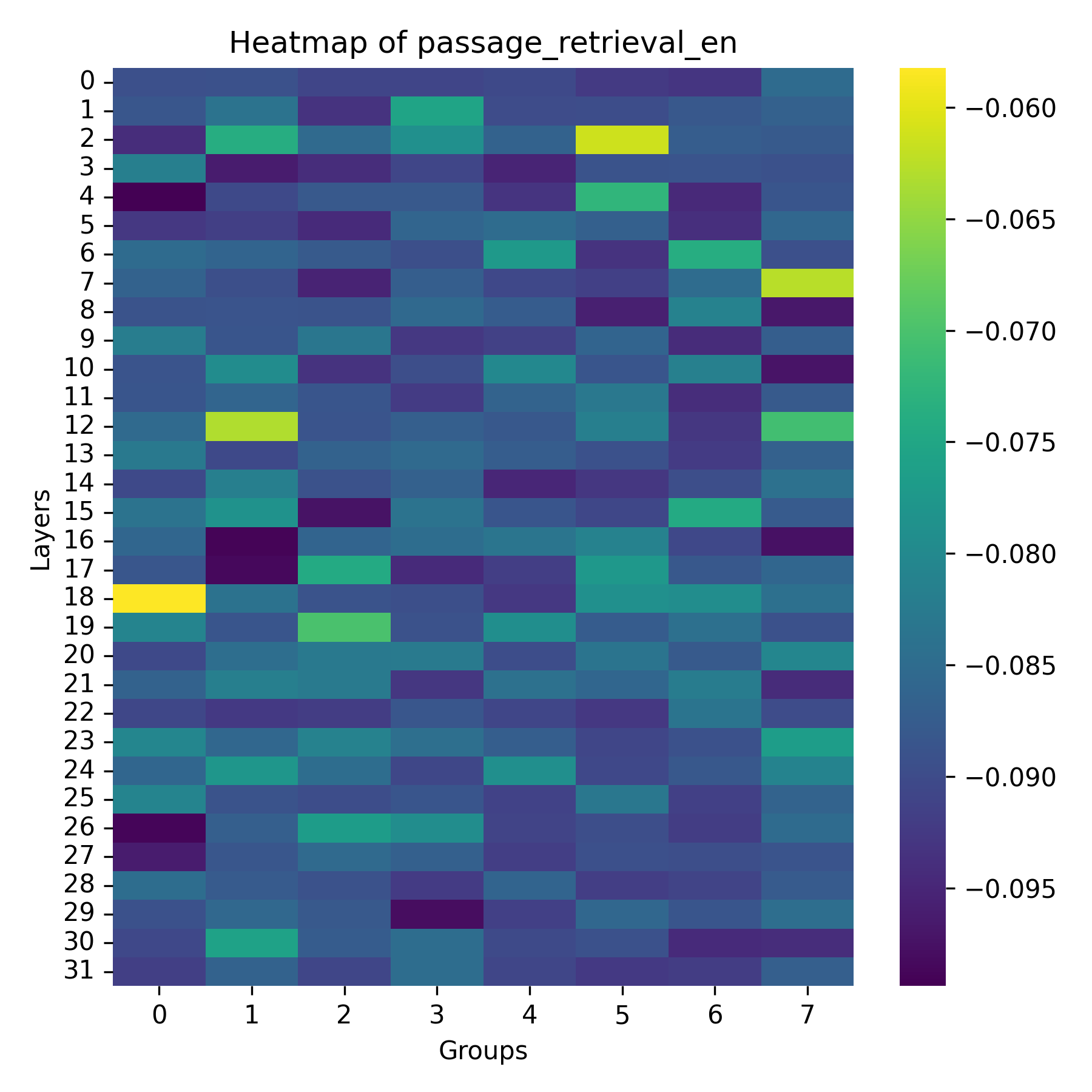}
			\caption* {(14) PRe}
			\label{cifar:opensetnoise}
		\end{minipage}%
	}%
    \subfigure{
		\begin{minipage}[b]{0.19\textwidth}
			\includegraphics[width=1\textwidth]{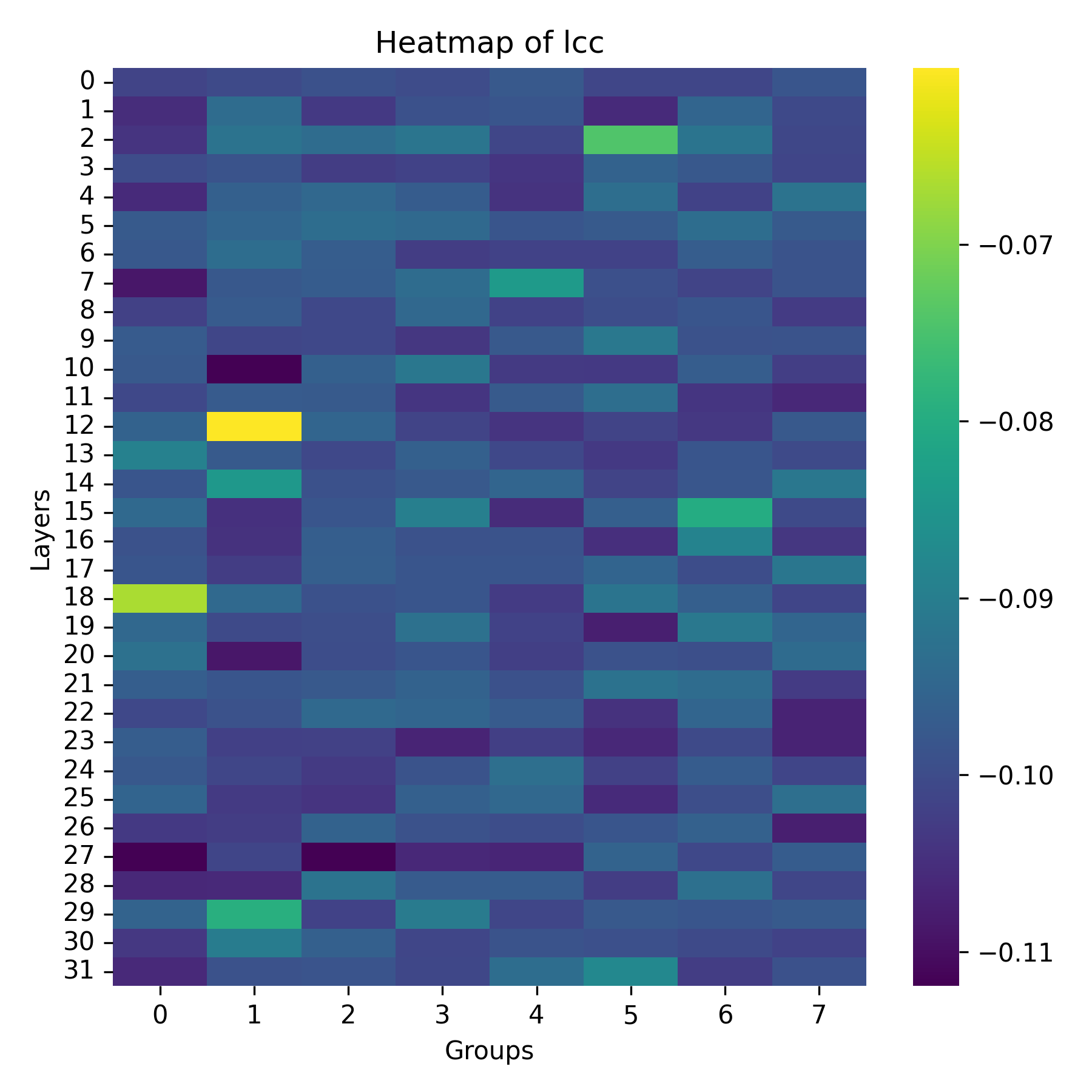}
			\caption* {(15) Lcc}
			\label{cifar:closedsetnoise}
		\end{minipage}%
	}%
    \subfigure{
    	\begin{minipage}[b]{0.19\textwidth}
    		\includegraphics[width=1\textwidth]{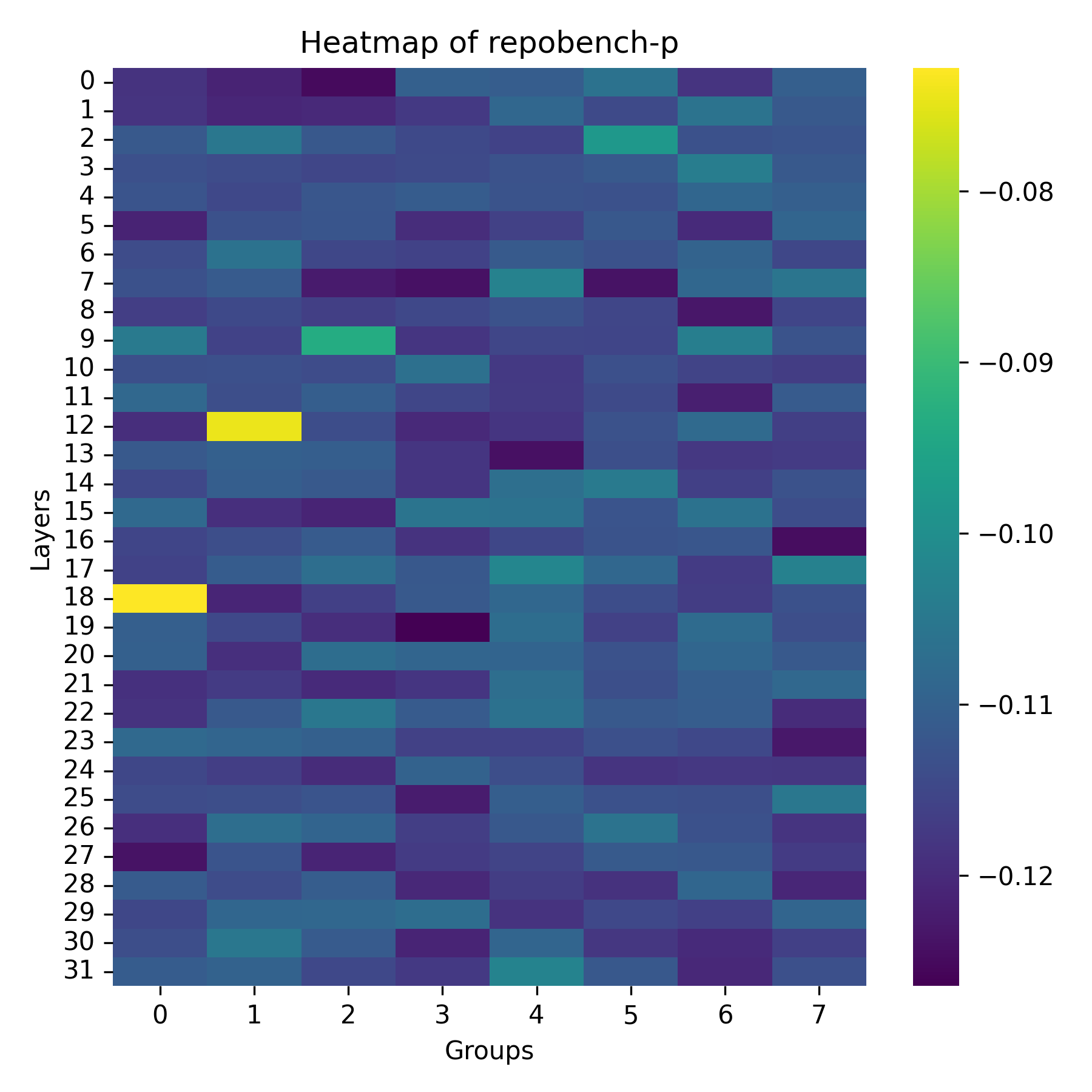}
    		\caption* {(16) RB-P}
    		\label{cifar:datanoise}
    	\end{minipage}%
    }%
    
\caption {Heatmap of Mistral-7B-Instruct-v0.2. }
\label{fig:heatmap_mistral}
\end{figure*}

Figures~\ref{fig:heatmap_llama} and~\ref{fig:heatmap_mistral} illustrate the distribution of the Sliced Shapley values computed for selected coalition sizes 
$H=\{32,64,96,128\}$ in our experiment. We observe that the distributions of Sliced Shapley values exhibit significant differences across datasets of different task categories, while showing relatively smaller variations within datasets of the same domain type.

\subsection{Distribution of j-coalition Complementary Contribution}\label{sec:coalition_distribution}
\begin{figure*}[t]
    \centering
	\subfigure{
		\begin{minipage}[b]{0.19\textwidth}
			\includegraphics[width=1\textwidth]{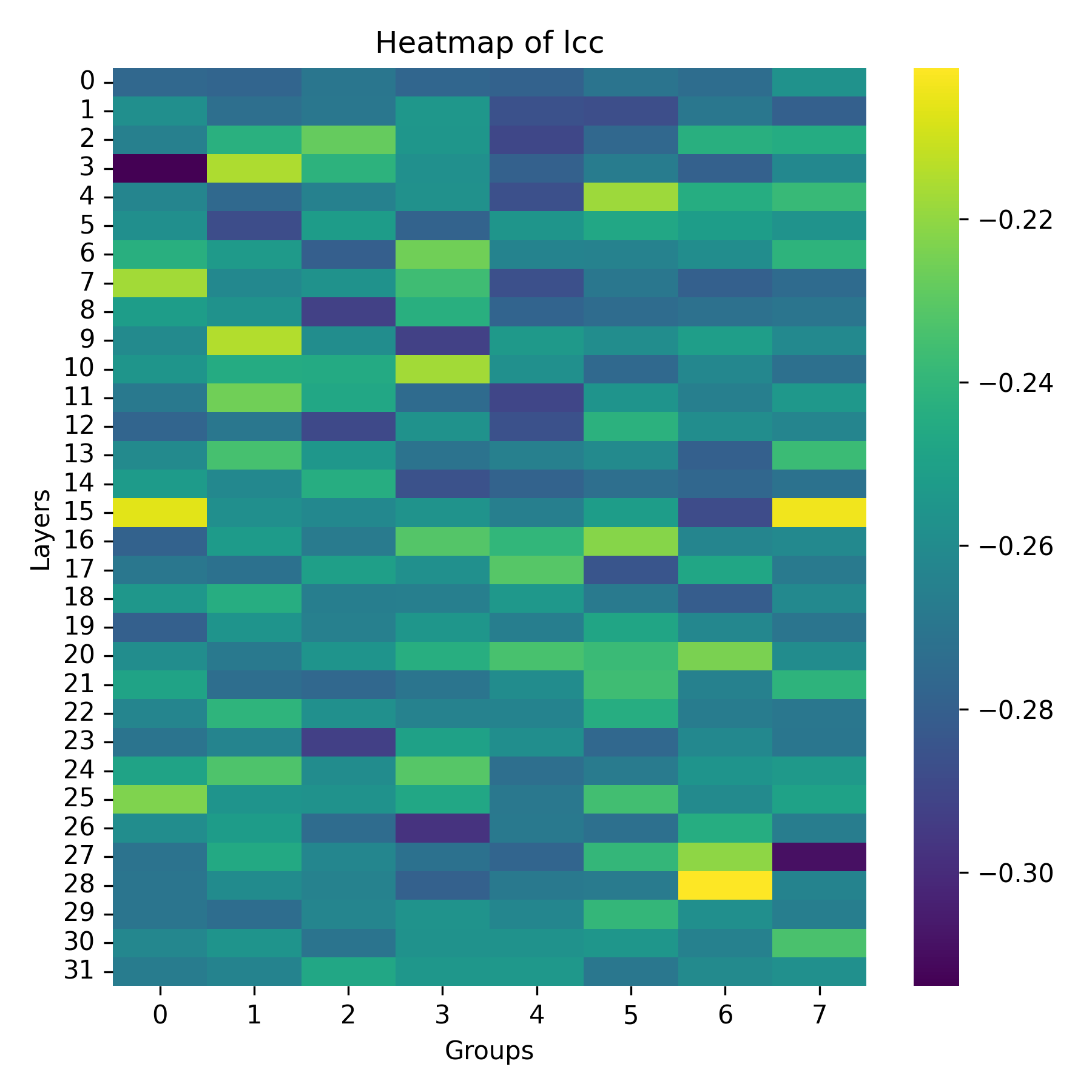}
			\caption* {(1) Coalition size 32}
			\label{cifar:opensetnoise}
		\end{minipage}%
	}%
    \subfigure{
		\begin{minipage}[b]{0.19\textwidth}
			\includegraphics[width=1\textwidth]{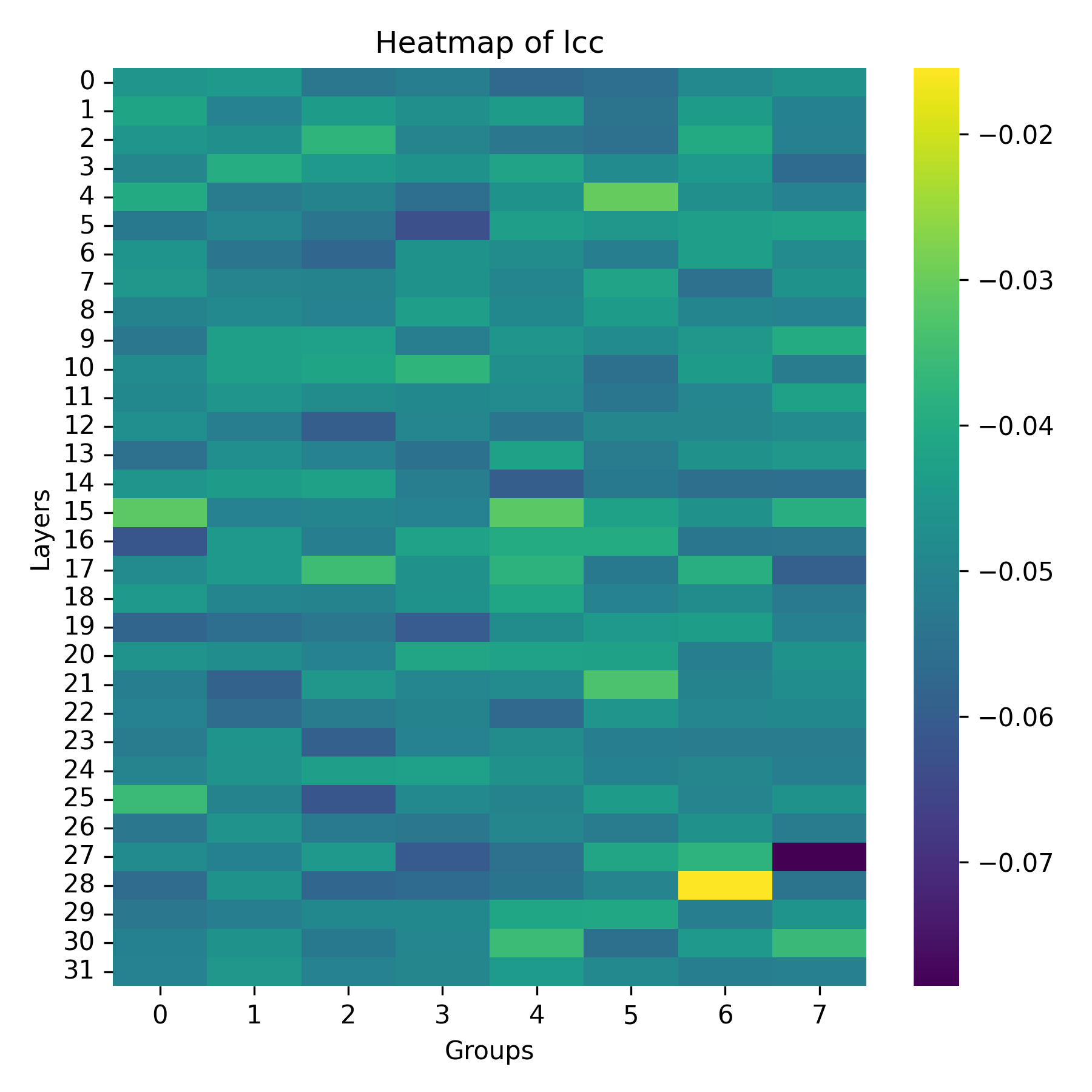}
			\caption* {(2) Coalition size 64}
			\label{cifar:closedsetnoise}
		\end{minipage}%
	}%
    \subfigure{
    	\begin{minipage}[b]{0.19\textwidth}
    		\includegraphics[width=1\textwidth]{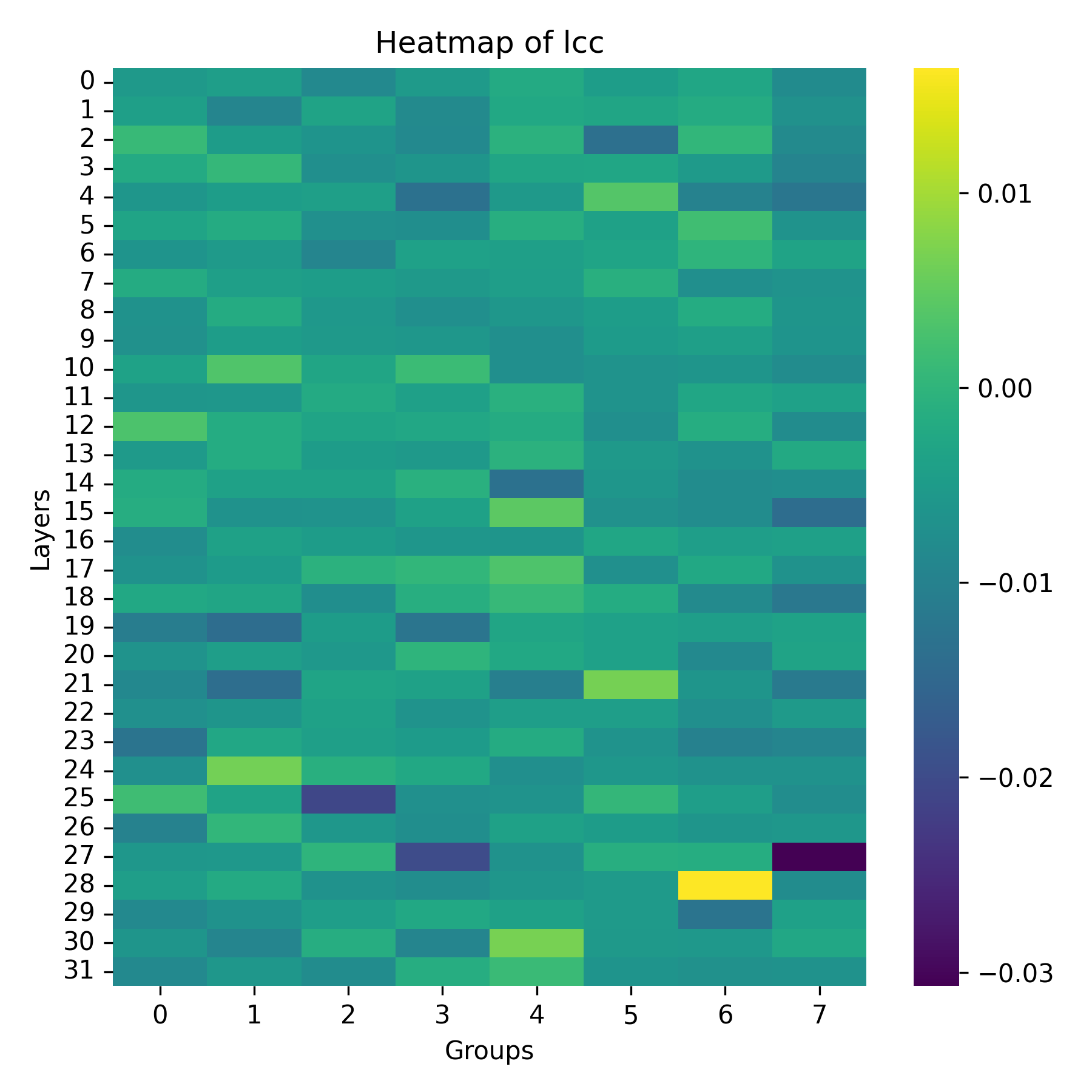}
    		\caption* {(3) Coalition size 96}
    		\label{cifar:datanoise}
    	\end{minipage}%
    }%
    \subfigure{
		\begin{minipage}[b]{0.19\textwidth}
			\includegraphics[width=1\textwidth]{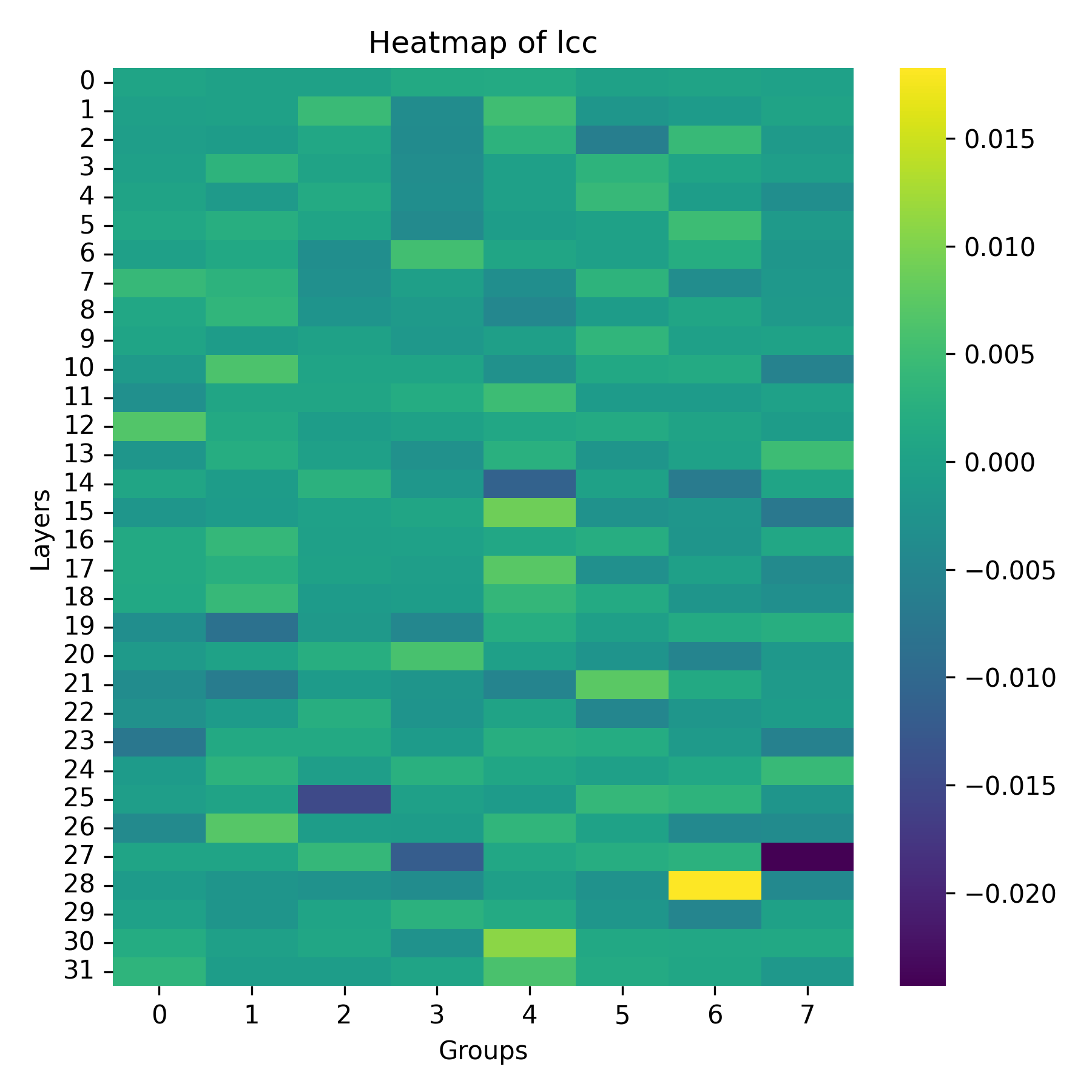}
			\caption* {(4) Coalition size 128}
			\label{gradient:cifar}
		\end{minipage}%
	}%
    \\

    \subfigure{
		\begin{minipage}[b]{0.19\textwidth}
			\includegraphics[width=1\textwidth]{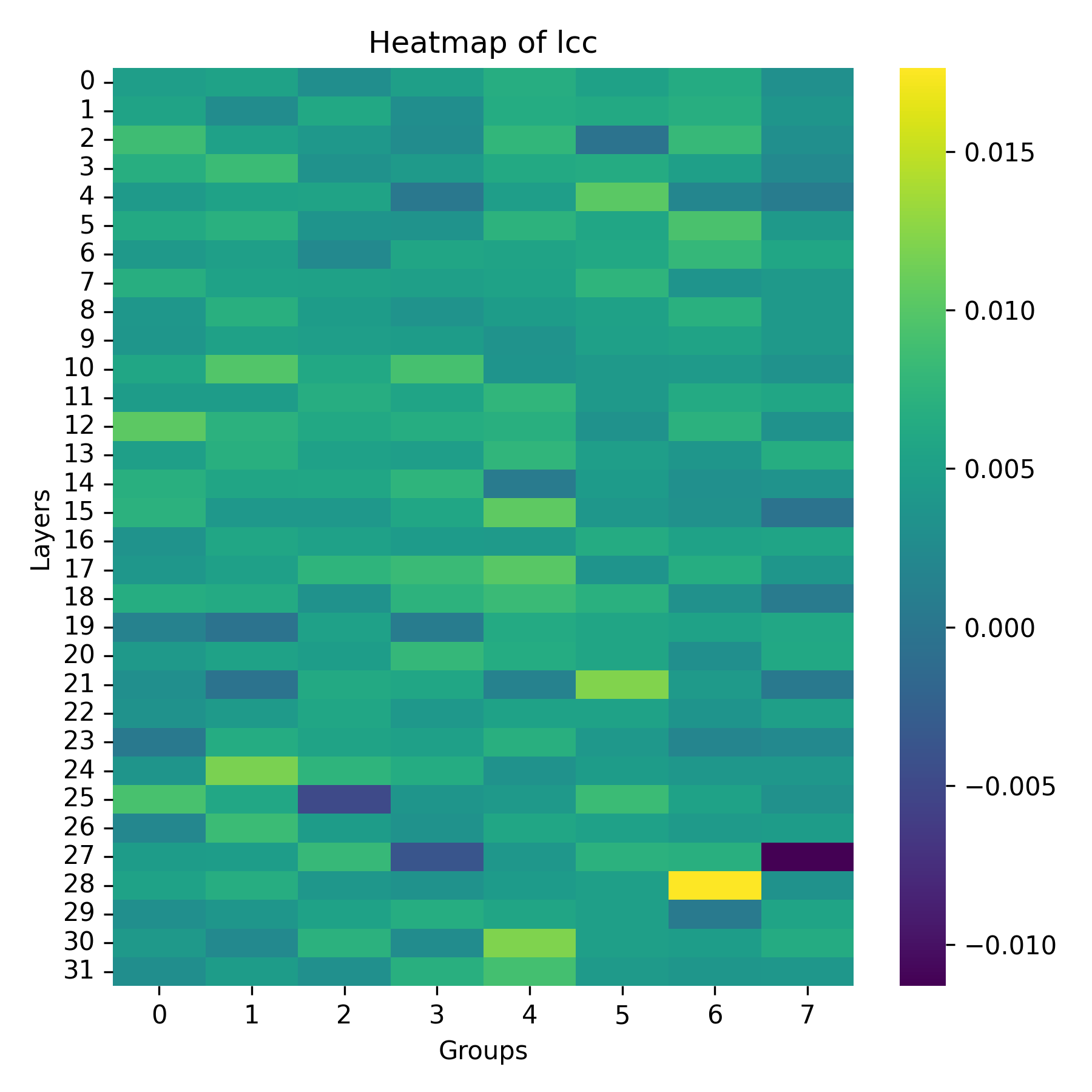}
			\caption* {(5) Coalition size 160}
			\label{cifar:opensetnoise}
		\end{minipage}%
	}%
    \subfigure{
		\begin{minipage}[b]{0.19\textwidth}
			\includegraphics[width=1\textwidth]{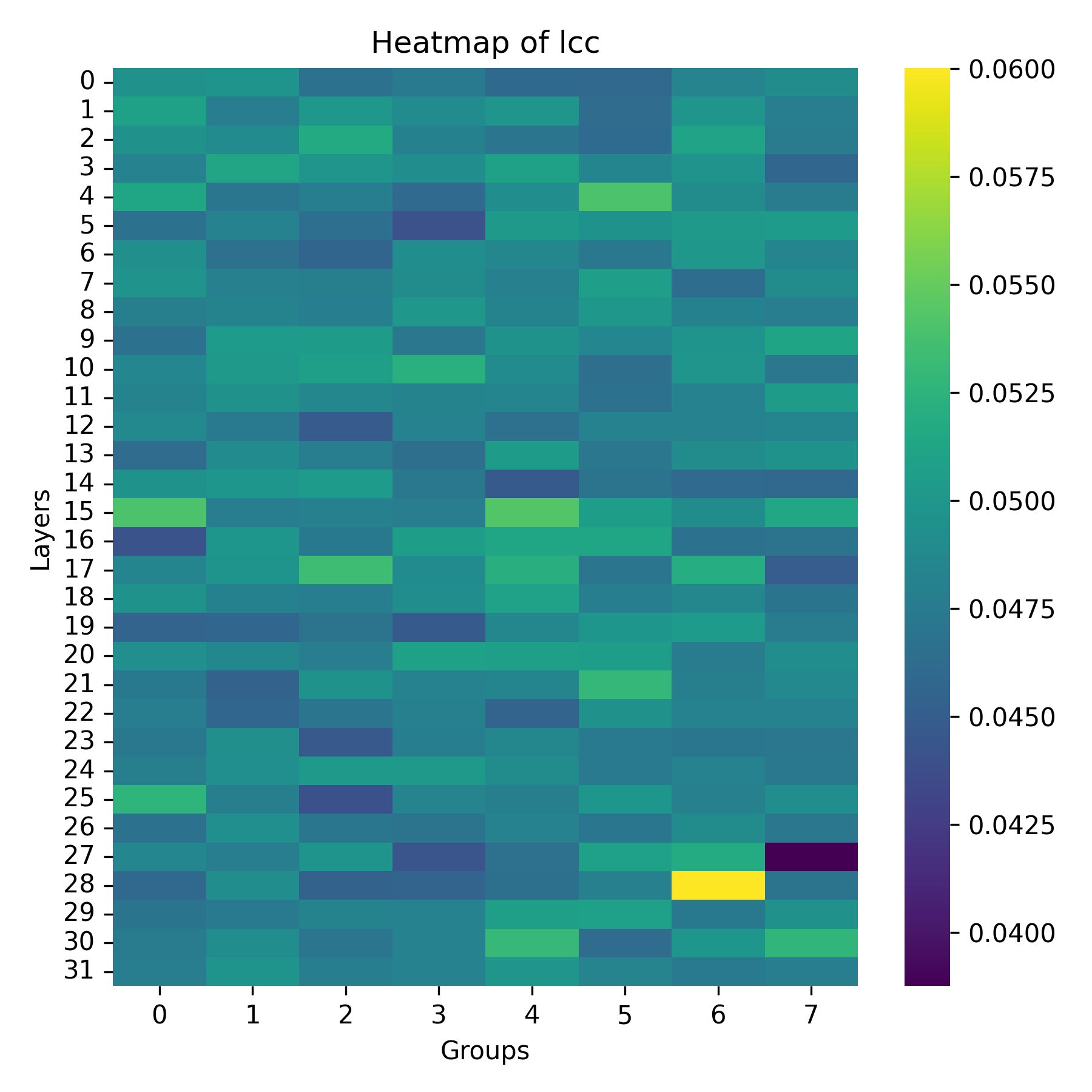}
			\caption* {(6) Coalition size 192}
			\label{cifar:closedsetnoise}
		\end{minipage}%
	}%
    \subfigure{
    	\begin{minipage}[b]{0.19\textwidth}
    		\includegraphics[width=1\textwidth]{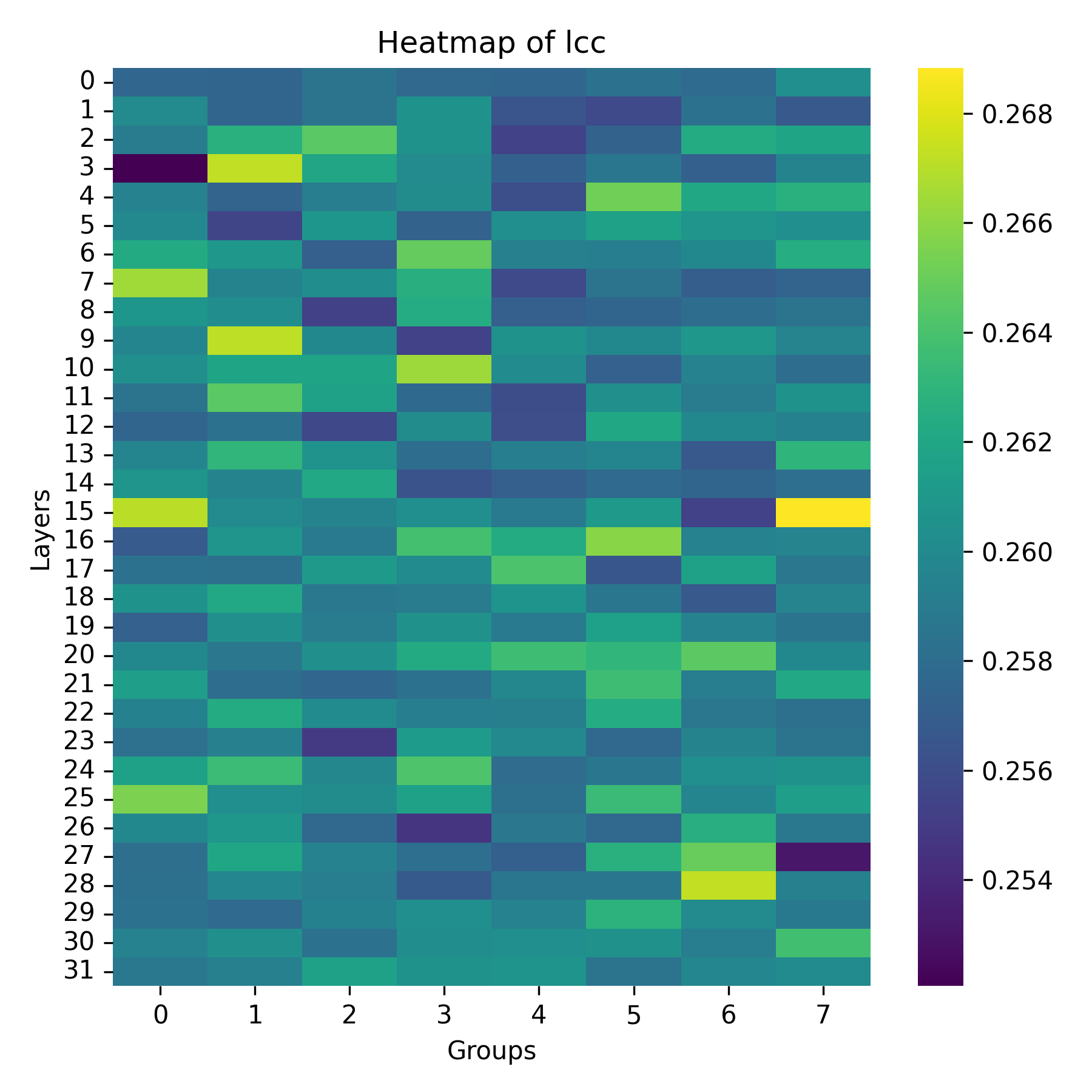}
    		\caption* {(7) Coalition size 224}
    		\label{cifar:datanoise}
    	\end{minipage}%
    }%
    \subfigure{
		\begin{minipage}[b]{0.19\textwidth}
			\includegraphics[width=1\textwidth]{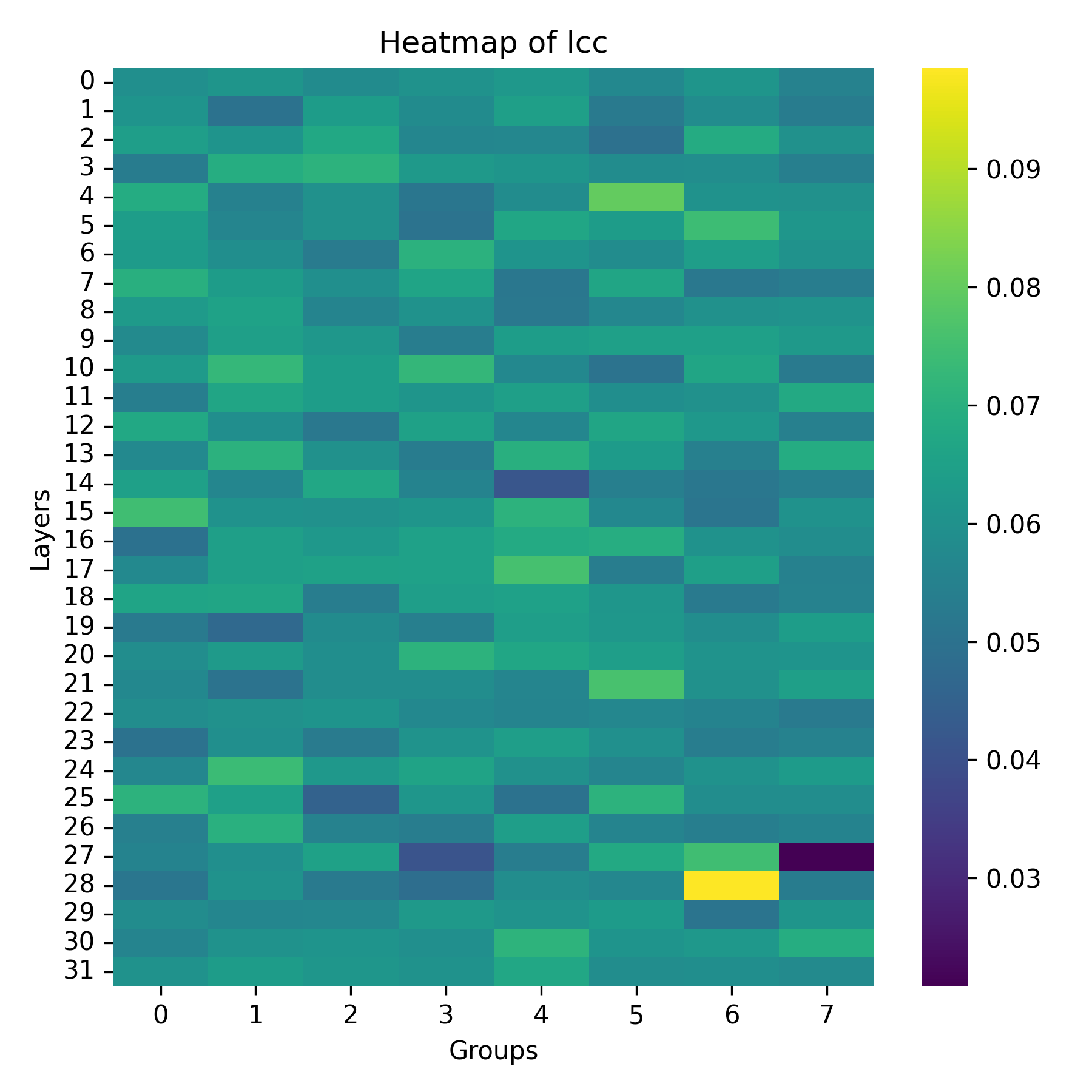}
			\caption* {(8) Average}
			\label{gradient:cifar}
		\end{minipage}%
	}%
 
\caption {The expected complementary contributions for the \emph{lcc} dataset across different coalition sizes.}
\label{fig:heatmap_lcc_coalition}
\end{figure*}

\begin{figure*}[t]
    \centering
	\subfigure{
		\begin{minipage}[b]{0.19\textwidth}
			\includegraphics[width=1\textwidth]{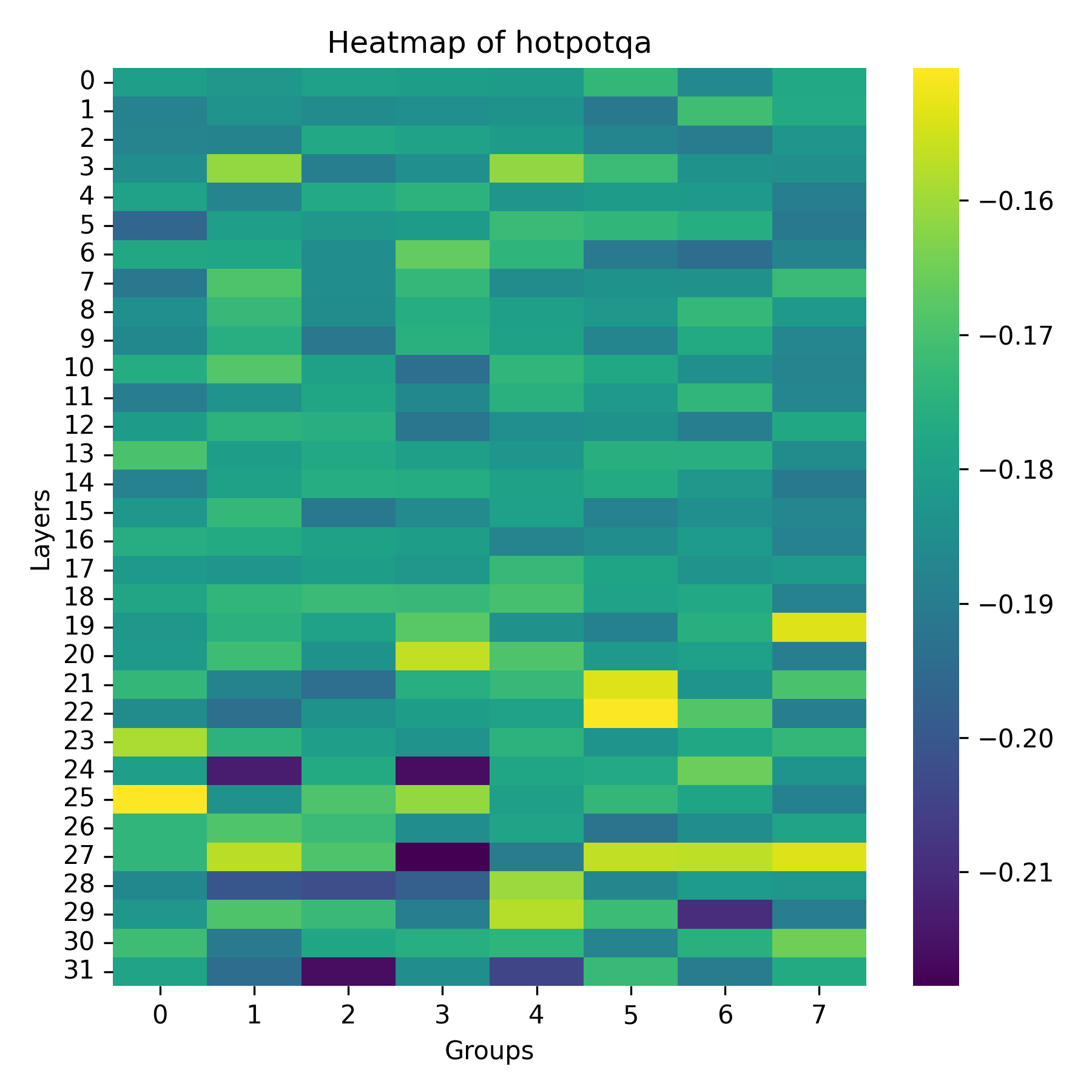}
			\caption* {(1) Coalition size 32}
			\label{cifar:opensetnoise}
		\end{minipage}%
	}%
    \subfigure{
		\begin{minipage}[b]{0.19\textwidth}
			\includegraphics[width=1\textwidth]{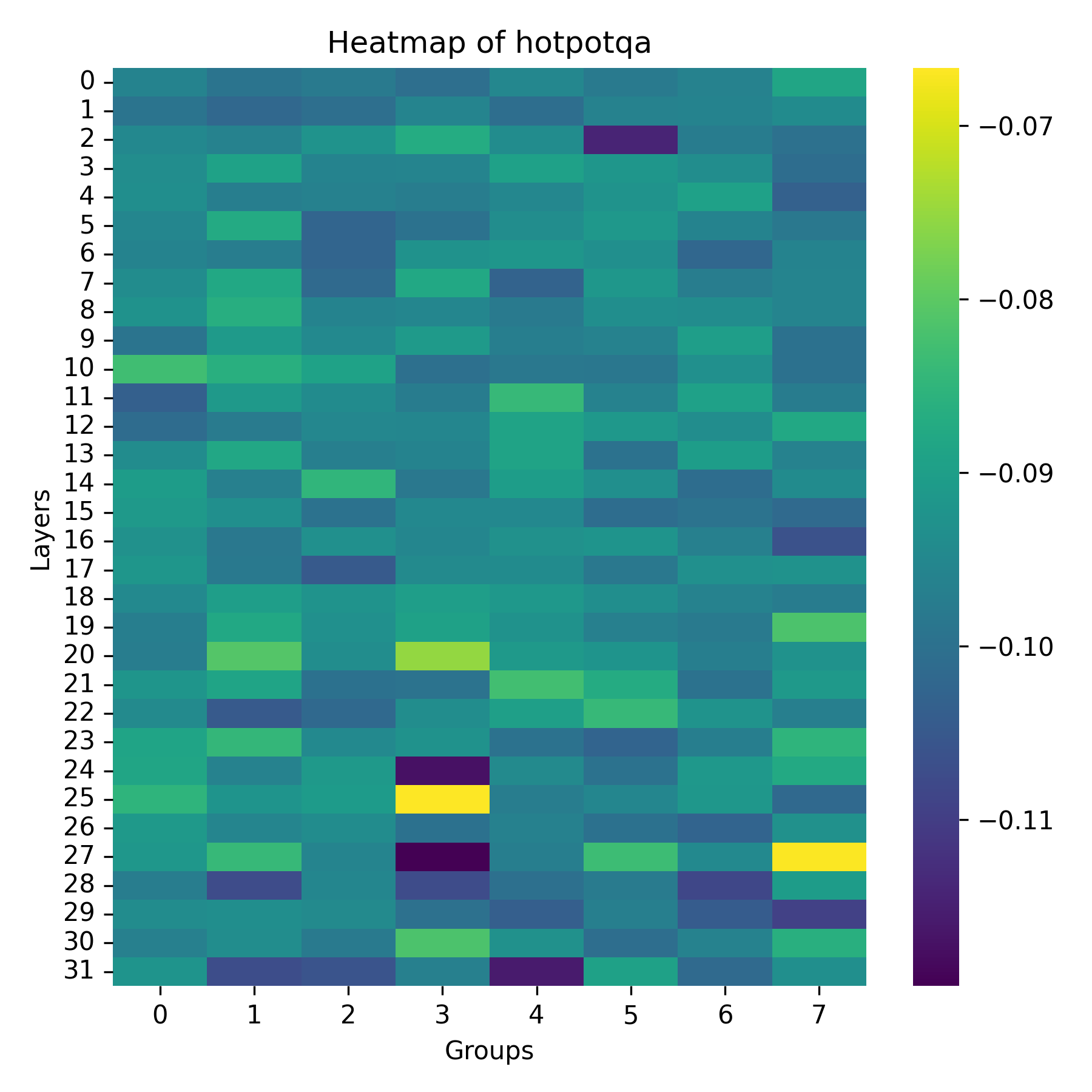}
			\caption* {(2) Coalition size 64}
			\label{cifar:closedsetnoise}
		\end{minipage}%
	}%
    \subfigure{
    	\begin{minipage}[b]{0.19\textwidth}
    		\includegraphics[width=1\textwidth]{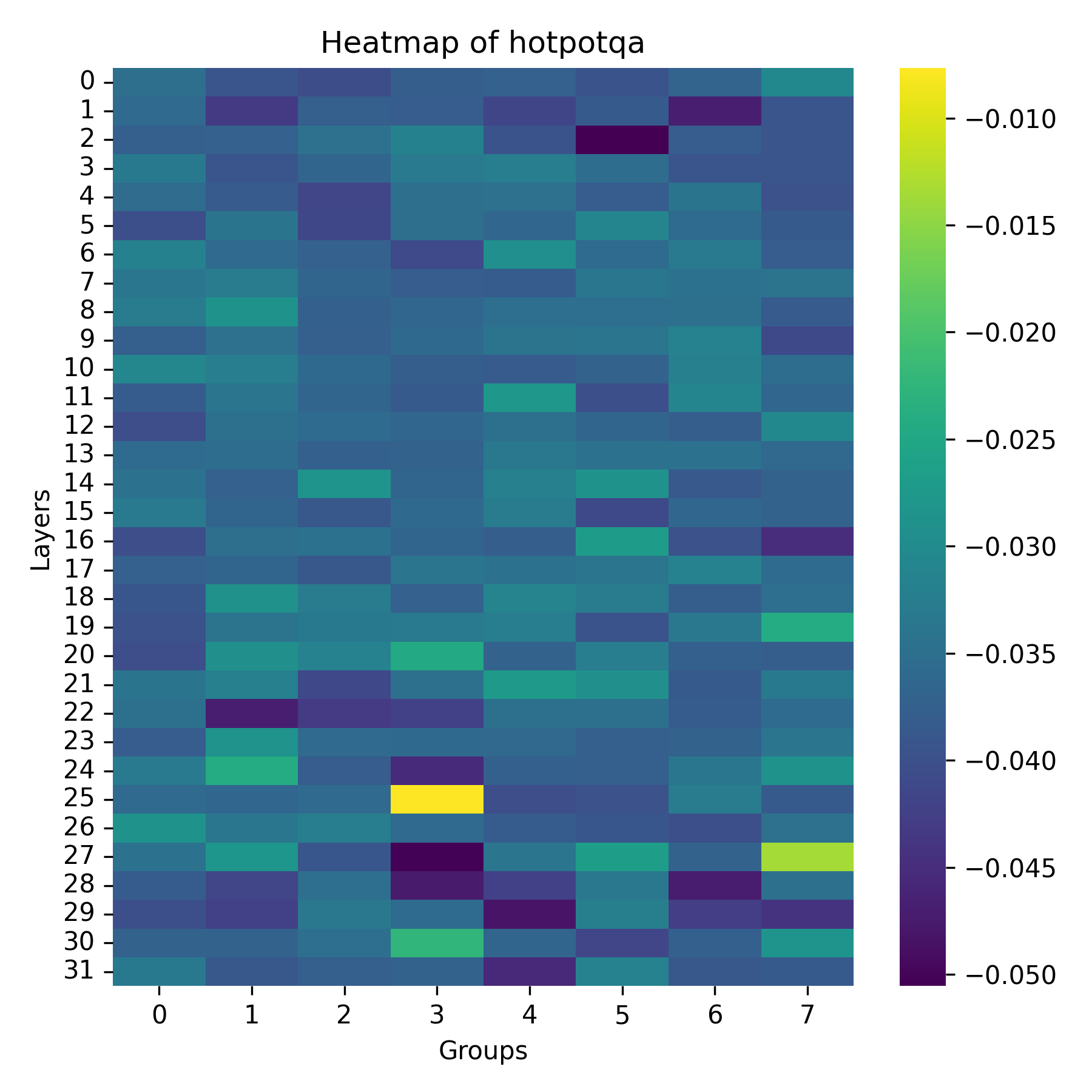}
    		\caption* {(3) Coalition size 96}
    		\label{cifar:datanoise}
    	\end{minipage}%
    }%
    \subfigure{
		\begin{minipage}[b]{0.19\textwidth}
			\includegraphics[width=1\textwidth]{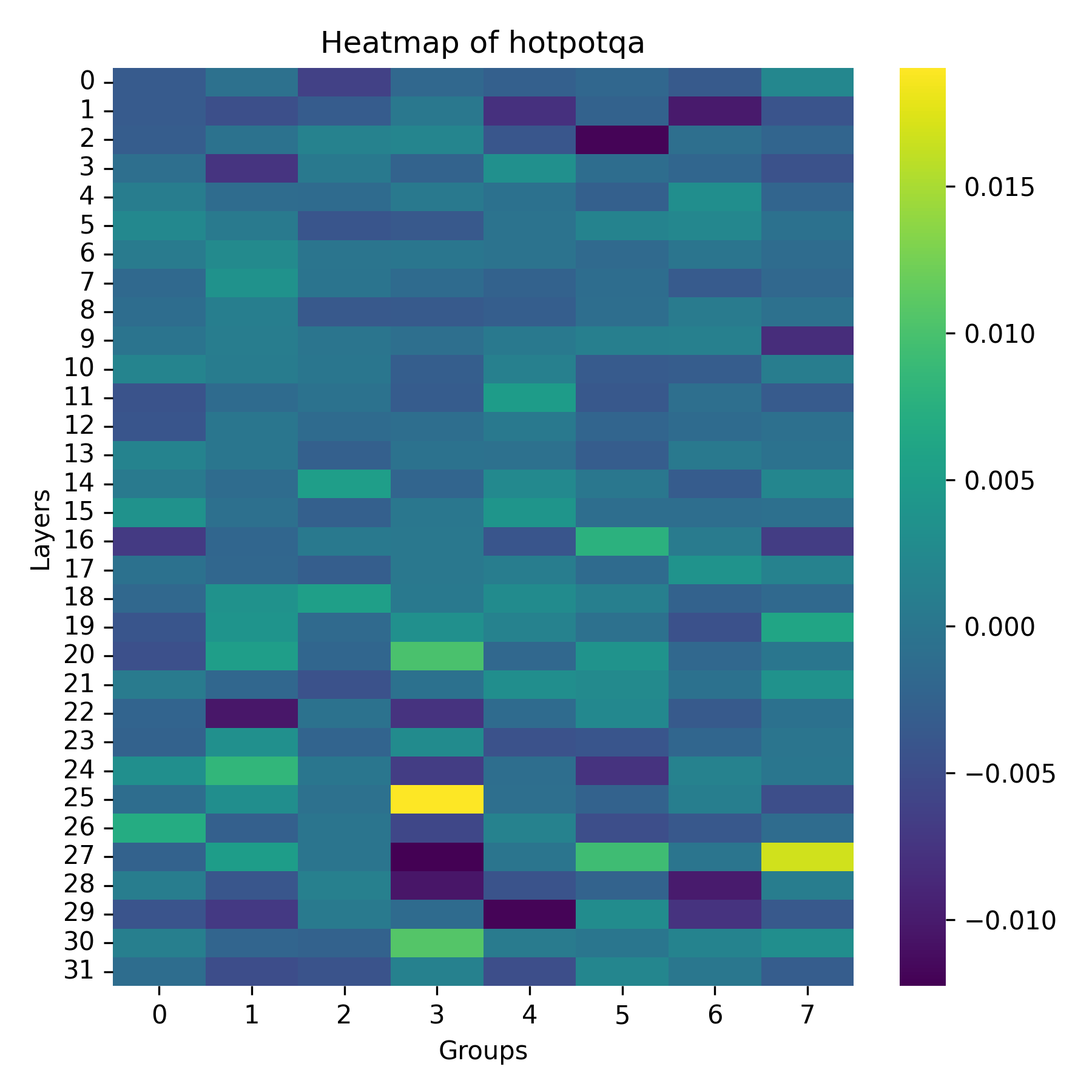}
			\caption* {(4) Coalition size 128}
			\label{gradient:cifar}
		\end{minipage}%
	}%

    \subfigure{
		\begin{minipage}[b]{0.19\textwidth}
			\includegraphics[width=1\textwidth]{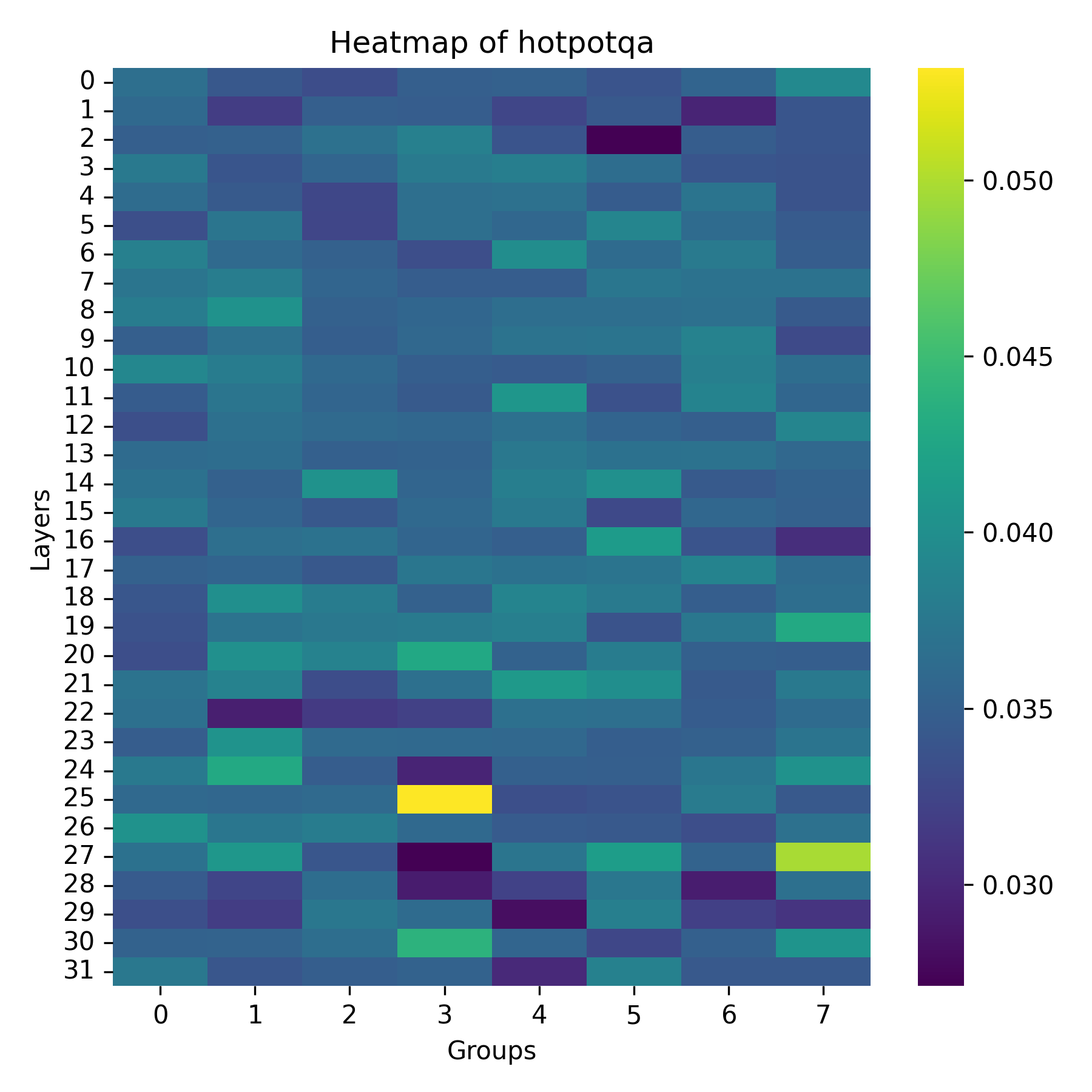}
			\caption* {(5) Coalition size 160}
			\label{cifar:opensetnoise}
		\end{minipage}%
	}%
    \subfigure{
		\begin{minipage}[b]{0.19\textwidth}
			\includegraphics[width=1\textwidth]{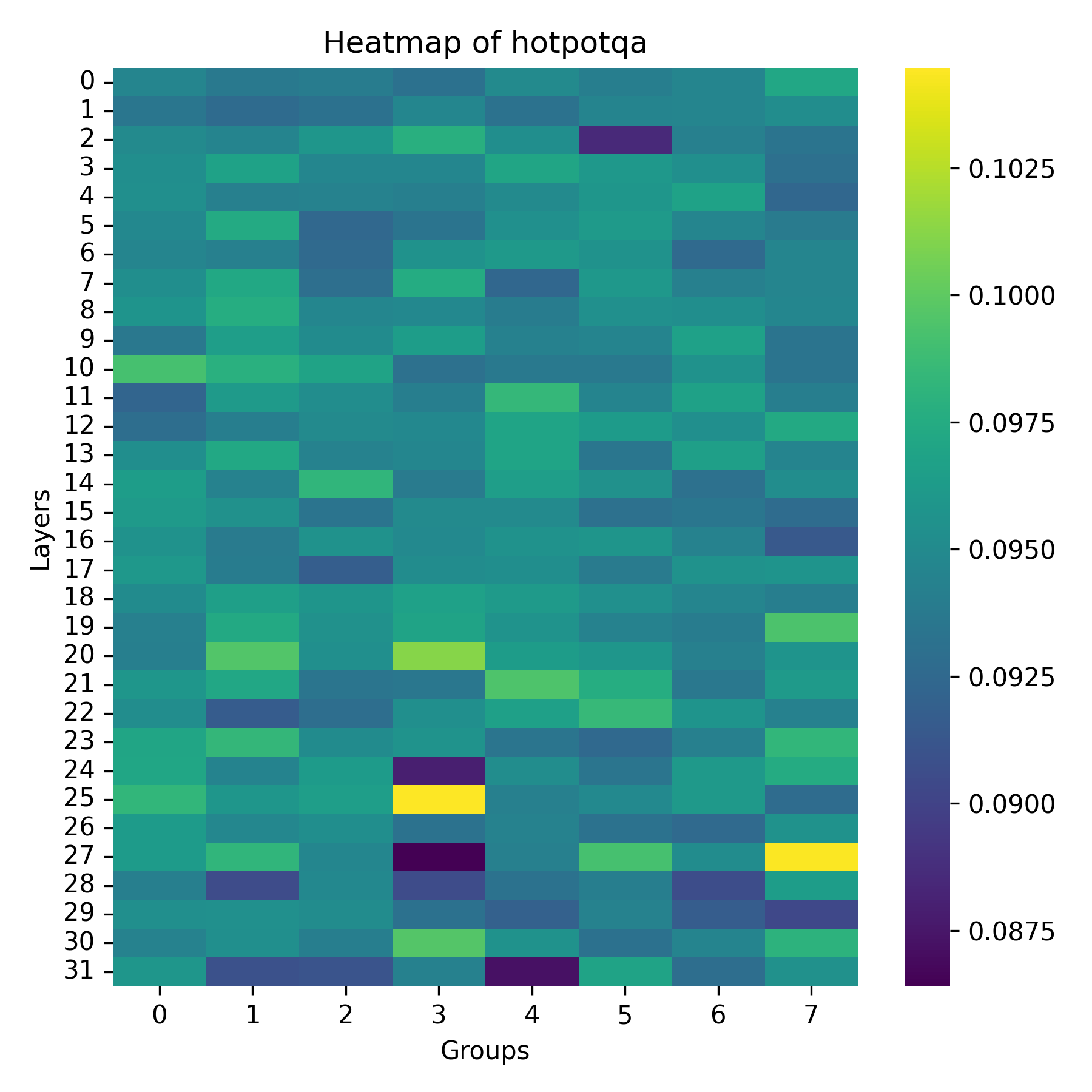}
			\caption* {(6) Coalition size 192}
			\label{cifar:closedsetnoise}
		\end{minipage}%
	}%
    \subfigure{
    	\begin{minipage}[b]{0.19\textwidth}
    		\includegraphics[width=1\textwidth]{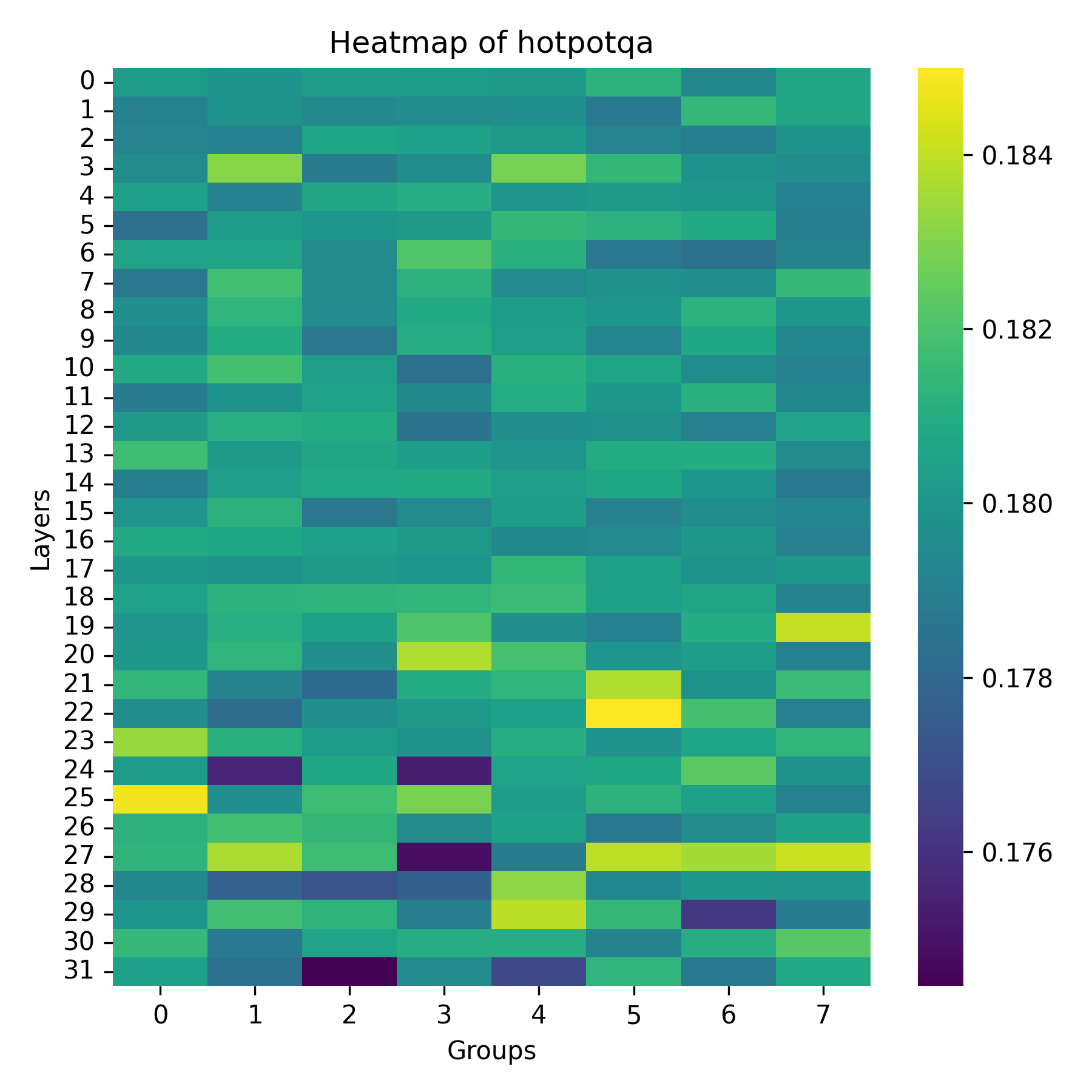}
    		\caption* {(7) Coalition size 224}
    		\label{cifar:datanoise}
    	\end{minipage}%
    }%
    \subfigure{
		\begin{minipage}[b]{0.19\textwidth}
			\includegraphics[width=1\textwidth]{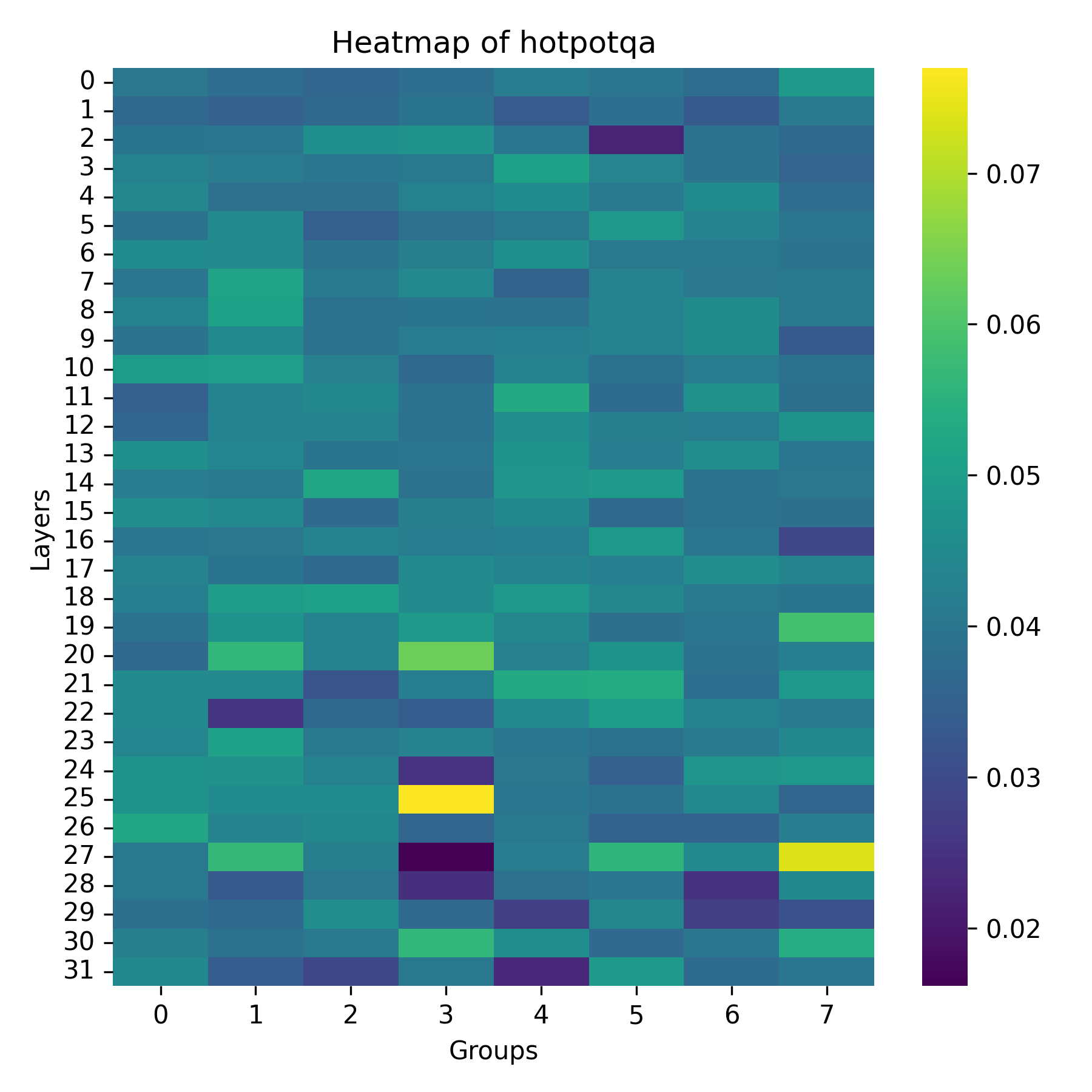}
			\caption* {(8) Average }
			\label{gradient:cifar}
		\end{minipage}%
	}%
 
\caption {The expected complementary contributions for the \emph{hotpotqa} dataset across different coalition sizes.}
\label{fig:heatmap_hotpotqa_coalition}
\end{figure*}

In Figures~\ref{fig:heatmap_lcc_coalition} and~\ref{fig:heatmap_hotpotqa_coalition}, we present the distributions of the expected complementary contributions of heads in Llama-3-8B-Instruct model on the \emph{hotpotqa} dataset (multi-document question answering) and the \emph{lcc} dataset (code generation), with coalition sizes of 
$\{32,64,96,128,160,192,224\}$. We observe strong correlations in the distributions across all coalition sizes. Additionally, the distributions of the expected complementary contributions for coalition sizes $S$ and $n-|S|$ are nearly identical, exhibiting symmetry around the size of 128. To optimize computational efficiency, we restrict the calculation of complementary contributions to coalitions with sizes below 128. These observations provide a justification for our approach of computing complementary contributions using only a small subset of coalition sizes, as it effectively captures the contributions of the heads.

\subsection{Generalization}
\label{section:generalization}
To validate the generalization capability of our method, we conduct cross-dataset evaluations on two task categories: 1. Multi-Document QA including 2WikiMQA and Musique datasets. 2. Code Processing including Lcc and RB-P datasets.

Following Section 4.2, we mask top and low-ranked attention heads but cross-apply head importance scores between datasets within the same task (e.g., mask 2WikiMQA using Musique-derived scores). As shown in Table~\ref{tab:generalization_llama} and Table~\ref{tab:generalization_mistral}, our method maintains superior accuracy over baselines across both models, confirming that learned importance scores can generalize across datasets within shared task domains.

\subsection{Needle-in-a-Haystack Test}\label{sec:needle_test}
To evaluate the performance of different KV cache compression methods in long-context retrieval tasks, we conduct a Needle-in-a-Haystack benchmark test using the Mistral-7B-v0.2 model. With the average KV cache size 128, we systematically insert target texts (needles) at ten equidistant positions (11\%, 22\%, ..., 100\%) across varying context lengths ranging from 1,000 to 31,000 tokens (in 1,000-token increments). Experimental results demonstrate that CoKV outperforms other baseline methods, achieving an average score of 95.89\% - the closest performance to the uncompressed FullKV benchmark.

\section{Proof}\label{sec:proof}
In this section, we give the proof of Theorem~\ref{theorem:SSV_sampling}. Denote $\mathcal{H}$ the selected coalition sizes.
The approximation of $\mathcal{SV}_{i,j}(1\leq i,j\leq n)$ is unbiased, which can be proven following Corollary 1 in~\cite{sun2024shapley}. So it is evident that $\overline{\mathcal{SSV}_{i}}$, being the weighted average of $\overline{\mathcal{SV}_{i,j}}$, serves as an unbiased estimator of $\mathcal{SSV}_{i}$.
Hence, we have 
\begin{align*}
    &\mathbb{P}(|\overline{\mathcal{SSV}_i^{\mathcal{H}}}-\mathcal{SSV}_i^{\mathcal{H}}|\geq \epsilon)\\
    \leq & \mathbb{P}(\sum_{j\in \mathcal{H}}|\overline{\mathcal{SV}_{i,j}}-\mathcal{SV}_{i,j}|\geq \epsilon) \\
    \leq & \sum_{j\in \mathcal{H}} \mathbb{P}(|\overline{\mathcal{SV}_{i,j}}-\mathcal{SV}_{i,j}|\geq \frac{\epsilon}{\mathcal{|H|}}) \\
\end{align*}
Then,we have
\begin{align*}
    &\sum_{j\in \mathcal{H}} \mathbb{P}(|\overline{\mathcal{SV}_{i,j}}-\mathcal{SV}_{i,j}|\geq \frac{\epsilon}{\mathcal{|H|}})\\
    \leq & 2|\mathcal{H}|\exp(-\frac{2(\frac{\epsilon}{|\mathcal{H}|})^2}{\sum_{k=1}^{\mathcal{M}/|\mathcal{H}|} (b_j-a_j)^2}) \\
    & \leq 2|\mathcal{H}|\exp(-\frac{2(\frac{\epsilon}{\mathcal{H}})^2}{\frac{\mathcal{M}r^2}{|\mathcal{H}|}}),
\end{align*}
according to Hoeffding's inequality where $(a_j,b_j)$ denotes the range of complementary contribution of j-coalitions, and r is $\max(b_1-a_1,\cdots,b_j-a_j)$. . Since we want the right hand side to be at most $\delta$, we have $\mathcal{M} \geq \frac{\mathcal{H}r^2ln\frac{2\mathcal{H}}{\delta}}{2\epsilon^2}$. Thus, Alogorithm~\ref{Alg:approSV} returns an $(\epsilon,\delta)$-approximation of Sliced Shapley value with time complexity $\mathcal{O}( \frac{T|\mathcal{H}|ln\frac{2|\mathcal{H}|}{\delta}}{\epsilon^2})$ where T is the time cost of evaluating each complementary contribution.
    The analysis of the time complexity of approximating Shapley value starts from $\mathbb{P}(|\overline{\mathcal{SV}_1}-\mathcal{SV}_i|\geq \epsilon)
    \leq \mathbb{P}(\sum_{j=1}^n|\overline{\mathcal{SV}_{i,j}}-\mathcal{SV}_{i,j}|\geq \epsilon)$
    Following similar steps, we can proof that the time complexity of approximating Shapley value is $\mathcal{O}( \frac{Tnln\frac{2n}{\delta}}{\epsilon^2})$.  
Thus, we complete the proof.

\begin{table*}[h]
\centering \caption{{Benchmark Results of Llama-3-8B-Instruct}} \label{tab:llama_results}
\begin{adjustbox}{width=\textwidth}
\begin{tabular}{lCCCCCCCCCCCCCCCCC}
\toprule
\multirow{2}{*}{Method} & \multicolumn{3}{c}{Single-Doc. QA} & \multicolumn{3}{c}{Multi-Doc. QA} & \multicolumn{3}{c}{Summarization} & \multicolumn{3}{c}{Few-shot Learning} & \multicolumn{2}{c}{Synthetic} & \multicolumn{2}{c}{Code} \\
\cmidrule(lr){2-4} \cmidrule(lr){5-7} \cmidrule(lr){8-10} \cmidrule(lr){11-13} \cmidrule(lr){14-15} \cmidrule(lr){16-17}
& \rotatebox{-45}{\small NtrQA} 
& \rotatebox{-45}{\small Qasper} 
& \rotatebox{-45}{\small MF-en} 
& \rotatebox{-45}{\small HotpotQA} 
& \rotatebox{-45}{\small 2WikiMQA} 
& \rotatebox{-45}{\small Musique} 
& \rotatebox{-45}{\small GovReport} 
& \rotatebox{-45}{\small QMSum} 
& \rotatebox{-45}{\small MultiNews} 
& \rotatebox{-45}{\small TREC} 
& \rotatebox{-45}{\small TriviaQA} 
& \rotatebox{-45}{\small SAMSum} 
& \rotatebox{-45}{\small PCount} 
& \rotatebox{-45}{\small PRe} 
& \rotatebox{-45}{\small Lcc} 
& \rotatebox{-45}{\small RB-P} \\
\midrule
Full Cache       & 24.12 & 31.24 & 39.85 & 45.23 & 34.56 & 21.09 & 28.38 & 23.24 & 26.52 & 74.12 & 90.96 & 42.37 & 4.55 & 71.76 & 58.1 & 51.64 \\
\bottomrule

\multicolumn{17}{c}{\textbf{KV size=64}} \\  
\midrule[0.5pt]

SnapKV           
&  19.94
&  13.21
&  28.91
&  40.06
&  28.58
&  18.12
&  17.29
&  21.71
&  17.05
&  49.41
&  89.00
&  35.48
&  3.99
&  71.57
&  54.35
&  50.42\\
Pyramid          
&  20.11
&  16.54
&  32.67
&  40.25
&  27.71
&  17.54
&  18.67
&  22.37
&  20.03
&  62.55
&  89.89
&  36.63
&  4.30
&  71.76
&  54.27
&  50.96\\
Ada-SnapKV       
&  20.40
&  14.46
&  32.62
&  42.39
&  31.48
&  17.58
&  18.57
&  22.18
&  18.71
&  58.82
&  90.13
&  35.25
&  4.41
&  71.57
&  54.02
&  51.68\\
HeadKV-R2        
&  20.30
&  16.76
&  35.96
&  38.08
&  26.41
&  17.98
&  18.68
&  21.75
&  20.58
&  67.06
&  88.19
&  37.30
&  3.21
&  71.76
&  56.20
&  54.49\\
CoKV             &20.77 & 19.67 & 35.11 & 44.37 & 34.36 & 17.83 & 17.89 & 22.33 & 18.55 & 71.76 & 90.73 & 38.51 & 4.71 &  71.76 & 55.45 & 55.82 \\
\bottomrule

\multicolumn{17}{c}{\textbf{KV size=128}} \\  
\midrule[0.5pt]

SnapKV           
&  20.37
&  14.73
&  34.24
&  43.32
&  28.94
&  19.74
&  19.68
&  22.15
&  20.68
&  64.71
&  90.69
&  39.03
&  4.41
&  71.76
&  58.48
&  51.70\\
Pyramid          
&  20.32
&  19.28
&  33.81
&  41.13
&  28.21
&  19.94
&  19.70
&  22.97
&  21.11
&  67.65
&  89.89
&  37.77
&  4.30
&  71.76
&  55.93
&  51.30\\
Ada-SnapKV       
&  20.86
&  18.14
&  35.17
&  45.12
&  30.39
&  20.43
&  19.93
&  21.84
&  21.25
&  69.41
&  90.29
&  38.08
&  4.75
&  71.76
&  57.99
&  53.16\\
HeadKV-R2        & 21.30 & 21.28 & 39.85 & 42.07 & 29.91 & 19.92 & 20.18 & 22.54 & 22.87 & 71.18 & 90.63 & 38.58 & 4.46 & 71.76 & 60.75 & 57.17 \\
CoKV             & 20.40 & 23.25 & 38.93 & 45.11 & 37.60 & 20.40 & 19.78 & 23.16 & 21.14 & 73.59 & 91.21 & 40.96 & 4.71 & 71.76 & 58.34 & 59.37 \\
\bottomrule

\multicolumn{17}{c}{\textbf{KV size=256}} \\  
\midrule[0.5pt]

SnapKV           
&  22.98
&  21.02
&  36.27
&  44.24
&  31.02
&  19.72
&  20.90
&  22.63
&  22.45
&  69.41
&  90.77
&  39.64
&  4.26
&  71.76
&  59.44
&  54.35\\
Pyramid          
&  22.18
&  22.83
&  35.95
&  41.85
&  31.74
&  21.14
&  21.27
&  22.65
&  22.83
&  71.18
&  90.83
&  40.50
&  4.35
&  71.37
&  57.69
&  51.49\\
Ada-SnapKV       
&  23.58
&  23.76
&  35.65
&  43.83
&  32.24
&  20.50
&  21.26
&  22.77
&  22.69
&  71.76
&  90.87
&  40.36
&  4.21
&  71.76
&  58.79
&  54.70\\
HeadKV-R2        
&  23.13
&  25.55
&  39.97
&  43.60
&  31.12
&  21.26
&  22.02
&  22.68
&  24.47
&  71.76
&  90.63
&  38.32
&  5.13
&  71.08
&  61.81
&  59.25\\
CoKV             & 22.69 & 28.23 & 42.34 & 46.32 & 36.38 & 21.17 & 21.17 & 23.64 & 23.08 & 72.94 & 90.93 & 42.07 & 4.71 & 71.76 & 62.40 & 61.92 \\
\bottomrule

\multicolumn{17}{c}{\textbf{KV size=512}} \\  
\midrule[0.5pt]

SnapKV           
&  22.92
&  22.86
&  39.33
&  43.89
&  32.70
&  20.87
&  22.24
&  22.39
&  23.97
&  71.18
&  90.87
&  41.14
&  4.54
&  71.76
&  59.98
&  55.00\\
Pyramid          
&  23.59
&  25.70
&  38.21
&  44.34
&  32.48
&  20.59
&  22.94
&  22.49
&  24.07
&  72.35
&  90.87
&  40.92
&  4.75
&  71.76
&  58.22
&  52.54\\
Ada-SnapKV       & 23.47 & 28.41 & 39.02 & 44.87 & 32.77 & 20.52 & 23.14 & 22.96 & 24.47 & 72.12 & 90.93 & 39.85 & 4.71 & 71.76 & 58.59 & 54.65 \\
HeadKV-R2        &22.52 & 29.32 & 40.34 & 45.64 & 34.52 & 20.53 & 23.92 & 22.61 & 25.73 & 72.35 & 90.93 &39.28 & 4.41 & 71.76 & 61.59 & 59.22 \\
CoKV             & 24.56 & 29.18 & 40.60 & 46.11 & 37.53 & 21.33 & 23.02 & 23.51 & 24.77 & 72.94 & 91.09 & 41.29 & 4.76 & 71.50 & 63.06 & 63.55 \\
\bottomrule

\multicolumn{17}{c}{\textbf{KV size=1024}} \\  
\midrule[0.5pt]

SnapKV          
&  23.95
&  26.95
&  37.81
&  44.03
&  30.88
&  20.93
&  24.26
&  23.09
&  25.79
&  72.35
&  90.87
&  41.43
&  4.31
&  71.76
&  59.29
&  54.91 \\
Pyramid          
& 23.62 & 26.76 & 39.44 & 45.79 & 33.41 & 19.87 & 23.57 & 22.98 & 25.13 & 73.02 & 90.93 & 40.86 & 4.71 & 71.76 & 58.43 & 53.67 \\
Ada-SnapKV       
&  23.52
&  28.33
&  40.39
&  45.20
&  32.95
&  20.11
&  24.55
&  23.33
&  25.37
&  73.53
&  90.87
&  41.38
&  4.46
&  71.76
&  58.88
&  54.65\\
HeadKV-R2        
&  23.35
&  29.60
&  40.09
&  45.82
&  35.81
&  21.39
&  25.57
&  23.32
&  26.30
&  74.12
&  90.77
&  40.27
&  4.19
&  71.76
&  61.58
&  59.03\\
CoKV             & 24.01 & 31.70 & 40.64 & 48.13 & 37.89 & 20.64 & 23.02 & 23.89 & 25.71 & 74.12 & 91.01 & 42.02 & 4.71 & 71.20 & 63.33 & 63.74 \\
\bottomrule

\end{tabular}
\vspace{0.5em}
\footnotesize
\end{adjustbox}
\end{table*}
\begin{table*}[h]
\centering  
\caption{{Results of Mistral-7B-Instruct-v0.2}}
\label{tab:mistral_results}
\begin{adjustbox}{width=\textwidth}
\begin{tabular}{lCCCCCCCCCCCCCCCCC}
\toprule
\multirow{2}{*}{Method} & \multicolumn{3}{c}{Single-Doc. QA} & \multicolumn{3}{c}{Multi-Doc. QA} & \multicolumn{3}{c}{Summarization} & \multicolumn{3}{c}{Few-shot Learning} & \multicolumn{2}{c}{Synthetic} & \multicolumn{2}{c}{Code} \\
\cmidrule(lr){2-4} \cmidrule(lr){5-7} \cmidrule(lr){8-10} \cmidrule(lr){11-13} \cmidrule(lr){14-15} \cmidrule(lr){16-17}
& \rotatebox{-45}{\small NtrQA} 
& \rotatebox{-45}{\small Qasper} 
& \rotatebox{-45}{\small MF-en} 
& \rotatebox{-45}{\small HotpotQA} 
& \rotatebox{-45}{\small 2WikiMQA} 
& \rotatebox{-45}{\small Musique} 
& \rotatebox{-45}{\small GovReport} 
& \rotatebox{-45}{\small QMSum} 
& \rotatebox{-45}{\small MultiNews} 
& \rotatebox{-45}{\small TREC} 
& \rotatebox{-45}{\small TriviaQA} 
& \rotatebox{-45}{\small SAMSum} 
& \rotatebox{-45}{\small PCount} 
& \rotatebox{-45}{\small PRe} 
& \rotatebox{-45}{\small Lcc} 
& \rotatebox{-45}{\small RB-P} \\
\midrule
Full Cache       
&  26.40
&  31.07
&  49.38
&  37.60
&  26.07
&  17.81
&  31.87
&  23.16
&  27.15
&  70.59
&  85.73
&  43.26
&  1.52
&  58.52
&  55.10
&  49.45 \\
\bottomrule

\multicolumn{17}{c}{\textbf{KV size=64}} \\  
\midrule[0.5pt]

SnapKV           
&  16.99
&  18.26
&  38.29
&  29.51
&  23.24
&  13.46
&  18.24
&  20.48
&  18.05
&  48.82
&  81.45
&  36.18
&  2.54
&  43.79
&  46.13
&  39.30\\
Pyramid          
&  17.51
&  18.60
&  40.49
&  31.92
&  22.08
&  13.81
&  18.68
&  20.94
&  18.80
&  57.06
&  81.71
&  37.42
&  1.68
&  46.23
&  46.05
&  40.03\\
Ada-SnapKV       
&  17.93
&  18.68
&  40.03
&  29.99
&  22.67
&  14.92
&  18.84
&  20.87
&  18.53
&  54.12
&  81.43
&  37.25
&  2.30
&  45.20
&  46.84
&  39.37\\
HeadKV-R2       
&  22.75
&  25.37
&  45.36
&  36.52
&  25.39
&  13.82
&  20.45
&  22.06
&  21.48
&  65.29
&  83.56
&  37.95
&  2.43
&  50.78
&  47.76
&  42.86\\
CoKV             & 21.07 & 21.41 & 42.87 & 37.74 & 28.93 & 15.60 & 18.03 & 21.08 & 19.70 & 67.65 & 86.52 & 39.54 & 3.68 & 54.22 & 49.20 & 42.13 \\
\bottomrule

\multicolumn{17}{c}{\textbf{KV size=128}} \\  
\midrule[0.5pt]

SnapKV     
&  23.02
&  20.73
&  41.91
&  31.39
&  22.88
&  14.55
&  20.92
&  21.83
&  21.25
&  62.35
&  83.21
&  38.99
&  3.14
&  51.16
&  49.94
&  43.61\\
Pyramid          
&  22.06
&  21.82
&  43.73
&  32.33
&  24.12
&  13.80
&  20.27
&  21.65
&  21.34
&  65.29
&  83.78
&  38.37
&  2.63
&  53.59
&  49.21
&  42.69\\
Ada-SnapKV       
&  22.32
&  22.71
&  44.40
&  32.63
&  23.29
&  13.79
&  21.15
&  22.50
&  21.77
&  66.47
&  84.28
&  39.68
&  3.04
&  51.87
&  49.57
&  44.84\\
HeadKV-R2        
&  24.81
&  27.66
&  48.29
&  36.87
&  26.66
&  14.75
&  23.30
&  22.88
&  23.26
&  67.65
&  84.93
&  39.75
&  2.50
&  49.31
&  50.79
&  45.57\\
CoKV             & 24.42 & 24.12 & 46.95 & 38.28 & 28.85 & 17.18 & 21.11 & 21.91 & 22.02 & 68.82 & 86.14 & 40.48 & 4.21 & 54.12 & 51.08 & 46.25 \\
\bottomrule

\multicolumn{17}{c}{\textbf{KV size=256}} \\  
\midrule[0.5pt]

SnapKV           
&  23.01
&  23.47
&  45.38
&  33.15
&  24.12
&  13.93
&  22.80
&  22.89
&  22.85
&  67.65
&  84.62
&  40.39
&  2.36
&  59.18
&  51.34
&  46.74\\
Pyramid   
& 22.98
& 25.66
& 46.12
& 34.47
& 25.81
& 13.98
& 22.86
& 22.54
& 22.88
& 68.90
& 85.07
& 40.92
& 2.39
& 58.74
& 53.13
& 46.59\\
Ada-SnapKV       
&  23.54
&  26.02
&  45.92
&  34.45
&  26.09
&  14.12
&  22.79
&  22.64
&  23.32
&  68.82
&  85.32
&  41.93
&  2.04
&  58.62
&  52.10
&  47.70\\
HeadKV-R2        
&  25.40
&  27.42
&  47.05
&  37.98
&  25.57
&  17.08
&  25.31
&  22.72
&  25.03
&  69.41
&  84.93
&  40.24
&  2.58
&  52.94
&  53.48
&  49.21\\
CoKV             & 25.70 & 26.10 & 48.43 & 38.96 & 30.06 & 17.33 & 23.42 & 22.55 & 23.73 & 70.00 & 86.19 & 42.35 & 3.65 & 56.37 & 53.97 & 48.79 \\
\bottomrule

\multicolumn{17}{c}{\textbf{KV size=512}} \\  
\midrule[0.5pt]

SnapKV           
&  25.24
&  26.30
&  47.85
&  37.16
&  25.07
&  14.57
&  24.43
&  22.98
&  24.61
&  68.82
&  85.72
&  43.04
&  2.00
&  58.63
&  54.06
&  49.03\\
Pyramid          
&  24.43
&  27.09
&  48.49
&  37.57
&  25.35
&  16.20
&  24.40
&  22.85
&  24.16
&  68.82
&  85.81
&  42.07
&  1.87
&  56.93
&  53.05
&  48.22\\
Ada-SnapKV       
&  25.01
&  26.76
&  49.10
&  37.12
&  26.68
&  15.63
&  24.42
&  22.94
&  24.61
&  69.41
&  85.56
&  41.88
&  1.87
&  57.93
&  54.09
&  48.94\\
HeadKV-R2        
&  25.80
&  28.73
&  48.34
&  37.43
&  27.03
&  17.28
&  28.22
&  23.22
&  26.65
&  70.59
&  85.72
&  40.15
&  2.69
&  56.15
&  53.24
&  49.22\\
CoKV        & 25.25 & 28.13 & 49.91 & 38.87 & 32.33 & 18.27 & 25.00 & 23.08 & 25.50 & 70.59 & 86.37 &  43.46 & 3.06 & 59.20 & 55.54 & 49.38 \\
\bottomrule

\multicolumn{17}{c}{\textbf{KV size=1024}} \\  
\midrule[0.5pt]

SnapKV           
&  26.38
&  29.70
&  48.13
&  37.36
&  25.52
&  16.88
&  27.31
&  22.63
&  26.10
&  69.41
&  85.72
&  42.43
&  1.54
&  56.87
&  55.05
&  49.33 \\
Pyramid        
&  25.09
&  28.59
&  47.78
&  37.74
&  25.83
&  17.53
&  25.88
&  23.05
&  25.91
&  68.24
&  85.95
&  42.77
&  1.59
&  57.82
&  54.47
&  48.85 \\
Ada-SnapKV       
&  25.70
&  29.95
&  47.50
&  37.68
&  26.18
&  17.10
&  26.63
&  22.93
&  26.10
&  70.00
&  85.72
&  43.16
&  1.68
&  56.28
&  54.52
&  49.10\\
HeadKV-R2      
&  27.48
&  29.94
&  49.49
&  37.49
&  26.45
&  18.69
&  30.73
&  23.31
&  26.74
&  70.59
&  85.92
&  42.05
&  3.15
&  56.37
&  54.73
&  49.30\\
CoKV             & 26.15 & 29.82 & 49.47 & 38.54 & 34.39 & 17.98 & 27.76 &  23.33&  26.49 &  70.59 & 86.23 & 43.54 & 2.48 & 59.32 & 55.47 & 49.92 \\
\bottomrule

\end{tabular}
\footnotesize
\end{adjustbox}
\end{table*}
\begin{table*}[h]
\centering 
\caption{{Results of masking groups with Llama-3-8B-Instruct}}
\label{tab:mask_llama}
\begin{adjustbox}{width=\textwidth}
\begin{tabular}{lCCCCCCCCCCCCCCCCC}
\toprule
\multirow{2}{*}{Method} & \multicolumn{3}{c}{Single-Doc. QA} & \multicolumn{3}{c}{Multi-Doc. QA} & \multicolumn{3}{c}{Summarization} & \multicolumn{3}{c}{Few-shot Learning} & \multicolumn{2}{c}{Synthetic} & \multicolumn{2}{c}{Code} \\
\cmidrule(lr){2-4} \cmidrule(lr){5-7} \cmidrule(lr){8-10} \cmidrule(lr){11-13} \cmidrule(lr){14-15} \cmidrule(lr){16-17}
& \rotatebox{-45}{\small NtrQA} 
& \rotatebox{-45}{\small Qasper} 
& \rotatebox{-45}{\small MF-en} 
& \rotatebox{-45}{\small HotpotQA} 
& \rotatebox{-45}{\small 2WikiMQA} 
& \rotatebox{-45}{\small Musique} 
& \rotatebox{-45}{\small GovReport} 
& \rotatebox{-45}{\small QMSum} 
& \rotatebox{-45}{\small MultiNews} 
& \rotatebox{-45}{\small TREC} 
& \rotatebox{-45}{\small TriviaQA} 
& \rotatebox{-45}{\small SAMSum} 
& \rotatebox{-45}{\small PCount} 
& \rotatebox{-45}{\small PRe} 
& \rotatebox{-45}{\small Lcc} 
& \rotatebox{-45}{\small RB-P} \\
\midrule
Full Cache       & 24.12 & 31.24 & 39.85 & 45.23 & 34.56 & 21.09 & 28.38 & 23.24 & 26.52 & 74.12 & 90.96 & 42.37 & 4.55 & 71.76 & 58.10 & 51.64 \\
\bottomrule

\multicolumn{17}{c}{\textbf{Masking 16 groups}} \\  
\midrule[0.5pt]

Random
&  20.93
&  28.48
&  33.69
&  44.93
&  20.01
&  20.6
&  28.43
&  23.7
&  26.67
&  74.12
&  91.07
&  41.12
&  4.26
&  71.76
&  49.83
&  40.55\\

HeadKV-R2(top)
&  19.45
&  12.97
&  27.75
&  34.2
&  17.33
&  14.32
&  19.74
&  22.76
&  22.05
&  67.06
&  87.91
&  35.53
&  4.71
&  68.49
&  26.62
&  26.53\\

CoKV(top)
&  6.55
&  9.46
&  9.47
&  10.19
&  12.27
&  5.67
&  5.73
&  16.96
&  4.47
&  43.53
&  71.21
&  23.77
&  3.91
&  34.98
&  11.58
&  17.18\\

HeadKV-R2(low)
&  21.83
&  14.36
&  33.34
&  31.37
&  27.23
&  12.55
&  27.29
&  23.82
&  26.99
&  74.12
&  91.03
&  42.18
&  4.12
&  70.59
&  37.35
&  38.55\\

CoKV(low)
&  23.74
&  33.76
&  41.71
&  49.27
&  40.48
&  19.99
&  29.13
&  23.25
&  27.79
&  74.12
&  91.45
&  42.37
&  4.71
&  70.55
&  63.38
&  61.26\\
\bottomrule

\multicolumn{17}{c}{\textbf{Masking 32 groups}} \\  
\midrule[0.5pt]

Random           
&  20.69
&  18.60
&  29.63
&  39.12
&  18.50
&  6.94
&  22.40
&  22.33
&  26.45
&  74.12
&  89.82
&  33.80
&  4.71
&  61.12
&  30.78
&  40.71\\
HeadKV-R2(top)         
&  17.33
&  6.98
&  9.37
&  13.50
&  9.37
&  5.11
&  13.18
&  20.86
&  15.24
&  45.88
&  75.30
&  27.21
&  4.76
&  66.21
&  11.24
&  13.64\\
CoKV(top)             
&  1.40
&  3.49
&  3.78
&  7.94
&  9.32
&  2.32
& 2.64
& 11.74
&  0.58
&  34.71
&  21.37
&  6.96
& 4.14
&  16.93
&  3.54
&  5.17\\

HeadKV-R2(low)
&  21.51
&  11.16
&  25.33
&  19.52
&  14.48
&  7.42
&  16.73
&  23.91
&  14.58
&  74.12
&  89.09
&  40.69
&  4.66
&  70.09
&  33.13
&  32.39\\ 

CoKV(low)&  22.45
&  33.06
&  38.34
&  45.82
&  39.62
&  20.18
& 28.39
& 24.04
&  26.67
&  74.12
&  91.14
&  41.70
& 4.71
&  71.76
&  52.24
&  64.94 \\

\bottomrule

\multicolumn{17}{c}{\textbf{Masking 64 groups}} \\  
\midrule[0.5pt]

Random           
&  13.22
&  7.34
&  20.57
&  20.58
&  9.11
&  6.76
&  7.50
&  21.22
&  19.18
&  72.35
&  71.92
&  36.09
&  4.71
&  52.80
&  21.27
&  18.07\\
HeadKV-R2(top)         
&  7.49
&  2.95
&  5.05
&  11.06
&  12.01
&  2.46
&  3.63
&  14.43
&  5.06
&  34.71
&  48.92
&  8.05
&  3.97
&  70.67
&  21.03
&  16.14\\
CoKV (top)           
&  0.76
&  1.76
&  2.45
&  4.85
&  5.58
&  1.93
& 2.48
& 5.65
&  0.20
&  34.12
&  3.33
&  7.34
& 3.16
&  12.18
&  2.45
&  3.83\\

HeadKV-R2(low) 
&  19.23
&  12.19
&  21.33
&  19.61
&  14.21
&  6.63
&  6.45
&  20.17
&  6.16
&  71.76
&  77.40
&  31.52
&  4.41
&  53.48
&  16.00
&  14.58
\\ 

CoKV(low) 
&  21.98
&  29.85
&  38.95
&  44.21
&  36.65
&  17.71
& 28.04
& 24.49
&  25.92
&  74.71
&  91.66
&  40.80
&  4.54
&  71.76
&  47.04
&  52.77\\

\bottomrule

\multicolumn{17}{c}{\textbf{Masking 96 groups}} \\  
\midrule[0.5pt]

Random           
&  5.19
&  4.04
&  6.85
&  8.15
&  10.33
&  5.08
&  2.21
&  10.77
&  2.82
&  40.00
&  61.54
&  13.38
&  4.64
&  54.29
&  15.37
&  9.81\\
HeadKV-R2(top)           
&  2.89
&  4.34
&  7.90
&  11.83
&  9.14
&  2.93
&  4.37
&  13.21
&  3.80
&  34.12
&  30.32
&  8.46
&  4.78
&  71.76
&  13.55
&  14.76\\
CoKV(top)      
&  1.36
&  1.14
&  1.82
&  3.66
&  3.79
&  1.48
& 1.20
& 4.63
&  0.13
&  34.12
&  2.40
&  7.52
& 0.54
&  6.71
&  2.41
&  3.54\\

HeadKV-R2(low) 
&  19.28
&  8.23
&  15.65
&  20.89
&  16.80
&  8.00
&  3.32
&  11.81
&  0.99
&  58.82
&  58.70
&  15.72
&  4.71
&  61.88
&  10.56
&  11.05\\

CoKV(low) 
&  20.24
&  18.97
&  35.28
&  41.37
&  30.02
&  13.87
& 19.95
& 17.33
&  20.76
&  74.71
&  84.08
&  41.23
& 4.71
&  68.24
&  38.11
&  38.08\\

\bottomrule

\multicolumn{17}{c}{\textbf{Masking 128 groups}} \\  
\midrule[0.5pt]

Random           
&  3.34
&  2.50
&  5.33
&  10.59
&  5.12
&  2.73
&  2.15
&  9.19
&  0.16
&  44.12
&  31.33
&  9.05
&  4.18
&  66.74
&  12.27
&  9.23\\
HeadKV-R2(top)           
&  2.34
&  2.17
&  5.38
&  7.21
&  7.19
&  1.85
&  1.80
&  10.34
&  0.31
&  34.71
&  26.08
&  7.87
&  4.71
&  66.92
&  13.94
&  11.76\\
CoKV(top)             
&  0.59
&  0.80
&  1.38
&  2.96
&  3.42
&  1.11
& 1.16
& 4.05
&  0.13
&  34.12
&  2.89
&  7.17
& 1.09
&  7.52
&  2.91
&  3.55\\

HeadKV-R2(low) 
&  12.02
&  7.97
&  8.92
&  14.87
&  12.83
&  5.26
&  2.41
&  9.12
&  1.42
&  55.88
&  40.96
&  10.2
&  4.71
&  68.42
&  10.14
&  6.03\\

CoKV(low) 
&  15.31
&  12.15
&  28.44
&  35.35
&  23.27
&  10.67
& 2.93
& 12.24
&  9.41
&  73.82
&  76.32
&  37.70
& 4.71
&  68.24
&  22.20
&  24.93\\

\bottomrule

\end{tabular}
\vspace{0.5em}
\footnotesize
\end{adjustbox}
\end{table*}
\begin{table*}[h]
\centering
\caption{{Results of masking groups with Mistral-7B-Instruct-v0.2}}
\label{tab:mask_mistral}
\begin{adjustbox}{width=\textwidth}
\begin{tabular}{lCCCCCCCCCCCCCCCCC}
\toprule
\multirow{2}{*}{Method} & \multicolumn{3}{c}{Single-Doc. QA} & \multicolumn{3}{c}{Multi-Doc. QA} & \multicolumn{3}{c}{Summarization} & \multicolumn{3}{c}{Few-shot Learning} & \multicolumn{2}{c}{Synthetic} & \multicolumn{2}{c}{Code} \\
\cmidrule(lr){2-4} \cmidrule(lr){5-7} \cmidrule(lr){8-10} \cmidrule(lr){11-13} \cmidrule(lr){14-15} \cmidrule(lr){16-17}
& \rotatebox{-45}{\small NtrQA} 
& \rotatebox{-45}{\small Qasper} 
& \rotatebox{-45}{\small MF-en} 
& \rotatebox{-45}{\small HotpotQA} 
& \rotatebox{-45}{\small 2WikiMQA} 
& \rotatebox{-45}{\small Musique} 
& \rotatebox{-45}{\small GovReport} 
& \rotatebox{-45}{\small QMSum} 
& \rotatebox{-45}{\small MultiNews} 
& \rotatebox{-45}{\small TREC} 
& \rotatebox{-45}{\small TriviaQA} 
& \rotatebox{-45}{\small SAMSum} 
& \rotatebox{-45}{\small PCount} 
& \rotatebox{-45}{\small PRe} 
& \rotatebox{-45}{\small Lcc} 
& \rotatebox{-45}{\small RB-P} \\
\midrule
Full Cache       
&  26.40
&  31.07
&  49.38
&  37.60
&  26.07
&  17.81
&  31.87
&  23.16
&  27.15
&  70.59
&  85.73
&  43.26
&  1.52
&  58.52
&  55.10
&  49.45\\
\bottomrule

\multicolumn{17}{c}{\textbf{Masking 16 groups}} \\  
\midrule[0.5pt]

Random 
&  25.92
&  31.73
&  50.29
&  37.84
&  27.19
&  17.83
&  24.91
&  21.92
&  27.04
&  70.59
&  85.93
&  43.8
&  3.22
&  53.82
&  52.38
&  48.24\\

HeadKV-R2(top)

&  23.38
&  16.66
&  37.13
&  37.41
&  22.76
&  14.29
&  18.8
&  21.74
&  23.23
&  54.12
&  82.96
&  35.22
&  4.12
&  21.76
&  39.49
&  35.66\\

CoKV(top)
&  16.1
&  23.35
&  18.49
&  14.34
&  13.39
&  7.89
&  20.5
&  19.98
&  17.25
&  38.24
&  52.51
&  26.32
&  4.17
&  40.85
&  24.6
&  14.35\\

HeadKV-R2(low)
&  24.78
&  29.37
&  48.78
&  38.07
&  24.88
&  16.93
&  31.25
&  23.08
&  27.64
&  71.18
&  84.55
&  42.52
&  2.1
&  58.82
&  54.22
&  49.4\\

CoKV(low)
&  26.57
&  32.3
&  49.94
&  40.38
&  34.0
&  19.11
&  31.25
&  22.97
&  26.85
&  70.59
&  87.3
&  44.39
&  3.29
&  58.03
&  56.6
&  50.74\\

\bottomrule

\multicolumn{17}{c}{\textbf{Masking 32 groups}} \\  
\midrule[0.5pt]

Random   
&  22.62
&  31.72
&  47.20
&  38.13
&  22.55
&  11.92
&  25.64
&  23.27
&  26.75
&  68.82
&  84.55
&  41.34
&  1.93
&  49.71
&  50.14
&  47.18\\
HeadKV-R2(top)  
&  20.82
&  15.40
&  28.72
&  34.31
&  20.31
&  12.86
&  13.56
&  19.83
&  17.80
&  46.47
&  79.25
&  30.10
&  4.71
&  24.31
&  33.41
&  30.47\\
CoKV(top)  &  9.05
& 15.38
&  7.61
&  9.88
& 8.07
& 6.38
& 0.59
&  11.72
&  4.70
&  35.88
&  26.87
&  11.85
& 4.65
& 10.88
&  15.23
&  11.14     \\

HeadKV-R2(low)
&  23.76
&  27.40
&  44.80
&  32.85
&  23.55
&  13.28
&  24.37
&  22.71
&  28.09
&  71.18
&  79.24
&  42.24
&  4.26
&  49.90
&  52.89
&  48.85\\ 
CoKV(low)
&  26.70
& 30.44
&  49.57
&  40.41
& 32.28
& 18.33
& 30.26
&  23.27
&  26.85
&  70.59
&  87.48
&  44.04
& 2.93
& 56.27
&  56.34
&  50.38\\

\bottomrule

\multicolumn{17}{c}{\textbf{Masking 64 groups}} \\  
\midrule[0.5pt]

Random   
&  13.43
&  24.46
&  30.97
&  22.62
&  16.93
&  15.65
&  14.07
&  22.16
&  19.86
&  55.29
&  82.16
&  35.85
&  4.12
&  38.94
&  38.07
&  28.39\\
HeadKV-R2(top) 
&  11.04
&  9.09
&  17.45
&  18.57
&  13.79
&  8.07
&  9.83
&  17.30
&  12.60
&  35.29
&  55.36
&  18.65
&  4.54
&  19.85
&  26.25
&  21.23\\
CoKV(top) 
&  3.28
& 3.50
&  4.65
&  4.30
& 3.42
& 2.55
& 0.79
&  4.66
&  1.08
&  34.71
&  8.41
&  6.00
& 3.53
& 3.53
&  11.22
&  11.57       \\

HeadKV-R2(low) 
&  18.81
&  21.42
&  35.18
&  18.03
&  14.26
&  7.41
&  22.56
&  22.41
&  20.24
&  57.65
&  75.72
&  37.03
&  4.11
&  45.46
&  38.78
&  39.22\\ 
CoKV(low) 
&  26.87
& 25.74
&  48.19
&  39.61
& 30.86
& 16.88
& 24.45
&  22.84
&  27.29
&  71.18
&  87.16
&  43.43
& 3.34
& 50.18
&  53.76
&  47.52\\

\bottomrule

\multicolumn{17}{c}{\textbf{Masking 96 groups}} \\  
\midrule[0.5pt]

Random    
&  4.84
&  6.33
&  13.77
&  12.00
&  10.41
&  8.43
&  0.88
&  17.55
&  21.83
&  51.76
&  63.48
&  22.32
&  4.47
&  34.19
&  21.30
&  17.65\\
HeadKV-R2(top) 
&  9.21
&  7.05
&  11.34
&  13.30
&  14.22
&  3.99
&  7.67
&  15.43
&  8.84
&  34.71
&  29.87
&  9.97
&  4.44
&  30.16
&  17.73
&  16.24\\
CoKV(top)  
&  2.13
& 4.13
&  4.58
&  4.09
& 6.52
& 0.64
& 0.00
&  2.44
&  0.15
&  34.71
&  2.16
&  4.40
& 4.12
& 2.94
&  7.16
&  8.39\\

HeadKV-R2(low)
&  8.17
&  10.62
&  18.76
&  13.07
&  10.10
&  5.44
&  3.75
&  19.42
&  6.51
&  46.47
&  50.84
&  23.98
&  4.57
&  29.89
&  34.95
&  32.57\\ 
CoKV(low)
&  24.62
& 24.71
&  48.04
&  38.72
& 30.29
& 16.37
& 19.35
&  22.84
&  27.18
&  70.59
&  79.48
&  42.01
& 3.75
& 48.29
&  50.78
&  43.53\\

\bottomrule

\multicolumn{17}{c}{\textbf{Masking 128 groups}} \\  
\midrule[0.5pt]

Random   
&  4.15
&  8.45
&  9.73
&  8.38
&  7.80
&  2.07
&  0.51
&  13.19
&  3.40
&  42.94
&  34.04
&  8.82
&  3.85
&  3.53
&  23.74
&  18.34\\
HeadKV-R2(top)   
&  5.22
&  4.78
&  8.63
&  7.04
&  6.15
&  3.89
&  5.64
&  14.59
&  5.64
&  35.88
&  25.98
&  8.36
&  3.82
&  18.53
&  18.68
&  18.52\\
CoKV(top)
&  1.33
& 9.43
&  1.03
&  4.24
& 5.54
& 1.41
& 0.09
&  0.78
&  0.01
&  33.53
&  1.06
&  4.50
& 2.94
& 2.94
&  6.94
&  6.22\\
HeadKV-R2(low)
&  4.41
&  4.53
&  11.12
&  12.8
&  7.20
&  6.64
&  0.46
&  10.48
&  0.61
&  47.65
&  31.61
&  10.45
&  2.91
&  9.92
&  24.09
&  24.48\\ 
CoKV(low) 
&  20.43
& 19.12
&  44.82
&  34.23
& 23.31
& 13.97
& 14.22
&  21.28
&  24.65
&  70.59
&  73.98
&  39.73
& 4.10
& 45.21
&  42.14
&  38.14\\

\bottomrule

\end{tabular}
\footnotesize
\end{adjustbox}
\end{table*}
\begin{center}
\begin{small}
\begin{table*}[ht]
\caption{Time and MAE of the Sliced Shapley values estimation.}\label{table:qasper_mse}
\centering
\begin{tabular}{|*{13}{c|}} 
\bottomrule
Sample Num &  50 &  100 &  150 &  200 &  250 &  300 &  350 &  400 &  450 &  500 \\
\bottomrule
    Time 
&  4.19
&  8.37
&  12.56
&  16.75
&  20.93
&  25.12
&  29.3
&  33.49
&  37.68
&  41.86
 \\
MAE 
&  8.2e-3
&  5.3e-3
&  4.8e-3
&  4.8e-3
&  3.8e-3
&  3.4e-3
&  3.2e-3
&  2.9e-3
&  2.8e-3
&  2.8e-3\\
\bottomrule
\end{tabular}
\end{table*}
\end{small}
\end{center}

\begin{table}[h]
\centering 
\caption{Generalization results of masking groups with Llama3-8B-Instruct}
\label{tab:generalization_llama}
\begin{adjustbox}{width=0.5\textwidth}
\begin{tabular}{lCCCC}
\toprule
\multirow{2}{*}{Method} & \multicolumn{2}{c}{Multi-Doc. QA} & \multicolumn{2}{c}{Code}  \\
\cmidrule(lr){2-3} \cmidrule(lr){4-5} 
& \rotatebox{-45}{\small 2WikiMQA} 
& \rotatebox{-45}{\small Musique} 
& \rotatebox{-45}{\small Lcc} 
& \rotatebox{-45}{\small RB-P}  \\
\midrule
Full Cache       & 34.56 & 21.09 & 58.10 & 51.64 \\
\bottomrule

\multicolumn{5}{c}{\textbf{Masking 16 groups}} \\  
\midrule[0.5pt]

Random
&  20.01
&  20.6
&  49.83
&  40.55\\

HeadKV-R2(top)
&  17.33
&  14.32
&  26.62
&  26.53\\

CoKV(top)
& 10.78
& 5.43
& 14.41
& 15.33  \\

HeadKV-R2(low)
&  27.23
&  12.55
&  37.35
&  38.55\\

CoKV(low)
& 39.92
& 20.9
& 64.04
& 61.22  \\
\bottomrule

\multicolumn{5}{c}{\textbf{Masking 32 groups}} \\  
\midrule[0.5pt]

Random           
&  18.50
&  6.94
&  30.78
&  40.71\\
HeadKV-R2(top)         
&  9.37
&  5.11
&  11.24
&  13.64\\
CoKV(top)
& 6.71
& 3.45
& 4.39
& 5.78  \\

HeadKV-R2(low)
&  14.48
&  7.42
&  33.13
&  32.39\\ 

CoKV(low)
& 38.1
& 18.22
& 64.75
& 58.28  \\

\bottomrule

\multicolumn{5}{c}{\textbf{Masking 64 groups}} \\  
\midrule[0.5pt]

Random           
&  9.11
&  6.76
&  21.27
&  18.07\\
HeadKV-R2(top)         
&  12.01
&  2.46
&  21.03
&  16.14\\
CoKV(top)
& 5.68
& 1.82
& 2.5
& 3.66  \\

HeadKV-R2(low) 
&  14.21
&  6.63
&  16.00
&  14.58
\\ 

CoKV(low)
& 34.17
& 16.29
& 49.97
& 48.93  \\

\bottomrule

\multicolumn{5}{c}{\textbf{Masking 96 groups}} \\  
\midrule[0.5pt]

Random           
&  10.33
&  5.08
&  15.37
&  9.81\\
HeadKV-R2(top)           
&  9.14
&  2.93
&  13.55
&  14.76\\
CoKV(top)
& 4.38
& 1.28
& 2.74
& 3.07  \\

HeadKV-R2(low) 
&  16.80
&  8.00
&  10.56
&  11.05\\

CoKV(low)
& 28.08
& 12.92
& 38.62
& 40.55  \\

\bottomrule

\multicolumn{5}{c}{\textbf{Masking 128 groups}} \\  
\midrule[0.5pt]

Random           
&  5.12
&  2.73
&  12.27
&  9.23\\
HeadKV-R2(top)           
&  7.19
&  1.85
&  13.94
&  11.76\\
CoKV(top)
& 2.93
& 0.94
& 2.48
& 3.84  \\

HeadKV-R2(low) 
&  12.83
&  5.26
&  10.14
&  6.03\\

CoKV(low)
& 24.34
& 9.37
& 23.38
& 24.11  \\

\bottomrule

\end{tabular}
\footnotesize
\end{adjustbox}
\end{table}
\begin{table}[h]
\centering 
\caption{Generalization results of masking groups with Mistral-7B-v0.2}
\label{tab:generalization_mistral}
\begin{adjustbox}{width=0.5\textwidth}
\begin{tabular}{lCCCC}
\toprule
\multirow{2}{*}{Method} & \multicolumn{2}{c}{Multi-Doc. QA} & \multicolumn{2}{c}{Code}  \\
\cmidrule(lr){2-3} \cmidrule(lr){4-5} 
& \rotatebox{-45}{\small 2WikiMQA} 
& \rotatebox{-45}{\small Musique} 
& \rotatebox{-45}{\small Lcc} 
& \rotatebox{-45}{\small RB-P}  \\
\midrule
Full Cache       
&  26.07
&  17.81
&  55.10
&  49.45\\
\bottomrule

\multicolumn{5}{c}{\textbf{Masking 16 groups}} \\  
\midrule[0.5pt]

Random 
&  27.19
&  17.83
&  52.38
&  48.24\\

HeadKV-R2(top)

&  22.76
&  14.29
&  39.49
&  35.66\\

CoKV(top)
& 13.02
& 6.99
& 17.97
& 23.38  \\

HeadKV-R2(low)
&  24.88
&  16.93
&  54.22
&  49.4\\

CoKV(low)
& 26.25
& 18.18
& 54.58
& 50.03  \\

\bottomrule

\multicolumn{5}{c}{\textbf{Masking 32 groups}} \\  
\midrule[0.5pt]

Random   
&  22.55
&  11.92
&  50.14
&  47.18\\
HeadKV-R2(top)  
&  20.31
&  12.86
&  33.41
&  30.47\\
CoKV(top)
& 10.23
& 5.16
& 11.8
& 13.64  \\

HeadKV-R2(low)
&  23.55
&  13.28
&  52.89
&  48.85\\

CoKV(low)
& 26.61
& 17.62
& 55.35
& 49.92  \\

\bottomrule

\multicolumn{5}{c}{\textbf{Masking 64 groups}} \\  
\midrule[0.5pt]

Random   
&  16.93
&  15.65
&  38.07
&  28.39\\
HeadKV-R2(top) 
&  13.79
&  8.07
&  26.25
&  21.23\\
CoKV(top)
& 4.52
& 2.11
& 13.14
& 13.31  \\

HeadKV-R2(low) 
&  14.26
&  7.41
&  38.78
&  39.22\\ 
CoKV(low)
& 33.11
& 16.97
& 52.68
& 49.54  \\

\bottomrule

\multicolumn{5}{c}{\textbf{Masking 96 groups}} \\  
\midrule[0.5pt]

Random    
&  10.41
&  8.43
&  21.30
&  17.65\\
HeadKV-R2(top) 
&  14.22
&  3.99
&  17.73
&  16.24\\
CoKV(top)
& 2.09
& 3.04
& 10.96
& 8.32  \\

HeadKV-R2(low)
&  10.10
&  5.44
&  34.95
&  32.57\\ 
CoKV(low)
& 31.51
& 17.39
& 47.71
& 45.37  \\

\bottomrule

\multicolumn{5}{c}{\textbf{Masking 128 groups}} \\  
\midrule[0.5pt]

Random   
&  7.80
&  2.07
&  23.74
&  18.34\\
HeadKV-R2(top)   
&  6.15
&  3.89
&  18.68
&  18.52\\
CoKV(top)
& 1.19
& 3.42
& 9.81
& 6.0  \\
HeadKV-R2(low)
&  7.20
&  6.64
&  24.09
&  24.48\\ 
CoKV(low)
& 23.76
& 12.12
& 42.01
& 36.7  \\

\bottomrule

\end{tabular}
\footnotesize
\end{adjustbox}
\end{table}
\begin{figure*}[htbp]
    \centering
    \subfigure{
        \begin{minipage}[b]{0.45\textwidth}
		\includegraphics[width=1\linewidth]{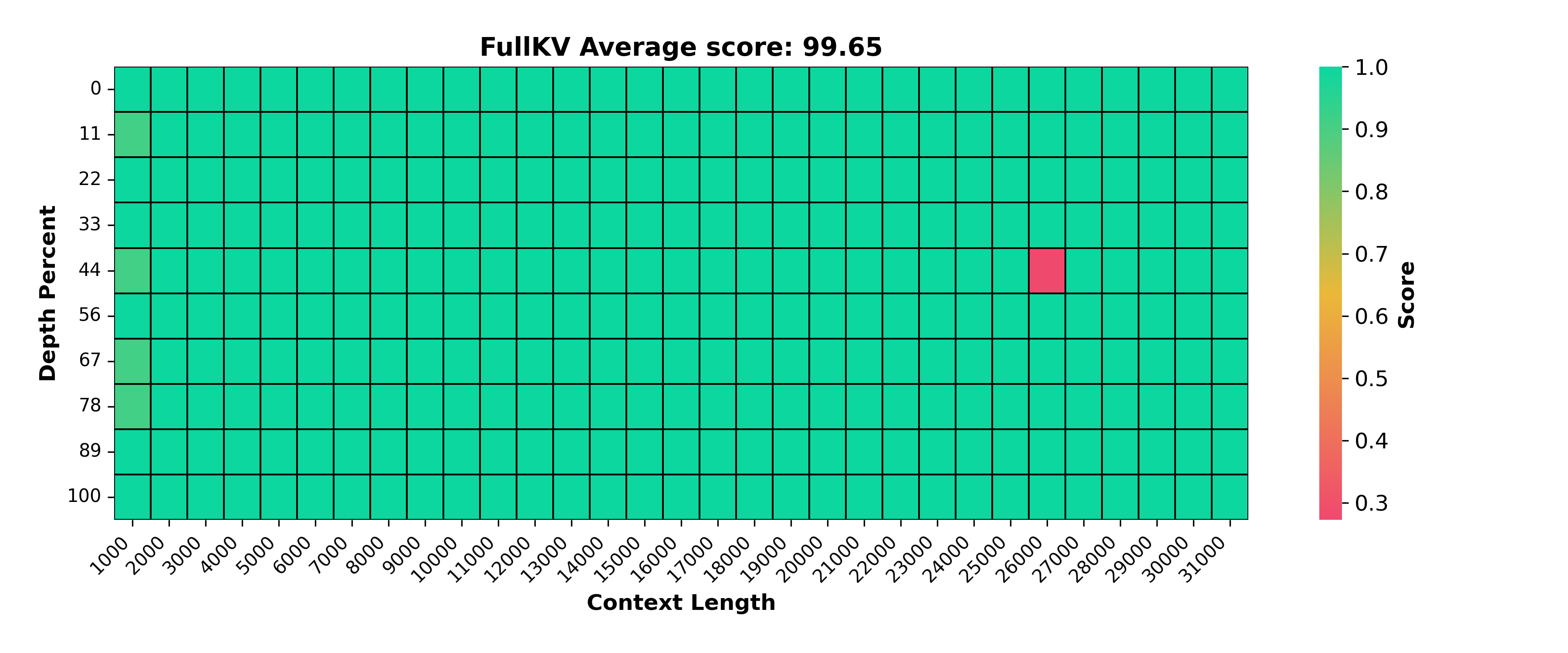}
            \label{fig:base}
        \end{minipage}%
    }
    \subfigure{
        \begin{minipage}[b]{0.45\textwidth}
            \includegraphics[width=1\linewidth]{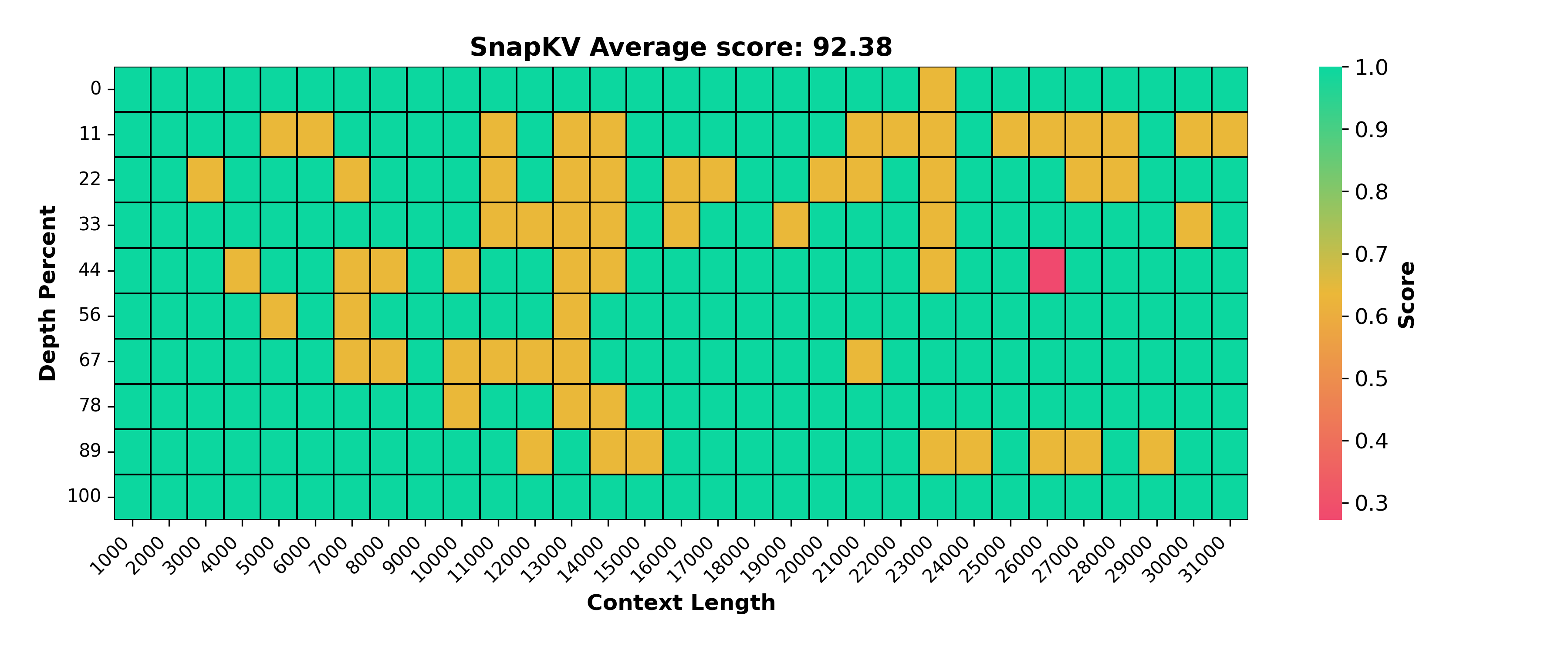}
            \label{fig:snap}
        \end{minipage}
    }
    \subfigure{
        \begin{minipage}[b]{0.45\textwidth}
            \includegraphics[width=1\linewidth]{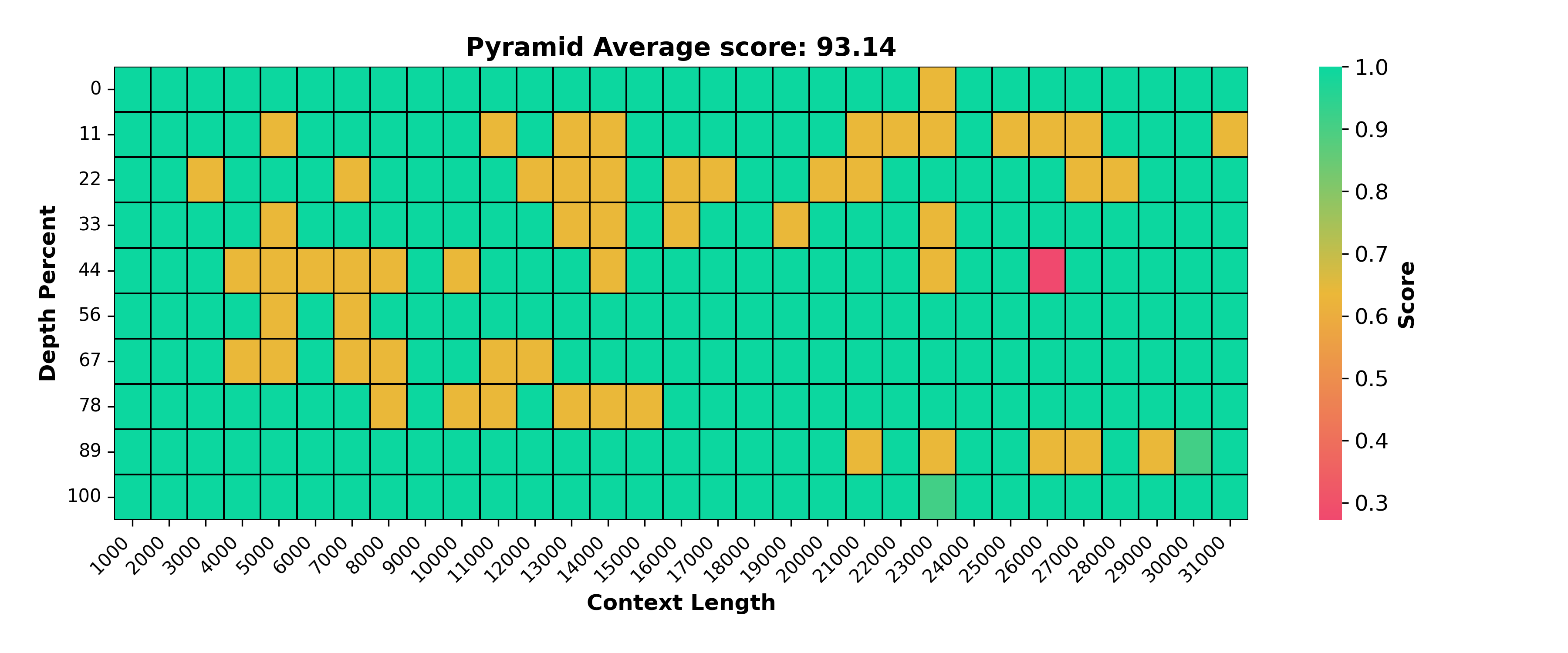}
            \label{fig:pyramid}
        \end{minipage}{}
    }
    \subfigure{
        \begin{minipage}[b]{0.45\textwidth}
            \includegraphics[width=1\linewidth]{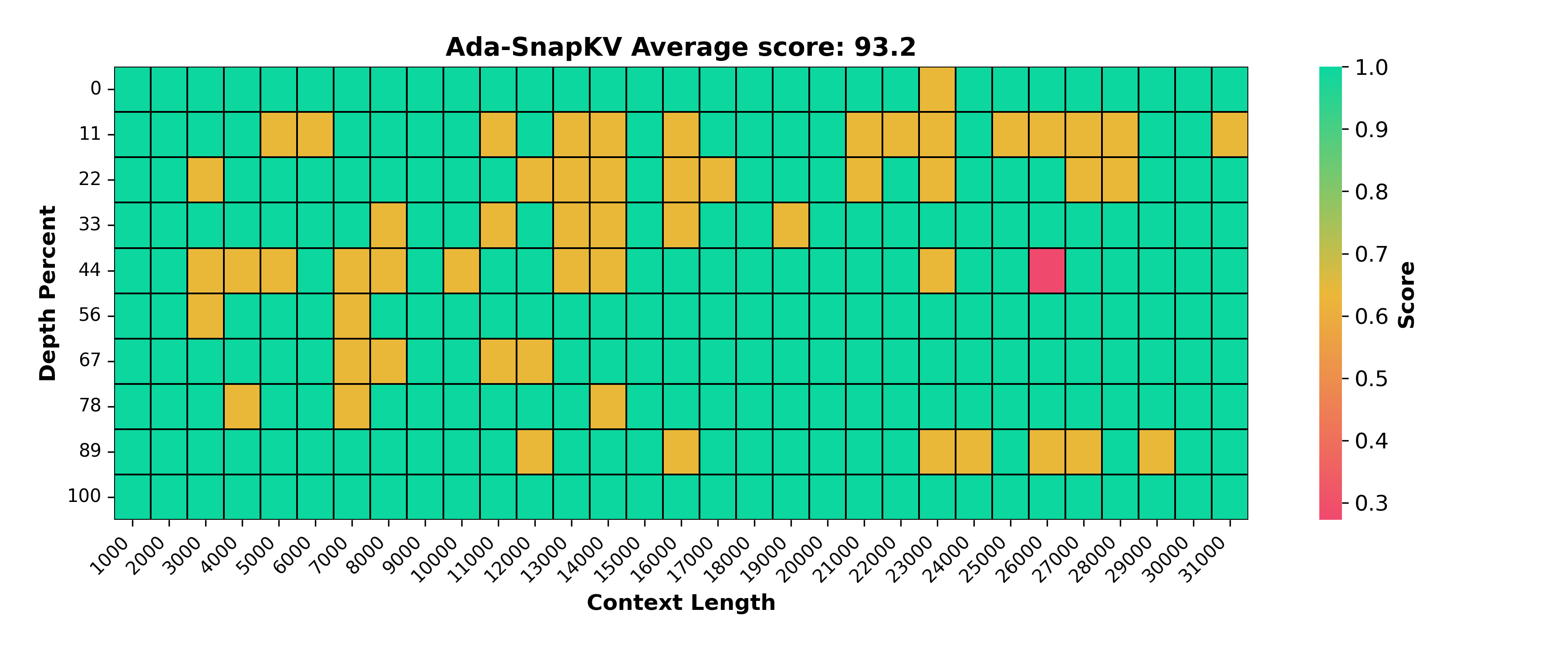}
        \label{fig:adakv}
        \end{minipage}{}
    }
    \subfigure{
        \begin{minipage}[b]{0.45\textwidth}
            \includegraphics[width=1\linewidth]{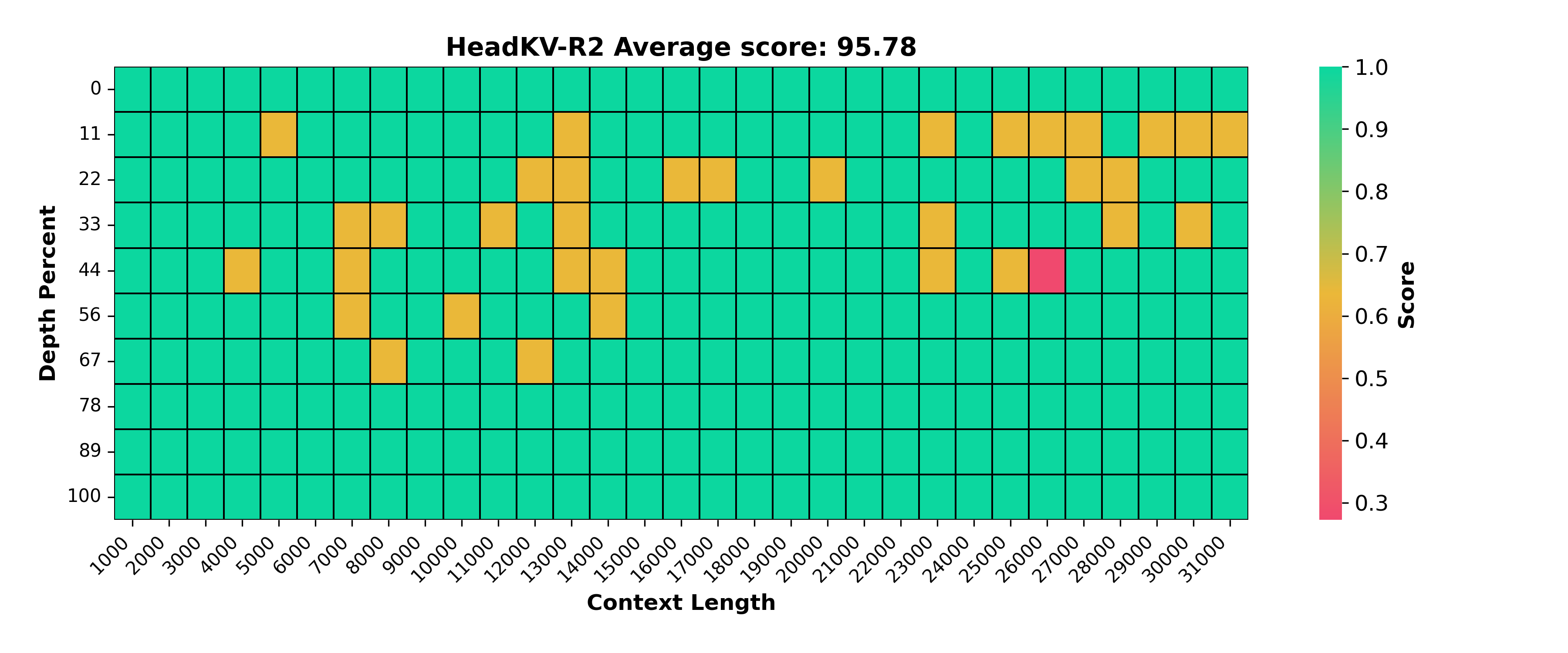}
             \label{fig:headkv}
        \end{minipage}{}
    }
    \subfigure{
        \begin{minipage}[b]{0.45\textwidth}
            \includegraphics[width=1\linewidth]{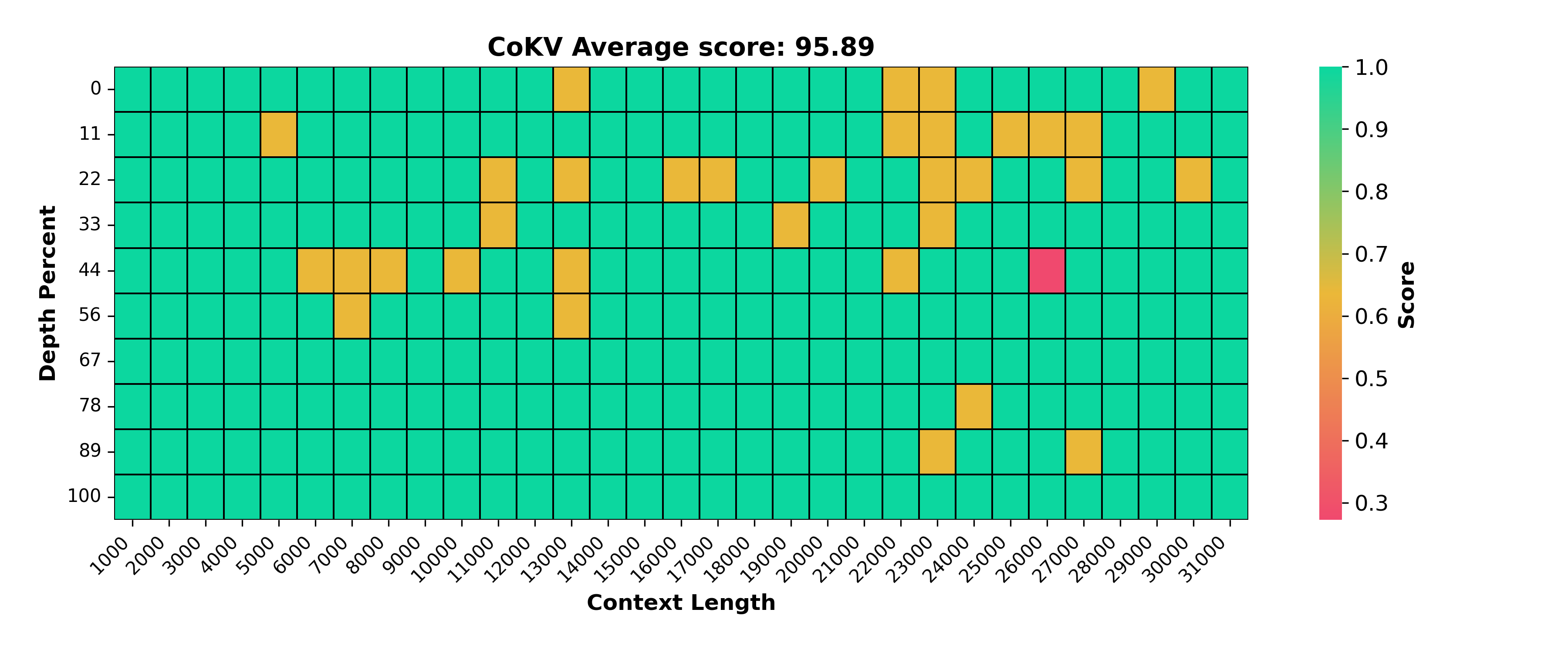}
            \label{fig:cokv}
        \end{minipage}{}
    }

    \caption{Needle-in-a-Haystack test results on  Mistral-7B-v0.2 with average KV cache = 128}
    \label{fig:all_niah}
\end{figure*}

\end{document}